\documentclass[runningheads]{llncs}

\usepackage{eccv}

\usepackage{eccvabbrv}

\usepackage{graphicx}
\usepackage{booktabs}
\usepackage{multirow}
\usepackage{makecell}
\usepackage[table]{xcolor}
\usepackage{wrapfig}
\usepackage[accsupp]{axessibility}

\usepackage{array}
\usepackage{tcolorbox}
\usepackage{enumitem}
\usepackage{algorithm}
\usepackage{algpseudocode}
\usepackage{listings}
\usepackage{tabularx}
\tcbuselibrary{breakable,listings}
\makeatletter
\renewcommand{\theHALG@line}{\thealgorithm.\arabic{ALG@line}}
\makeatother
\newtcblisting{promptlisting}[2][]{
  breakable,
  listing only,
  colback=eccvblue!3,
  colframe=eccvblue!65!black,
  coltitle=black,
  boxrule=0.7pt,
  arc=2mm,
  left=2mm,
  right=2mm,
  top=1mm,
  bottom=1mm,
  title={#2},
  fonttitle=\bfseries\fontfamily{ptm}\selectfont,
  colbacktitle=eccvblue!14,
  listing options={
    basicstyle=\fontfamily{ptm}\selectfont\normalsize,
    breaklines=true,
    breakatwhitespace=true,
    breakautoindent=false,
    breakindent=0pt,
    columns=fullflexible,
    keepspaces=true,
    showstringspaces=false
  },
  #1
}

\lstdefinestyle{caselog}{
  basicstyle=\ttfamily\fontsize{5}{5.6}\selectfont,
  breaklines=true,
  breakatwhitespace=false,
  breakautoindent=false,
  columns=fullflexible,
  keepspaces=true,
  showstringspaces=false,
  frame=single,
  framerule=0.3pt,
  rulecolor=\color{black!15},
  backgroundcolor=\color{black!2},
  xleftmargin=0pt,
  xrightmargin=0pt,
  aboveskip=2pt,
  belowskip=0pt
}

\newcolumntype{Y}{>{\raggedright\arraybackslash}X}
\newcommand{\casepanelwidth}{0.188\textwidth}
\newcommand{\caseimgwidth}{0.98\linewidth}
\newcommand{\caseinfowidth}{0.98\linewidth}
\newcommand{\caseinfoheight}{0.074\textheight}
\newcommand{\casefield}[2]{\textbf{#1}\par #2\par}
\newcommand{\casepositive}{\textcolor{green!45!black}{\textbf{POSITIVE}}}
\newcommand{\casenegative}{\textcolor{red!70!black}{\textbf{NEGATIVE}}}
\newcommand{\casejson}[1]{{\ttfamily\fontsize{3.55}{3.95}\selectfont\raggedright\detokenize{#1}\par}}
\newcommand{\caseinfo}[2]{%
  \begin{tcolorbox}[
    width=\caseinfowidth,
    height=\caseinfoheight,
    colback=black!2,
    colframe=black!18,
    boxrule=0.35pt,
    arc=0.6mm,
    left=0.9mm,
    right=0.9mm,
    top=0.9mm,
    bottom=0.9mm,
    valign=top
  ]
    \begin{minipage}[t]{\linewidth}
      \raggedright
      \fontsize{4.45}{5.0}\selectfont
      #2
    \end{minipage}
  \end{tcolorbox}%
}
\newcommand{\casepanel}[3]{%
  \begin{minipage}[t]{\casepanelwidth}
    \centering
    \begin{tcolorbox}[
      width=\linewidth,
      colback=white,
      colframe=black!28,
      boxrule=0.45pt,
      arc=0.8mm,
      left=0mm,
      right=0mm,
      top=0mm,
      bottom=0.55mm
    ]
      \vspace{0.15mm}
      {\centering\fontsize{6.2}{6.8}\selectfont\textbf{#1}\par}
      \vspace{0.35mm}
      #2
      \vspace{0.65mm}
      #3
    \end{tcolorbox}
    \vspace{0pt}
  \end{minipage}%
}
\newcommand{\caseimage}[1]{\includegraphics[width=\caseimgwidth]{#1}}

\newcommand{\caseblankplaceholder}{%
  \begin{tcolorbox}[
    width=\caseimgwidth,
    height=0.425\textwidth,
    colback=white,
    colframe=black!12,
    boxrule=0.35pt,
    arc=0.6mm,
    left=0mm,
    right=0mm,
    top=0mm,
    bottom=0mm,
    valign=center
  ]
  \end{tcolorbox}%
}

\usepackage{hyperref}
\hypersetup{hidelinks}

\begin{document}

\title{\texorpdfstring{AnchorGUI: Asymmetric Memory for \\ Dual-Scale Learning in GUI Navigation}{AnchorGUI: Asymmetric Memory for Dual-Scale Learning in GUI Navigation}}
\titlerunning{Asymmetric Memory for Dual-Scale Learning in GUI Navigation}

\author{Shengjie Jin\inst{1} \and
Zelong Sun\inst{1} \and
Hengbo Xu\inst{1} \and
Yanbiao Ma\inst{1} \and
Zhiwu Lu\inst{1}\thanks{Corresponding author.}}

\authorrunning{S.~Jin et al.}

\institute{Gaoling School of Artificial Intelligence, Renmin University of China, Beijing, China\\
\email{\{jinshengjie,luzhiwu\}@ruc.edu.cn}}

\maketitle

\begin{abstract}
Vision-Language Models (VLMs) enable autonomous GUI navigation, but agents still struggle to process and learn from dense, continuous visual histories. This bottleneck hinders both immediate error correction within a single episode (\textbf{intra-trial}) and experience distillation across multiple attempts (\textbf{cross-trial}). We trace these challenges to an empirical \emph{informational asymmetry} in GUI navigation: while expected transitions can often be compressed into lightweight textual summaries, unexpected outcomes benefit from preserved screenshots as causal evidence for accurate diagnosis. Building on this insight, we propose \textbf{AnchorGUI}, a unified framework driven by the \textbf{Cognitive State Anchor (CSA)}. The CSA acts as a per-step primitive that actively compares expected and observed transitions, converting passive multimodal trajectories into explicit prediction-error signals. These signals orchestrate a \textbf{dual-scale learning mechanism} via an asymmetric memory. For \emph{intra-trial correction}, a sliding window selectively retains visual evidence for detected mismatches, providing immediate, visually-grounded feedback. For \emph{cross-trial distillation}, this asymmetric memory focuses the computationally expensive credit assignment search space on likely failure steps. Experiments across four benchmarks validate the effectiveness of our approach. On AndroidWorld, AnchorGUI achieves a 57.3\% success rate with a 2.4$\times$ token reduction per step. Furthermore, cross-trial distillation reaches 69.2\% success (+11.9\% gain), significantly outperforming standard reflection methods while maintaining sub-linear context scaling.
\end{abstract}

\section{Introduction}

\begin{figure*}[t]
\centering
\includegraphics[width=0.98\textwidth]{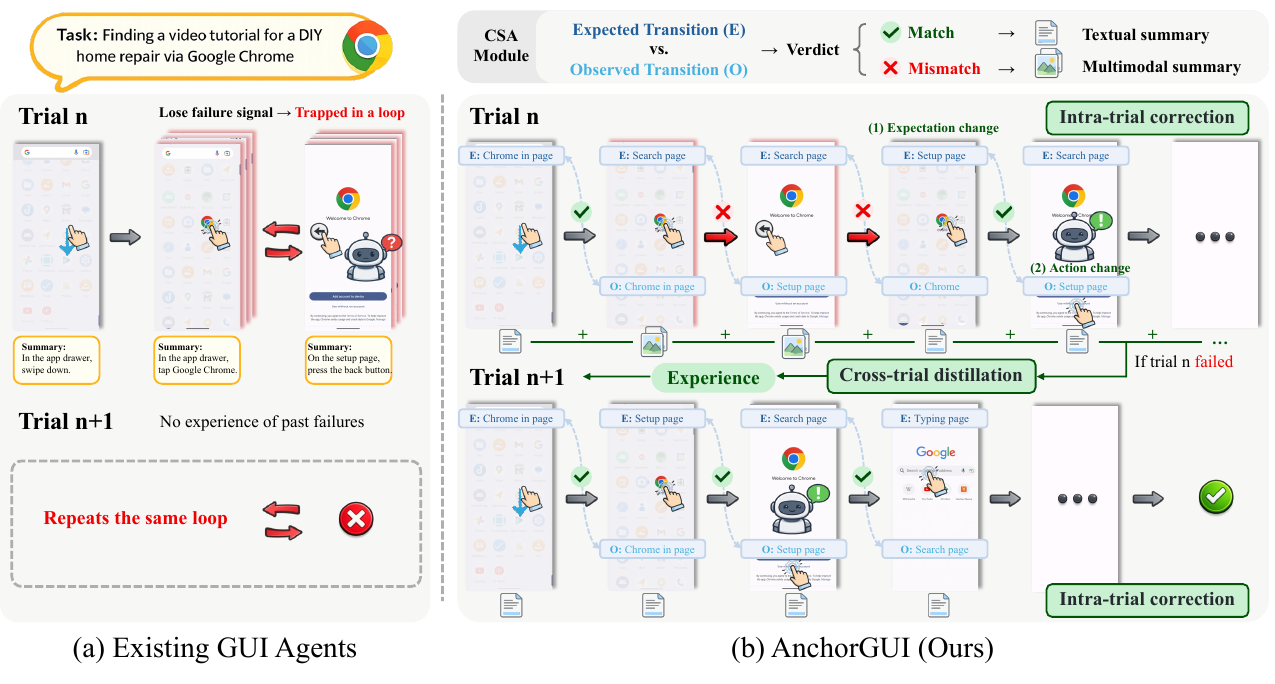}
\caption{\textbf{Comparison of existing GUI agents and AnchorGUI.} 
(a) Existing methods indiscriminately compress visual histories, discarding critical visual evidence of unexpected states. This causes agents to lose failure signals and repeat the same mistakes across trials. 
(b) \textbf{AnchorGUI} introduces Cognitive State Anchors (CSA) to explicitly compare expected versus observed transitions. By selectively retaining original screenshots for mismatched steps while compressing matched ones, it builds an asymmetric memory that enables efficient intra-trial correction and cross-trial distillation while avoiding multimodal token explosion.}
\label{fig_intro}
\end{figure*}

Graphical User Interface (GUI) navigation requires autonomous agents to perceive visual interfaces and execute actions to accomplish user-specified goals~\cite{zhang2024large,wang2024gui,nguyen2025gui}. This capability underpins a wide range of applications, from automated software testing to intelligent personal assistants~\cite{zhao2024gui,liu2025llm}. Recent advances in Vision-Language Models (VLMs)~\cite{team2024gemini,hurst2024gpt,bai2025qwen3} have substantially expanded the potential of such agents, enabling them to navigate complex visual environments~\cite{wang2023voyager,zhang2025appagent}. However, as illustrated in Figure~\ref{fig_intro}, existing GUI agents face a significant bottleneck in \textbf{effectively processing and learning from dense, continuous visual interaction histories}, both within a single episode (\textbf{intra-trial}) and across multiple attempts (\textbf{cross-trial}).

Currently, most existing GUI agents prioritize \textbf{intra-trial} decision-making, predicting the next action based on accumulated interaction history~\cite{xu2024aguvis,yang2025aria}. To mitigate multimodal token explosion, recent methods~\cite{qin2025ui,wu2025resum,liu2025pal,wu2025auto} typically resort to aggressive truncation or the compression of visual histories into textual summaries. However, this indiscriminate compression treats successful and failed actions identically, discarding error signals essential for behavioral correction~\cite{liu2025infigui,xiao2025ui} and preventing agents from recognizing and learning from their own mistakes. For instance, as shown in Figure~\ref{fig_intro}, consider an agent searching for a video that becomes stuck on a ``Welcome to Chrome'' setup screen. If the visual evidence of this unexpected state is compressed into a generic representation indistinguishable from a normal search interface, the agent may repeatedly press ``Back'' and relaunch the app, trapped in a loop it cannot recognize or escape.

In practice, human users naturally overcome such deadlocks through \textbf{cross-trial} learning, distilling experience from past failures to ensure future success. While global reflection mechanisms have proven effective in text-based environments~\cite{shinn2023reflexion,shridhar2020alfworld,yao2022react}, applying them to visually grounded GUI tasks introduces a severe \emph{credit assignment bottleneck}. Unlike discrete text sequences, GUI trajectories consist of high-dimensional visual observations that often obscure the causal chain between actions and outcomes. Pinpointing the specific action responsible for a failure among dozens of visually similar steps is conceptually intractable and computationally expensive. Consequently, agents struggle to extract actionable insights from failures, even after multiple attempts.

These challenges in both intra-trial and cross-trial learning share a common root: the lack of a principled mechanism to distinguish which steps benefit from visual evidence and which can be abstracted. Our experiments reveal an empirical \emph{informational asymmetry} in GUI navigation: the representational requirement of a step depends on whether the observed outcome aligns with the agent's expectations. When an action yields an anticipated visual transition (\eg, a menu opening as expected), the state change can be effectively encapsulated by a lightweight textual summary. Conversely, when the observation deviates from expectation, this mismatch signals a breakdown in causal understanding. In such cases, textual abstraction may be insufficient to capture the complex dependency between the initial visual state and the failed action; the original screenshot can be retained as causal evidence for accurate diagnosis. This asymmetry can be used to address both challenges: for intra-trial learning, retaining visual evidence of mismatches enables immediate error correction; for cross-trial learning, it allows the agent to precisely locate failure-inducing actions, effectively bounding the search space for credit assignment.

Building on this insight, we propose \textbf{AnchorGUI}, a unified framework that addresses both intra-trial and cross-trial learning through a single architectural principle. At its core, AnchorGUI introduces the \textbf{Cognitive State Anchor (CSA)}, a per-step structured representation comprising an \emph{Expected Transition}, an \emph{Observed Transition}, and a binary \emph{Verdict} indicating their alignment. The CSA anchors agent attention to specific states responsible for errors, preventing them from being lost in the stream of visual observations. Rather than passively recording trajectories, CSA actively compares pre-action expectations against post-action visual realities, converting dense multimodal streams into explicit prediction-error signals. These signals then orchestrate a \textbf{dual-scale learning mechanism}. For intra-trial correction, a sliding window maintains recent steps with their CSAs, selectively retaining visual observations for detected mismatches to provide immediate, visually-grounded feedback. For cross-trial distillation, this asymmetric memory aggregates across the entire episode, condensing expectation-consistent steps into text while preserving visual evidence primarily for detected mismatches. By exploiting informational asymmetry, AnchorGUI focuses the credit assignment search space on detected mismatches, enabling efficient experience extraction while drastically reducing multimodal token consumption at both temporal scales.

Our contributions are summarized as follows:
\textbf{(1) }We identify the principle of \textbf{informational asymmetry} in GUI navigation, which provides a principled basis for selective visual retention based on expectation alignment.
\textbf{(2)} We propose \textbf{AnchorGUI}, a unified framework with \textbf{CSA} module, which actively contrasts expectations with observations to selectively retain visual evidence during compression.
\textbf{(3)} We propose an asymmetric memory for both \textbf{intra-trial correction} and \textbf{cross-trial distillation}, which focuses credit assignment on detected mismatches.
\textbf{(4)} Experiments on four benchmarks demonstrate AnchorGUI's effectiveness: it achieves a \textbf{57.3\%} success rate on AndroidWorld with \textbf{2.4$\times$} lower token cost (single-trial), and improves to \textbf{69.2\%} (\textbf{+11.9\%}) via asymmetric distillation (cross-trial), outperforming standard reflection methods.

\section{Related Work}

\subsection{History Management in GUI Navigation}

Long-horizon GUI navigation requires agents to condition decisions on extended interaction histories under dynamically evolving UI states~\cite{rawles2024androidworld,xu2025androidlab}. To manage growing context, existing methods adopt three strategies: (i) \emph{context truncation}~\cite{qin2025ui}, which discards early steps entirely; (ii) \emph{periodic summarization}~\cite{wu2025resum,gao2025chain}, which compresses trajectories into text; and (iii) \emph{hierarchical planning}~\cite{wang2023mobile,wang2024mobile,ye2025mobile,agashe2025agent}, which decomposes tasks into subgoals to reduce the effective reasoning horizon. While these approaches mitigate token overhead, they share a common limitation: visual histories are primarily treated as passive context for next-step prediction, often discarding the fine-grained visual evidence required for diagnosing why a specific action failed. This loss of causal information severely hinders both immediate error correction and cross-trial knowledge transfer.

\subsection{Step-Level Verification in GUI Agents}

Recent work introduces step-level verification or feedback mechanisms to improve action reliability. GUI-Critic~\cite{wanyan2025look} employs a pre-execution critic to predict action outcomes, while MobileUse~\cite{li2025mobileuse} and Mobile-Agent-v3~\cite{ye2025mobile} integrates post-execution validation. These methods enable local error correction within a single trial. However, the verification signal is typically consumed immediately and then discarded, rarely being preserved or aggregated for future attempts. Consequently, an agent that encounters an unexpected pop-up in trial $k$ has no mechanism to explicitly recall this experience in trial $k+1$, forcing it to re-discover the same solution from scratch.

\subsection{Cross-Trial Learning in Agents}

Reflexion~\cite{shinn2023reflexion} demonstrates that verbal self-reflection enables effective cross-trial learning in text-based environments~\cite{shridhar2020alfworld,yao2022react}. However, directly applying this paradigm to GUI navigation introduces a multimodal reasoning challenge. Retaining full visual trajectories for cross-trial reasoning causes \emph{context collapse}: the VLM must locate failure causes within an overwhelming volume of largely irrelevant screenshots, diluting attention and degrading credit assignment. Na\"ively combining step-level verification with cross-trial reflection does not resolve this—reasoning still spans entire visual histories. The core challenge is to preserve only causal visual evidence while discarding redundant frames.
\section{Methodology}

\begin{figure}[t]
\centering
\includegraphics[width=\textwidth]{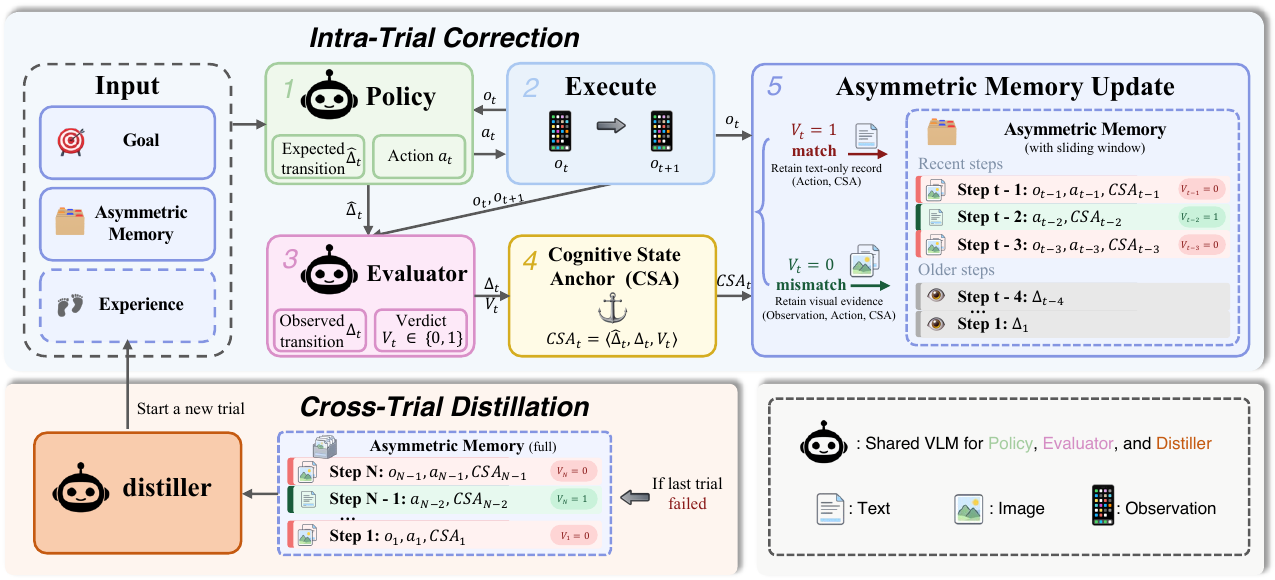}
\caption{
\textbf{Overview of AnchorGUI.}
At each step, the policy predicts an action together with an expected transition.
After execution, the evaluator compares the expected transition with the observed GUI change and produces a Cognitive State Anchor (CSA).
CSAs drive dual-scale learning:
(\textit{Top}) intra-trial correction via asymmetric memory updates, where mismatched steps retain visual observations while matched steps keep only text records within a sliding window, enabling efficient in-context correction; and
(\textit{Bottom}) cross-trial distillation that converts asymmetric memory into reusable experience when a trial fails, bounding the credit assignment search space.
}
\label{fig:framework}
\end{figure}

In this section, we present AnchorGUI, a unified framework designed to empower agents to effectively process and learn from dense, continuous visual interaction histories (see Figure~\ref{fig:framework}). While existing agents often struggle with the inherent high-dimensionality of GUI trajectories, AnchorGUI introduces a principled mechanism to distinguish and retain critical visual evidence. We first formulate the visually-grounded decision process (Sec.~\ref{sec:problem}). We then introduce the \textbf{Cognitive State Anchor (CSA)}, a structured representation that operationalizes the informational asymmetry between expected and observed transitions (Sec.~\ref{sec:csa}). Finally, we detail how this design drives a dual-scale learning mechanism, enabling efficient intra-trial correction and cross-trial distillation without suffering from multimodal token explosion or intractable credit assignment (Sec.~\ref{sec:dual}).

\subsection{Problem Formulation}
\label{sec:problem}

We define GUI navigation as a visually grounded Markov Decision Process (MDP)~\cite{sutton1998reinforcement}, formalized by the tuple
\begin{equation}
\mathcal{M} = \langle \mathcal{O}, \mathcal{A}, \mathcal{T}, G \rangle,
\end{equation}
where $\mathcal{O}$ denotes the observation space, $\mathcal{A}$ the action space, $\mathcal{T}$ the environment transition function, and $G$ a natural language goal describing the target task.

At each step $t$, the agent receives a visual observation
\begin{equation}
o_t \in \mathcal{O} = \mathbb{R}^{H \times W \times 3},
\end{equation}
corresponding to the current GUI screenshot. Let $\mathcal{H}_t$ denote the interaction history up to step $t$. Conditioned on the goal $G$ and history $\mathcal{H}_t$, the policy $\pi_{\theta}$ predicts an action
\begin{equation}
a_t \sim \pi_{\theta}(a \mid G, o_t, \mathcal{H}_t), \quad a_t \in \mathcal{A}.
\end{equation}

Following prior conventions~\cite{qin2025ui}, the action space $\mathcal{A}$ consists of predefined primitive GUI operations, including \emph{click}, \emph{swipe}, \emph{type}, and \emph{terminate}. Spatial actions are parameterized by continuous 2D coordinates $(x,y)$ defined on the visual observation $o_t$.
After executing action $a_t$, the environment transitions to the next observation according to
\begin{equation}
o_{t+1} \sim \mathcal{T}(o_t, a_t),
\end{equation}
where $\mathcal{T}$ captures stochastic GUI dynamics such as rendering delays, animation transitions, or unexpected pop-ups. The episode terminates when the agent selects the \emph{terminate} action or when a predefined maximum horizon $T$ is reached.

\subsection{Cognitive State Anchor: Bridging Expectation and Observation}
\label{sec:csa}

To enable reliable credit assignment without indiscriminate compression, we introduce the \emph{Cognitive State Anchor} (CSA). CSA acts as a principled interface between dense multimodal trajectories and reusable knowledge by explicitly modeling the alignment between the agent's internal belief and actual environment dynamics. It converts passive interaction histories into explicit prediction-error signals, forming the computational primitive for our framework.

AnchorGUI uses a single Vision-Language Model (VLM), parameterized by $\theta$, with role-specific prompts for the \emph{policy} ($\pi_{\theta}$), \emph{evaluator} ($\phi_{\theta}$), and \emph{distiller} ($\psi_{\theta}$). At step $t$, CSA is defined as:

\begin{equation}
CSA_t = \langle \hat{\Delta}_t, \Delta_t, V_t \rangle.
\end{equation}

\textbf{Expected Transition ($\hat{\Delta}_t$).} Before execution, the policy role $\pi_{\theta}$ jointly predicts the spatial action and its anticipated visual effect:
\begin{equation}
(a_t, \hat{\Delta}_t) = \pi_{\theta}(G, o_t, \mathcal{H}_t),
\end{equation}
where $\hat{\Delta}_t$ is formulated as a concise textual description (e.g., ``\textit{The 'Login' button will disappear, and the home feed will load}'').

\textbf{Observed Transition and Verdict ($\Delta_t, V_t$).} Upon reaching $o_{t+1}$, the evaluator role $\phi_{\theta}$ compares the pre-action expectation against the post-action reality:
\begin{equation}
(\Delta_t, V_t) = \phi_{\theta}(\hat{\Delta}_t, o_t, o_{t+1}),
\end{equation}
where $\Delta_t$ describes the actual UI change, and $V_t \in \{0,1\}$ is a binary verdict indicating expectation alignment ($1$ for expectation-consistent, $0$ for mismatch). To capture fine-grained visual differences, $o_t$ and $o_{t+1}$ are fed to $\phi_{\theta}$ as a temporally ordered image sequence.

\subsection{Dual-Scale Learning via Asymmetric Memory}
\label{sec:dual}

Building upon these prediction-error signals, CSA drives the agent's learning cycle across two complementary temporal scales, exploiting informational asymmetry to optimize both intra-trial correction and cross-trial transfer.

\subsubsection{Intra-Trial: Immediate Correction via Selective Visual Retention}

CSAs guide single-trial behavioral correction through \textbf{in-context learning}~\cite{brown2020language}. To support immediate error diagnosis under a bounded context, we apply an asymmetric retention strategy using a \emph{sliding window} $k$.

For recent steps ($t-k \le i < t$), we selectively retain visual observations based on the verdict $V_i$, preserving screenshots for detected expectation mismatches:
\begin{equation}
\mathcal{H}_{recent} = \bigoplus_{i=t-k}^{t-1}
\begin{cases}
\langle o_i, a_i, CSA_i \rangle & \text{if } V_i = 0, \\
\langle a_i, CSA_i \rangle & \text{if } V_i = 1,
\end{cases}
\end{equation}
where $\bigoplus$ denotes temporal concatenation. By explicitly including $V_i$ in the prompt context, the policy $\pi_{\theta}$ learns to dynamically adjust its behavior: steps with $V_i = 0$ retain their visual context ($o_i$) to act as immediate, visually-grounded causal evidence for errors, while steps with $V_i = 1$ drop redundant images to save tokens, reinforcing successful local strategies purely via text.

For earlier steps ($i < t-k$), retaining images is computationally prohibitive. We thus aggressively compress them into a lightweight textual sequence of observed transitions: $\mathcal{H}_{old} = \bigoplus_{i=0}^{t-k-1} \langle \Delta_i \rangle$. The final effective context is constructed as $\mathcal{H}_t = \mathcal{H}_{old} \oplus \mathcal{H}_{recent}$. This asymmetric sliding window reduces multimodal context overhead from $\mathcal{O}(T)$ to $\mathcal{O}(|\{i \in [t-k, t) : V_i=0\}|)$, supporting scalable reasoning across longer horizons.

\subsubsection{Cross-Trial: Distillation and Bounded Credit Assignment}

Upon trial completion, the agent distills reusable strategies (e.g., handling specific pop-ups) for future attempts. Let $\tau^{(k)}$ denote the $k$-th trial, spanning $T_k$ steps. Extending the selective retention strategy to the episode level, we construct a global \textbf{Asymmetric Memory} $\mathcal{H}_{asym}^{(k)}$ that aggregates the entire trial:
\begin{equation}
\mathcal{H}_{asym}^{(k)} = \bigoplus_{i=0}^{T_k}
\begin{cases}
\langle o_i, a_i, CSA_i \rangle & \text{if } V_i = 0, \\
\langle a_i, CSA_i \rangle & \text{if } V_i = 1.
\end{cases}
\end{equation}

As established, the \emph{informational asymmetry} in GUI navigation motivates that successful actions ($V_i=1$) can often be abstracted into text, whereas failed actions ($V_i=0$) benefit from the original visual state ($o_i$) to resolve causal ambiguity. By localizing negative verdicts, this asymmetric memory explicitly bounds the temporal search space for credit assignment. It reduces the multimodal reasoning complexity for the distiller from the full trajectory length $\mathcal{O}(T)$ to the detected subset of mismatched steps $\mathcal{O}(|\{i: V_i=0\}|)$, mitigating the combinatorial ambiguity inherent in global reflection.
The distiller role $\psi_{\theta}$ then processes this asymmetric memory to synthesize experience:
\begin{equation}
\mathcal{X}^{(k)} = \psi_{\theta}(\mathcal{H}_{asym}^{(k)}).
\end{equation}

Rather than na\"ively concatenating past reflections, $\mathcal{X}^{(k)}$ acts as an iteratively updated experience bank, formatted as a concise set of natural language strategic rules (e.g., ``\textit{If a pop-up blocks the screen, click the 'X' at the top right before proceeding}''). In the next trial $\tau^{(k+1)}$, the policy integrates this distilled experience to predict the next action, equipping $\pi_{\theta}$ with explicit strategies to avoid repeating past visual mistakes:
\begin{equation}
(a_t, \hat{\Delta}_t) = \pi_{\theta}(G, \mathcal{X}^{(k)}, o_t, \mathcal{H}_t).
\end{equation}

\section{Experiments}
\label{sec:experiments}

\subsection{Experimental Setup}
\label{sec:exp_setup}

\paragraph{Benchmarks.} We evaluate AnchorGUI across four GUI navigation benchmarks. To assess intra-trial reasoning and immediate error correction, we employ three offline datasets: AITZ~\cite{zhang2024android}, Android-Control-High~\cite{li2024effects}, and GUI-Odyssey~\cite{lu2025guiodyssey}. To evaluate cross-trial distillation, we conduct multi-trial experiments on AndroidWorld~\cite{rawles2024androidworld}, a dynamic benchmark that supports iterative agent-environment interactions across diverse tasks and applications.

\paragraph{Metrics.} For AITZ, Android-Control-High, and GUI-Odyssey, we report Type Match (TM), measuring whether the predicted action category aligns with the ground truth, and Exact Match (EM), which additionally requires all action parameters to be correct. For AndroidWorld, we report Success Rate (SR). In cross-trial evaluations, $\text{SR}@K$ denotes the cumulative success rate achieved by the $K$-th attempt across multiple interaction trials

\paragraph{Compared Methods.}
We compare against methods spanning two complementary dimensions. \textbf{Intra-trial baselines} operate in single-attempt settings, including: (1)~\textit{fine-tuning methods}: OS-Genesis~\cite{sun2025genesis}, Aguvis~\cite{xu2024aguvis}, OdysseyAgent~\cite{lu2025guiodyssey}, UI-TARS~\cite{qin2025ui}, GUI-Critic~\cite{wanyan2025look}, UGround~\cite{gou2024navigating}, Aria-UI~\cite{yang2025aria}, and Agent-S2~\cite{agashe2025agent}; (2)~\textit{zero-shot prompting methods}: ReAct~\cite{yao2022react}, ReSum~\cite{wu2025resum}, Truncation~\cite{qin2025ui}, M3A~\cite{rawles2024androidworld}, Self-Reflection~\cite{madaan2023self}, Mobile-Agent-v3~\cite{ye2025mobile}, and Chain-of-Memory~\cite{gao2025chain}. \textbf{Cross-trial strategies} distill experience across multiple attempts; we evaluate Reflexion~\cite{shinn2023reflexion} and Episodic Memory~\cite{shinn2023reflexion}, pairing each with different intra-trial backbones to isolate their individual contributions (Table~\ref{tab:online_cross}). Detailed baseline descriptions are provided in Appendix~\ref{sec:supp_baselines}.

\paragraph{Implementation Details.} We adopt Qwen3-VL-8B~\cite{bai2025qwen3} as the primary vision-language backbone model. We set the intra-trial sliding window size to $k{=}2$ and the maximum number of cross-trial attempts to $K{=}3$. All experiments are conducted with 3 different random seeds; we report the mean performance across runs. Complete hyperparameters and implementation details are provided in Appendix~\ref{sec:supp_implementation}. The source code will be released upon publication.

\begin{table}[t]
\centering
\caption{Performance comparison on offline GUI navigation benchmarks. All methods operate in single-trial setting. TM: Type Match; EM: Exact Match.}
\label{tab:offline_results}
\resizebox{\linewidth}{!}{
\begin{tabular}{llcccccc}
\toprule
\multirow{2}{*}{\textbf{Method}} & \multirow{2}{*}{\textbf{Model}} & \multicolumn{2}{c}{\textbf{AITZ}} & \multicolumn{2}{c}{\textbf{Android-Control}} & \multicolumn{2}{c}{\textbf{GUI-Odyssey}} \\
\cmidrule(lr){3-4} \cmidrule(lr){5-6} \cmidrule(lr){7-8}
& & \textbf{TM (\%)} & \textbf{EM (\%)} & \textbf{TM (\%)} & \textbf{EM (\%)} & \textbf{TM (\%)} & \textbf{EM (\%)} \\
\midrule

\multicolumn{8}{l}{\textbf{Fine-tuning Methods}} \\
\addlinespace[2pt]
OS-Genesis~\cite{sun2025genesis} & OS-Genesis-7B & 20.0 & 8.45 & 65.9 & 44.4 & 11.7 & 3.63 \\
Aguvis~\cite{xu2024aguvis} & Aguvis-7B & 35.7 & 19.0 & 65.6 & 54.2 & 26.7 & 13.5 \\
OdysseyAgent~\cite{lu2025guiodyssey} & OdysseyAgent & 59.2 & 31.6 & 58.8 & 32.7 & \textbf{90.8} & \textbf{73.7} \\
UI-TARS~\cite{qin2025ui} & UI-TARS-7B & \textbf{71.7} & \textbf{55.3} & \textbf{68.5} & \textbf{60.8} & 78.8 & 57.3 \\
\midrule
\addlinespace[2pt]
\multicolumn{8}{l}{\textbf{Zero-shot Methods}} \\
\addlinespace[2pt]

GPT-4o~\cite{hurst2024gpt} & GPT-4o & 70.0 & 35.3 & 63.1 & 30.9 & 37.5 & 14.2 \\
ReAct~\cite{yao2022react} & Qwen3-VL-8B & 71.4 & 57.5 & 71.8 & 70.0 & 75.4 & 58.9 \\
ReSum~\cite{wu2025resum} & Qwen3-VL-8B & 69.9 & 51.9 & 72.3 & 70.5 & 76.5 & 54.3 \\
Truncation~\cite{qin2025ui} & Qwen3-VL-8B & 71.0 & 56.7 & 72.0 & 70.5 & 78.3 & 60.3 \\
Self-Reflection~\cite{madaan2023self} & Qwen3-VL-8B & 73.9 & 58.0 & 71.9 & 70.3 & 80.1 & \textbf{60.7} \\
\rowcolor{gray!10}
\textbf{AnchorGUI (Ours)} & \textbf{Qwen3-VL-8B} & \textbf{75.2} & \textbf{59.1} & \textbf{74.6} & \textbf{73.0} & \textbf{80.6} & 60.6 \\

\bottomrule
\end{tabular}
}
\end{table}

\begin{table}[t]
\centering
\caption{Single-trial performance on AndroidWorld. SR: Success Rate.}
\label{tab:online_intra}
\resizebox{0.9\textwidth}{!}{
\tabcolsep10pt
\begin{tabular}{llc}
\toprule
\textbf{Method} & \textbf{Model} & \textbf{SR (\%)} \\
\midrule
\multicolumn{3}{l}{\textbf{Fine-tuning Methods}} \\
\addlinespace[2pt]
GUI-Critic~\cite{wanyan2025look} & GPT-4o + GUI-Critic-R1 & 29.4 \\
UGround~\cite{gou2024navigating} & GPT-4o + UGround & 32.8 \\
Aguvis~\cite{xu2024aguvis} & GPT-4o + Aguvis-7B & 37.1 \\
Aria-UI~\cite{yang2025aria} & GPT-4o + Aria-UI & 44.8 \\
UI-TARS~\cite{qin2025ui} & UI-TARS-72B-SFT & 46.6 \\
Agent-S2~\cite{agashe2025agent} & Claude-3.7-Sonnet + UI-TARS-72B-DPO & \textbf{54.3} \\
\midrule
\addlinespace[2pt]
\multicolumn{3}{l}{\textbf{Zero-shot Methods}} \\
\addlinespace[2pt]

GPT-4o\cite{hurst2024gpt} & GPT-4o & 34.5 \\
AndroidGen~\cite{lai2025androidgen} & GPT-4o & 46.8 \\
MobileUse~\cite{li2025mobileuse} & Qwen2.5-VL-7B & 21.6 \\
MobileUse~\cite{li2025mobileuse}  & Qwen2.5-VL-32B & 44.4 \\
ReAct~\cite{yao2022react} & Qwen3-VL-8B & 43.7 \\
ReSum~\cite{wu2025resum} & Qwen3-VL-8B & 45.6 \\
Truncation~\cite{qin2025ui} & Qwen3-VL-8B& 44.7  \\
Self-Reflection~\cite{madaan2023self} &Qwen3-VL-8B & 47.7 \\
M3A~\cite{rawles2024androidworld} & Qwen3-VL-8B & 39.8 \\
Mobile-Agent-v3~\cite{ye2025mobile} & Qwen3-VL-8B & 55.2 \\
Chain-of-Memory~\cite{gao2025chain} & Qwen3-VL-8B & 46.8 \\
\rowcolor{gray!10} \textbf{AnchorGUI (Ours)} & \textbf{Qwen3-VL-8B} & \textbf{57.3} \\
\bottomrule
\end{tabular}}
\end{table}

\begin{table}[t]
\centering
\caption{Cross-trial learning performance and token efficiency on AndroidWorld. We compare system-level baselines with controlled ablations using different distillation strategies. \textbf{Intra} tokens: average per-step cost including policy generation and step-level verification (e.g., evaluator for AnchorGUI, self-reflection for baseline methods). \textbf{Cross} tokens: distillation cost per trial.}
\label{tab:online_cross}
\resizebox{\linewidth}{!}{
\tabcolsep6pt
\begin{tabular}{l|ccc|c|cc}
\toprule
\multirow{2}{*}{\textbf{Method}} & \multicolumn{4}{c|}{\textbf{Success Rate (\%)}} & \multicolumn{2}{c}{\textbf{Tokens (Avg.)}} \\
\cmidrule{2-7}
& \textbf{SR@1} & \textbf{SR@2} & \textbf{SR@3} & \begin{tabular}{@{}c@{}}\textbf{$\Delta$} \\ \scriptsize{(T3-T1)}\end{tabular} & \begin{tabular}{@{}c@{}}\textbf{Intra} \\ \scriptsize{(/Step)}\end{tabular} & \begin{tabular}{@{}c@{}}\textbf{Cross} \\ \scriptsize{(/Trial)}\end{tabular} \\
\midrule
\multicolumn{7}{l}{\textbf{Baseline intra-trial + various cross-trial strategies}} \\
ReAct + Episodic Memory & 43.7 & 49.1 & 53.4 & +9.74 & 91.4k & - \\
ReAct + Reflexion & 43.7 & \textbf{53.0} & 55.2 & \textbf{+11.5} & \textbf{32.7k} & \textbf{59.2k} \\
Self-Reflection + Episodic Memory & \textbf{47.7} & 51.6 & 54.2 & +6.50 & 98.5k & - \\
Self-Reflection + Reflexion & \textbf{47.7} & 52.5 & \textbf{56.8} & +9.13 & 38.1k & 60.7k \\
\midrule
\multicolumn{7}{l}{\textbf{AnchorGUI intra-trial + various cross-trial strategies}} \\
Memoryless Retry & \textbf{57.3} & 61.2 & 62.9 & +5.51 & 13.5k & - \\
Episodic Memory & \textbf{57.3} & 60.5 & 62.6 & +5.29 & 78.9k & - \\
Reflexion & \textbf{57.3} & 63.1 & 64.4 & +7.02 & 13.6k & 66.5k \\
\rowcolor{gray!10} 
\textbf{Asymmetric Distillation (Ours)} & \textbf{57.3} & \textbf{65.4} & \textbf{69.2} & \textbf{+11.9} & \textbf{13.7k} & \textbf{24.4k} \\
\bottomrule
\end{tabular}
}
\end{table}

\begin{table}[ht!]
\centering
\caption{Ablation study on AndroidWorld. Each row incrementally adds one component to validate its contribution. \textbf{Intra} tokens: per-step cost including policy and verification. \textbf{Cross} tokens: per-trial distillation cost. $^\dagger$: memoryless retry without cross-trial distillation.}
\label{tab:component_ablation}
\resizebox{\linewidth}{!}{
\begin{tabular}{l|ccc|c|cc}
\toprule
\multirow{2}{*}{\textbf{Method / Component Added}} & \multicolumn{4}{c|}{\textbf{Success Rate (\%)}} & \multicolumn{2}{c}{\textbf{Tokens (Avg.)}} \\
\cmidrule{2-7}
& \textbf{SR@1} & \textbf{SR@2} & \textbf{SR@3} 
& \begin{tabular}{@{}c@{}}\textbf{$\Delta$} \\ \scriptsize{(T3-T1)}\end{tabular}
& \begin{tabular}{@{}c@{}}\textbf{Intra} \\ \scriptsize{(/Step)}\end{tabular}
& \begin{tabular}{@{}c@{}}\textbf{Cross} \\ \scriptsize{(/Trial)}\end{tabular} \\
\midrule
\multicolumn{7}{l}{\textbf{Intra-Trial Configurations}} \\
(1) Baseline (ReAct) & 43.7 & 50.9$^\dagger$ & 53.7$^\dagger$ & \textbf{+10.0} & 32.4k & - \\
(2) + Dual-Grained Memory with CSA & 55.3 & 60.5$^\dagger$ & \textbf{63.0}$^\dagger$ & +7.69 & 17.9k & - \\
(3) + Asymmetric Window Retention & \textbf{57.3} & \textbf{61.2}$^\dagger$ & 62.9$^\dagger$ & +5.51 & \textbf{13.5k}  & - \\
\midrule
\multicolumn{7}{l}{\textbf{Cross-Trial Configurations}} \\
(4) + Cross-Trial Distillation & \textbf{57.3} & 63.8 & 66.1 & +8.74 & 13.9k & 65.3k \\
\rowcolor{gray!10}
(5) \textbf{+ Asymmetric Distillation (AnchorGUI)} & \textbf{57.3} & \textbf{65.4} & \textbf{69.2} & \textbf{+11.9} & \textbf{13.7k} & \textbf{24.4k} \\
\bottomrule
\end{tabular}
}
\end{table}

\subsection{Main Results}
\label{sec:exp_main}

\paragraph{Intra-Trial Correction on Offline Benchmarks.}
AnchorGUI demonstrates strong zero-shot performance in static environments by explicitly preserving visual causal evidence through asymmetric retention.
As shown in Table~\ref{tab:offline_results}, compared to full-history baselines such as ReAct, our method achieves higher performance on the AITZ dataset (+3.8\% in TM and +1.6\% in EM), suggesting that selective visual retention effectively avoids the attention dilution caused by blindly stacking historical steps.
Furthermore, on the Android-Control benchmark, AnchorGUI surpasses text-compression baselines like ReSum by +2.3\% in TM and +2.5\% in EM. This trend implies that compressing post-action states into text discards the fine-grained visual evidence necessary to diagnose complex UI failures, whereas the comparative mechanism of the CSA retains this critical visual grounding.
Compared to Truncation, which discards history indiscriminately, the verdict-driven retention of AnchorGUI presents a more principled strategy, as arbitrary truncation wastes tokens on successful transitions and risks losing critical causal failure frames. While the fine-tuned OdysseyAgent achieves higher scores on the GUI-Odyssey benchmark due to domain-specific training, AnchorGUI maintains competitive cross-domain adaptability as a zero-shot method without requiring task-specific weight updates.

\paragraph{Intra-Trial Performance on AndroidWorld.}
In the dynamic AndroidWorld environment, the single-attempt success rate of AnchorGUI surpasses both standard single-agent baselines and complex multi-agent frameworks.
As shown in Table~\ref{tab:online_intra}, AnchorGUI achieves an SR@1 of 57.3\%, outperforming engineered systems such as Mobile-Agent-v3 (55.2\%) and Chain-of-Memory (46.8\%).
This outcome indicates that a single binary verdict is sufficient to align expectations throughout the decision process, bypassing the coordination overhead of multi-agent pipelines or the multi-stage procedures required to maintain short-term and long-term memory.
From an efficiency perspective, AnchorGUI not only improves the SR@1 of the identically backed ReAct baseline by +13.6\% but also effectively mitigates multimodal token explosion by reducing intra-trial consumption by a factor of 2.4 per step. This dual advantage suggests that the asymmetric memory mechanism successfully removes redundant images from successful steps, enabling a lightweight 8B model to maintain a high signal-to-noise ratio and outperform more complex architectures.

\paragraph{Cross-Trial Distillation Results.}
AnchorGUI exhibits consistent cross-trial improvement, which appears strongly correlated with its controlled treatment of the multimodal credit assignment problem.
As detailed in Table~\ref{tab:online_cross}, our approach reaches an SR@3 of 69.2\% with an absolute learning gain of +11.9\%, establishing a final performance margin of +14.0\% over ReAct + Reflexion (55.2\% SR@3). 
The significance of this gain is amplified when considering the principle of diminishing returns; achieving a +11.9\% improvement from a high initial SR@1 of 57.3\% represents a steeper learning curve than the +11.5\% gain Reflexion achieves from a much lower 43.7\% starting point.
Furthermore, AnchorGUI accomplishes this with a cross-trial distillation cost of only 24.4k tokens, compared to roughly 60k tokens required by Reflexion-style baselines. The performance plateau of the baselines suggests a severe credit assignment bottleneck caused by unbounded search spaces that dilute attention across full trajectories. By filtering out successful frames, AnchorGUI biases the model toward failure-related signals, enabling efficient, low-cost rule extraction.

\subsection{Ablation Studies}
\label{sec:exp_ablation}

\paragraph{Component Ablation Studies.}
The progressive ablation in Table~\ref{tab:component_ablation} examines how the principle of informational asymmetry influences both performance and efficiency across two temporal scales. 
At the intra-trial scale (Rows 1 to 3), introducing the explicit CSA and asymmetric retention improves the SR@1 to 57.3\% while simultaneously reducing the per-step token cost from 17.9k to 13.5k.
At the cross-trial scale (Rows 4 to 5), asymmetric distillation (Row 5) attains a higher SR@3 (69.2\%) than full visual distillation (Row 4, 66.1\%) while consuming substantially fewer tokens. 
This comparison indicates that retaining the entire visual history paradoxically weakens credit assignment by obscuring causally relevant failure signals with a large volume of successful frames. 
Additionally, comparing the memoryless retry gains helps isolate the role of environmental stochasticity; the smaller retry gain of Row 3 (+5.51\%) compared to Row 1 (+10.0\%) reflects the compressed room for stochastic improvement given Row 3's higher initial success rate. Consequently, the +11.9\% learning gain achieved by asymmetric distillation (Row 5) implies that the model successfully extracts valid cross-trial knowledge rather than merely benefiting from environmental stochasticity.

\begin{figure}[t]
    \centering
    \begin{minipage}[b]{0.48\textwidth}
        \centering
        \includegraphics[width=\textwidth]{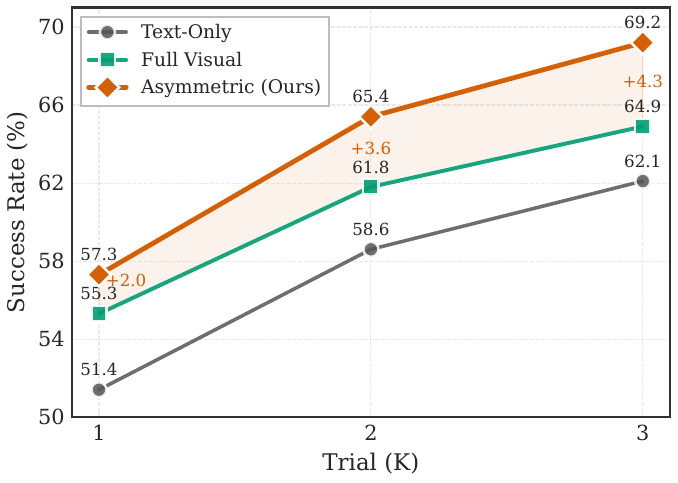}
        \caption{\textbf{Ablation on retention strategies.} Cross-trial learning curves of different visual retention strategies.}
        \label{fig:learning_curve}
    \end{minipage}
    \hfill
    \begin{minipage}[b]{0.48\textwidth}
        \centering
        \includegraphics[width=\textwidth]{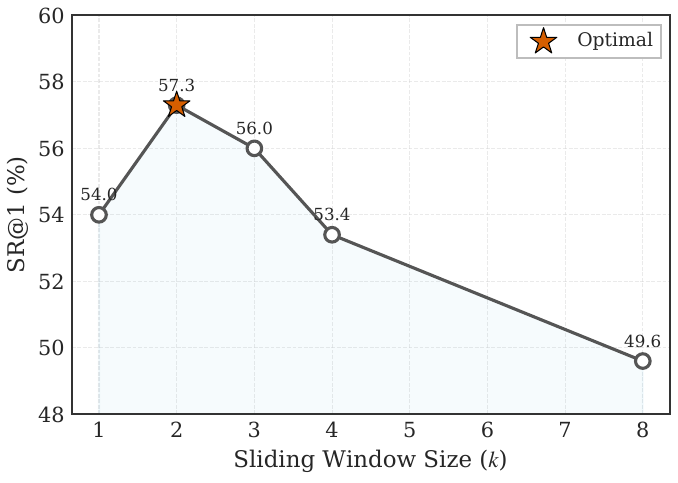}
        \caption{\textbf{Sliding window size impact.} Single-trial performance across varying sliding window sizes $k$.}
        \label{fig:sliding_window}
    \end{minipage}
\end{figure}

\paragraph{Efficacy of Asymmetric Visual Retention.} To validate our core hypothesis regarding informational asymmetry, we compare three unified retention strategies in Figure~\ref{fig:learning_curve}. The pure text retention strategy yields the lowest cross-trial performance with a 62.1\% success rate at the third attempt, suggesting that discarding all visual history severs the causal chain necessary to diagnose complex UI failures. Conversely, the full visual retention strategy improves the success rate to 64.9\% by preserving causal evidence, but it introduces significant visual noise by treating all steps equally. The asymmetric visual retention strategy achieves the steepest learning curve and the highest final success rate of 69.2\%. This trend shows that preserving visual evidence for detected mismatches filters redundant successful transitions and anchors attention on failure signals.

\paragraph{Optimal Context Bounds in Intra-Trial Dynamics.} Figure~\ref{fig:sliding_window} illustrates the trade-off in single-trial performance as the sliding window size varies. Setting the window to $k=2$ with asymmetric retention yields the optimal immediate correction capability with a 57.3\% success rate. A minimal window of $k=1$ drops performance to 54.0\%, indicating that overly extreme compression sacrifices the necessary sequential context required to diagnose cascading UI errors. More importantly, expanding the window beyond $k=3$ paradoxically degrades performance, bottoming out at 49.6\% for $k=8$. This decline reveals that even with asymmetric dropping, an excessively long temporal window accumulates historical noise and dilutes the policy attention.

\subsection{Efficiency and Scalability}
\label{sec:exp_efficiency}

\begin{figure}[t]
    \centering
    \begin{minipage}[b]{0.48\textwidth}
        \centering
        \includegraphics[width=\textwidth]{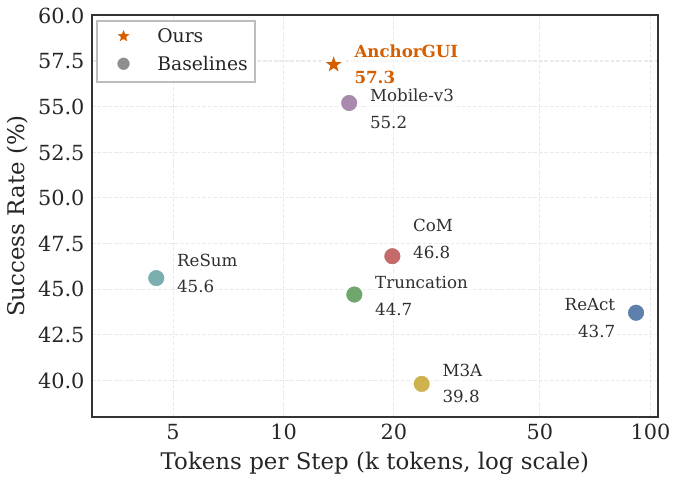}
        \caption{\textbf{Pareto frontier analysis.} Task success rate versus average token cost per step across different methods.}
        \label{fig:pareto}
    \end{minipage}
    \hfill
    \begin{minipage}[b]{0.48\textwidth}
        \centering
        \includegraphics[width=\textwidth]{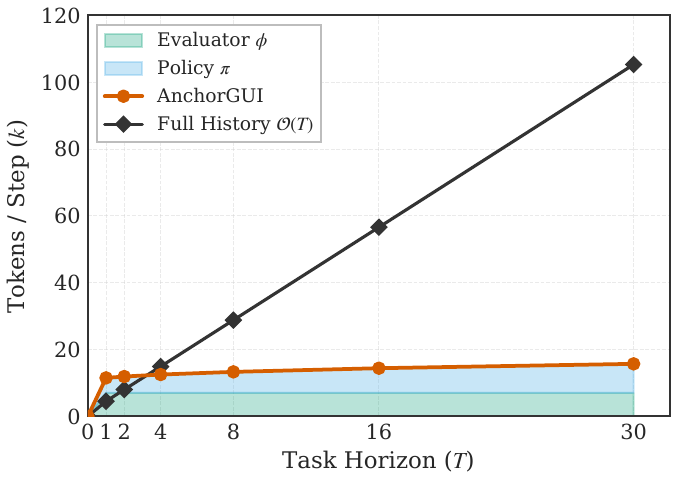}
        \caption{\textbf{Token scaling with task horizon.} Token consumption scaling as the episode length $T$ increases.}
        \label{fig:scaling}
    \end{minipage}
\end{figure}

\paragraph{Pareto Optimality in Performance and Efficiency.} Figure~\ref{fig:pareto} plots the Pareto frontier of task success rate versus total token cost per step, showing that AnchorGUI occupies the optimal top-left quadrant with a 57.3\% success rate at an average cost of only 13.7k tokens. In contrast, standard uncompressed methods like ReAct consume a prohibitive 91.4k tokens for a much lower 43.7\% success rate, and even engineered baselines like Mobile-Agent-v3 achieve 55.2\% while consuming 15.1k tokens. This demonstrates that the localized 7k token overhead of the evaluator is more than offset by its ability to aggressively prune redundant visual histories, making the overall system highly cost-effective.

\paragraph{Token Scaling with Task Horizon.} A central property of AnchorGUI is that its multimodal context complexity remains bounded by $\mathcal{O}(k)$, independent of episode length $\mathcal{O}(T)$.
As shown in Figure~\ref{fig:scaling}, token consumption remains stable as interaction steps increase; a standard trajectory accumulation approach such as ReAct explodes linearly, reaching 105.3k tokens by step 30. In contrast, the policy token consumption of AnchorGUI remains remarkably stable, growing marginally from 4.5k tokens at step 1 to just 8.7k tokens at step 30. Even when adding the fixed 7k token overhead from the evaluator, the total cost at step 30 remains tightly constrained to 15.7k tokens. This sub-linear scaling implies that AnchorGUI effectively eliminates the multimodal token explosion bottleneck, which is a fundamental prerequisite for deploying autonomous agents in long-horizon navigation tasks.

\subsection{Robustness and Generalization}
\label{sec:exp_robustness}

\paragraph{Reliability of the Cognitive State Anchor.} The efficacy of asymmetric memory depends on the accuracy of the evaluator’s binary verdict. We manually annotated 535 interaction steps (see Appendix~\ref{sec:supp_reliability} for details). The zero-shot evaluator achieves 99.2\% precision and 94.1\% recall. From a system perspective, the extremely low false positive rate (0.56\%) ensures that critical visual evidence of root failures is rarely discarded. The modest false negative rate (4.5\%) incurs only negligible token overhead by occasionally retaining redundant successful frames, without affecting downstream causal reasoning. Overall, modern VLMs can reliably act as automated judges of expectation alignment, providing a solid foundation for asymmetric history orchestration.

\begin{figure}[t]
    \centering
    \begin{subfigure}[b]{0.48\textwidth}
        \centering
        \includegraphics[width=\textwidth]{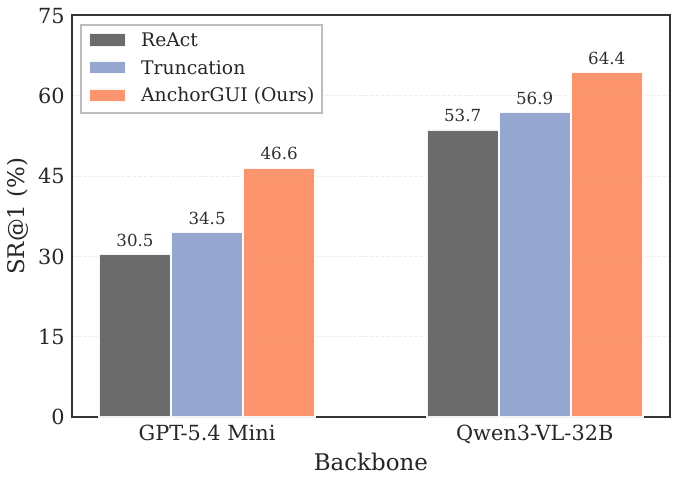}
        \caption{Intra-trial performance (SR@1)}
    \end{subfigure}
    \hfill
    \begin{subfigure}[b]{0.48\textwidth}
        \centering
        \includegraphics[width=\textwidth]{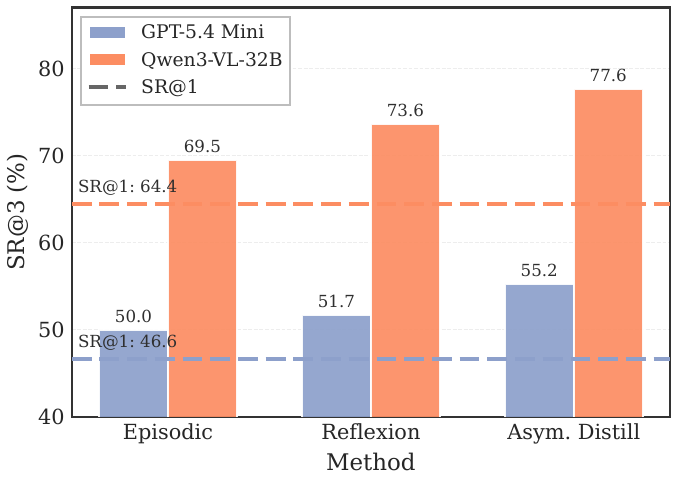}
        \caption{Cross-trial performance (SR@3)}
    \end{subfigure}
    \caption{\textbf{Strong-backbone generalization.} (a) Intra-trial performance (SR@1) and (b) cross-trial performance (SR@3) on AndroidWorld using GPT-5.4 Mini and Qwen3-VL-32B. All results are three-run averages on a fixed half subset.\protect\footnotemark}
    \label{fig:backbone_generalization}
    \vspace{-18pt}
\end{figure}
\footnotetext{For GPT-5.4 Mini evaluation, we follow UI-Ins~\cite{chen2025ui} by using GPT-5.4 Mini as the planner and Qwen3-VL-8B for grounding localization.}

\paragraph{Strong-Backbone Generalization.} Figure~\ref{fig:backbone_generalization} further evaluates GPT-5.4 Mini and Qwen3-VL-32B as strong backbones. AnchorGUI improves SR@1 over ReAct and Truncation for both, while asymmetric distillation yields the best SR@3, indicating that CSA-driven asymmetric memory remains beneficial for both stronger open-source and closed-source VLM settings.

\section{Conclusion}

In this paper, we propose \textbf{AnchorGUI} by identifying an empirical \emph{informational asymmetry}, where expected transitions often compress into text while unexpected mismatches benefit from original screenshots. At its core, the Cognitive State Anchor (CSA) compares pre-action expectations with post-action visual realities to generate explicit prediction-error signals. These signals drive asymmetric visual memory for dual-scale learning: \emph{intra-trial}, enabling immediate error correction while mitigating multimodal token explosion; \emph{cross-trial}, focusing the computationally expensive credit assignment search space on detected mismatches, allowing agents to efficiently distill insights from past failures. Extensive experiments show that AnchorGUI significantly boosts success rates and reduces token consumption, offering a principled and efficient framework for multimodal GUI agents to process and learn from dense visual histories.

\clearpage
\section*{Acknowledgements}
This work is partially supported by National Natural Science Foundation of China (62376274, 62437002). Zhiwu Lu is the corresponding author.

\bibliographystyle{splncs04}
\bibliography{main}

\begin{thebibliography}{10}
\providecommand{\url}[1]{\texttt{#1}}
\providecommand{\urlprefix}{URL }
\providecommand{\doi}[1]{https://doi.org/#1}

\bibitem{agashe2025agent}
Agashe, S., Wong, K., Tu, V., Yang, J., Li, A., Wang, X.E.: {Agent S2}: A
  compositional generalist-specialist framework for computer use agents. In:
  Conference on Language Modeling ({COLM}) (2025)

\bibitem{bai2025qwen3}
Bai, S., Cai, Y., Chen, R., Chen, K., Chen, X., Cheng, Z., Deng, L., Ding, W.,
  Gao, C., Ge, C., et~al.: {Qwen3-VL} technical report. arXiv preprint
  arXiv:2511.21631  (2025)

\bibitem{brown2020language}
Brown, T., Mann, B., Ryder, N., Subbiah, M., Kaplan, J.D., Dhariwal, P.,
  Neelakantan, A., Shyam, P., Sastry, G., Askell, A., Agarwal, S.,
  Herbert-Voss, A., Krueger, G., Henighan, T., Child, R., Ramesh, A., Ziegler,
  D., Wu, J., Winter, C., Hesse, C., Chen, M., Sigler, E., Litwin, M., Gray,
  S., Chess, B., Clark, J., Berner, C., McCandlish, S., Radford, A., Sutskever,
  I., Amodei, D.: Language models are few-shot learners. In: Advances in Neural
  Information Processing Systems ({NeurIPS}). vol.~33, pp. 1877--1901. Curran
  Associates, Inc. (2020)

\bibitem{chen2025ui}
Chen, L., Zhou, H., Cai, C., Zhang, J., Tong, P., Zhang, X., Kong, Q., Liu, C.,
  Liu, Y., Wang, W., Wang, Y., Jin, Q., Hoi, S.: {UI-Ins}: Enhancing {GUI}
  grounding with multi-perspective instruction as reasoning. In: International
  Conference on Learning Representations ({ICLR}) (2026)

\bibitem{gao2025chain}
Gao, X., Hu, C., Chen, B., Li, T.: {Chain-of-Memory}: Enhancing {GUI} agents
  for cross-application navigation. arXiv preprint arXiv:2506.18158  (2025)

\bibitem{team2024gemini}
{Gemini Team}, Georgiev, P., Lei, V.I., Burnell, R., Bai, L., Gulati, A.,
  Tanzer, G., Vincent, D., Pan, Z., Wang, S., et~al.: {Gemini} 1.5: Unlocking
  multimodal understanding across millions of tokens of context. arXiv preprint
  arXiv:2403.05530  (2024)

\bibitem{gou2024navigating}
Gou, B., Wang, D.R., Zheng, B., Xie, Y., Chang, C., Shu, Y., Sun, H., Su, Y.:
  Navigating the digital world as humans do: Universal visual grounding for
  {GUI} agents. In: International Conference on Learning Representations
  ({ICLR}). pp. 30851--30883 (2025)

\bibitem{hurst2024gpt}
Hurst, A., Lerer, A., Goucher, A.P., Perelman, A., Ramesh, A., Clark, A.,
  Ostrow, A., Welihinda, A., Hayes, A., Radford, A., et~al.: {GPT-4o} system
  card. arXiv preprint arXiv:2410.21276  (2024)

\bibitem{lai2025androidgen}
Lai, H., Gao, J., Liu, X., Xu, Y., Zhang, S., Dong, Y., Tang, J.: {AndroidGen}:
  Building an {Android} language agent under data scarcity. In: Annual Meeting
  of the Association for Computational Linguistics ({ACL}). pp. 2727--2749.
  Association for Computational Linguistics, Vienna, Austria (Jul 2025).
  \doi{10.18653/v1/2025.acl-long.138}

\bibitem{li2025mobileuse}
Li, N., Qu, X., Zhou, J., Wen, M., Du, K., Lou, X., Peng, Q., Wang, J., Zhang,
  W.: {MobileUse}: A hierarchical reflection-driven {GUI} agent for autonomous
  mobile operation. In: Advances in Neural Information Processing Systems
  ({NeurIPS}). vol.~38, pp. 40361--40388. Curran Associates, Inc. (2025)

\bibitem{li2024effects}
Li, W., Bishop, W., Li, A., Rawles, C., Campbell-Ajala, F., Tyamagundlu, D.,
  Riva, O.: On the effects of data scale on {UI} control agents. In: Advances
  in Neural Information Processing Systems ({NeurIPS}). vol.~37, pp.
  92130--92154. Curran Associates, Inc. (2024). \doi{10.52202/079017-2925}

\bibitem{liu2025llm}
Liu, G., Zhao, P., Liang, Y., Liu, L., Guo, Y., Xiao, H., Lin, W., Chai, Y.,
  Han, Y., Ren, S., Wang, H., Liang, X., Wang, W., Wu, T., Lu, Z., Chen, S.,
  LiLinghao, Wang, H., Xiong, G., Liu, Y., Li, H.: {{LLM}-Powered} {GUI} agents
  in phone automation: Surveying progress and prospects. Transactions on
  Machine Learning Research ({TMLR})  (2025)

\bibitem{liu2025infigui}
Liu, Y., Li, P., Xie, C., Hu, X., Han, X., Zhang, S., Yang, H., Wu, F.:
  {InfiGUI-R1}: Advancing multimodal {GUI} agents from reactive actors to
  deliberative reasoners. arXiv preprint arXiv:2504.14239  (2025)

\bibitem{liu2025pal}
Liu, Z., Li, J., Zhao, W.X., Gao, D., Li, Y., Wen, J.r.: {PAL-UI}: Planning
  with active look-back for vision-based {GUI} agents. arXiv preprint
  arXiv:2510.00413  (2025)

\bibitem{lu2025guiodyssey}
Lu, Q., Shao, W., Liu, Z., Du, L., Meng, F., Li, B., Chen, B., Huang, S.,
  Zhang, K., Luo, P.: {GUIOdyssey}: A comprehensive dataset for cross-app {GUI}
  navigation on mobile devices. In: IEEE/CVF International Conference on
  Computer Vision ({ICCV}). pp. 22404--22414. IEEE (Oct 2025).
  \doi{10.1109/ICCV51701.2025.02080}

\bibitem{madaan2023self}
Madaan, A., Tandon, N., Gupta, P., Hallinan, S., Gao, L., Wiegreffe, S., Alon,
  U., Dziri, N., Prabhumoye, S., Yang, Y., Gupta, S., Majumder, B.P., Hermann,
  K., Welleck, S., Yazdanbakhsh, A., Clark, P.: {Self-Refine}: Iterative
  refinement with self-feedback. In: Advances in Neural Information Processing
  Systems ({NeurIPS}). vol.~36, pp. 46534--46594. Curran Associates, Inc.
  (2023)

\bibitem{nguyen2025gui}
Nguyen, D., Chen, J., Wang, Y., Wu, G., Park, N., Hu, Z., Lyu, H., Wu, J.,
  Aponte, R., Xia, Y., Li, X., Shi, J., Chen, H., Lai, V.D., Xie, Z., Kim, S.,
  Zhang, R., Yu, T., Tanjim, M., Ahmed, N.K., Mathur, P., Yoon, S., Yao, L.,
  Kveton, B., Kil, J., Nguyen, T.H., Bui, T., Zhou, T., Rossi, R.A.,
  Dernoncourt, F.: {GUI} agents: A survey. In: Findings of the Association for
  Computational Linguistics ({ACL}). pp. 22522--22538. Association for
  Computational Linguistics, Vienna, Austria (Jul 2025).
  \doi{10.18653/v1/2025.findings-acl.1158}

\bibitem{qin2025ui}
Qin, Y., Ye, Y., Fang, J., Wang, H., Liang, S., Tian, S., Zhang, J., Li, J.,
  Li, Y., Huang, S., et~al.: {UI-TARS}: Pioneering automated {GUI} interaction
  with native agents. arXiv preprint arXiv:2501.12326  (2025)

\bibitem{rawles2024androidworld}
Rawles, C., Clinckemaillie, S., Chang, Y., Waltz, J., Lau, G., Fair, M., Li,
  A., Bishop, W., Li, W., Campbell-Ajala, F., Toyama, D., Berry, R.,
  Tyamagundlu, D., Lillicrap, T., Riva, O.: {AndroidWorld}: A dynamic
  benchmarking environment for autonomous agents. In: International Conference
  on Learning Representations ({ICLR}). pp. 406--441 (2025)

\bibitem{shinn2023reflexion}
Shinn, N., Cassano, F., Gopinath, A., Narasimhan, K., Yao, S.: {Reflexion}:
  Language agents with verbal reinforcement learning. In: Advances in Neural
  Information Processing Systems ({NeurIPS}). vol.~36, pp. 8634--8652. Curran
  Associates, Inc. (2023)

\bibitem{shridhar2020alfworld}
Shridhar, M., Yuan, X., C{\^o}t{\'e}, M.A., Bisk, Y., Trischler, A.,
  Hausknecht, M.: {ALFWorld}: Aligning text and embodied environments for
  interactive learning. In: International Conference on Learning
  Representations ({ICLR}) (2021)

\bibitem{sun2025genesis}
Sun, Q., Cheng, K., Ding, Z., Jin, C., Wang, Y., Xu, F., Wu, Z., Jia, C., Chen,
  L., Liu, Z., Kao, B., Li, G., He, J., Qiao, Y., Wu, Z.: {OS-Genesis}:
  Automating {GUI} agent trajectory construction via reverse task synthesis.
  In: Annual Meeting of the Association for Computational Linguistics ({ACL}).
  pp. 5555--5579. Association for Computational Linguistics, Vienna, Austria
  (Jul 2025). \doi{10.18653/v1/2025.acl-long.277}

\bibitem{sutton1998reinforcement}
Sutton, R.S., Barto, A.G.: Reinforcement learning: An introduction. MIT Press,
  Cambridge, MA (1998)

\bibitem{wang2023voyager}
Wang, G., Xie, Y., Jiang, Y., Mandlekar, A., Xiao, C., Zhu, Y., Fan, L.,
  Anandkumar, A.: {Voyager}: An open-ended embodied agent with {Large Language
  Models}. Transactions on Machine Learning Research ({TMLR})  (2024)

\bibitem{wang2024mobile}
Wang, J., Xu, H., Jia, H., Zhang, X., Yan, M., Shen, W., Zhang, J., Huang, F.,
  Sang, J.: {Mobile-Agent-v2}: Mobile device operation assistant with effective
  navigation via multi-agent collaboration. In: Advances in Neural Information
  Processing Systems ({NeurIPS}). vol.~37, pp. 2686--2710. Curran Associates,
  Inc. (2024). \doi{10.52202/079017-0088}

\bibitem{wang2023mobile}
Wang, J., Xu, H., Ye, J., Yan, M., Shen, W., Zhang, J., Huang, F., Sang, J.:
  {Mobile-Agent}: Autonomous multi-modal mobile device agent with visual
  perception. arXiv preprint arXiv:2401.16158  (2024)

\bibitem{wang2024gui}
Wang, S., Liu, W., Chen, J., Zhou, Y., Gan, W., Zeng, X., Che, Y., Yu, S., Hao,
  X., Shao, K., et~al.: {GUI} agents with {Foundation Models}: A comprehensive
  survey. arXiv preprint arXiv:2411.04890  (2024)

\bibitem{wanyan2025look}
Wanyan, Y., Zhang, X., Xu, H., Liu, H., Wang, J., Ye, J., Kou, Y., Yan, M.,
  Huang, F., Yang, X., Dong, W., Xu, C.: Look before you leap: A
  {GUI-Critic-R1} model for pre-operative error diagnosis in {GUI} automation.
  In: Advances in Neural Information Processing Systems ({NeurIPS}). vol.~38,
  pp. 3907--3929. Curran Associates, Inc. (2025)

\bibitem{wu2025auto}
Wu, W., Zhou, K., Yuan, R., Yu, V., Wang, S., Hu, Z., Huang, B.: Auto-scaling
  continuous memory for {GUI} agent. arXiv preprint arXiv:2510.09038  (2025)

\bibitem{wu2025resum}
Wu, X., Li, K., Zhao, Y., Zhang, L., Ou, L., Yin, H., Zhang, Z., Yu, X., Zhang,
  D., Jiang, Y., et~al.: {ReSum}: Unlocking long-horizon search intelligence
  via context summarization. arXiv preprint arXiv:2509.13313  (2025)

\bibitem{xiao2025ui}
Xiao, H., Wang, G., Chai, Y., Lu, Z., Lin, W., He, H., Fan, L., Bian, L., Hu,
  R., Liu, L., Ren, S., Wen, Y., Chen, X., Zhou, A., Li, H.: {UI-Genie}: A
  self-improving approach for iteratively boosting {MLLM}-based mobile {GUI}
  agents. In: Advances in Neural Information Processing Systems ({NeurIPS}).
  vol.~38, pp. 150376--150411. Curran Associates, Inc. (2025)

\bibitem{xu2025androidlab}
Xu, Y., Liu, X., Sun, X., Cheng, S., Yu, H., Lai, H., Zhang, S., Zhang, D.,
  Tang, J., Dong, Y.: {AndroidLab}: Training and systematic benchmarking of
  {Android} autonomous agents. In: Annual Meeting of the Association for
  Computational Linguistics ({ACL}). pp. 2144--2166. Association for
  Computational Linguistics, Vienna, Austria (Jul 2025).
  \doi{10.18653/v1/2025.acl-long.107}

\bibitem{xu2024aguvis}
Xu, Y., Wang, Z., Wang, J., Lu, D., Xie, T., Saha, A., Sahoo, D., Yu, T.,
  Xiong, C.: {Aguvis}: Unified pure vision agents for autonomous {GUI}
  interaction. In: International Conference on Machine Learning ({ICML}). pp.
  69772--69805. PMLR (2025)

\bibitem{yang2025aria}
Yang, Y., Wang, Y., Li, D., Luo, Z., Chen, B., Huang, C., Li, J.: {Aria-UI}:
  Visual grounding for {GUI} instructions. In: Findings of the Association for
  Computational Linguistics ({ACL}). pp. 22418--22433. Association for
  Computational Linguistics, Vienna, Austria (Jul 2025).
  \doi{10.18653/v1/2025.findings-acl.1152}

\bibitem{yao2022react}
Yao, S., Zhao, J., Yu, D., Du, N., Shafran, I., Narasimhan, K.R., Cao, Y.:
  {ReAct}: Synergizing reasoning and acting in language models. In:
  International Conference on Learning Representations ({ICLR}) (2023)

\bibitem{ye2025mobile}
Ye, J., Zhang, X., Xu, H., Liu, H., Wang, J., Zhu, Z., Zheng, Z., Gao, F., Cao,
  J., Lu, Z., et~al.: {Mobile-Agent-v3}: Fundamental agents for {GUI}
  automation. arXiv preprint arXiv:2508.15144  (2025)

\bibitem{zhang2024large}
Zhang, C., He, S., Qian, J., Li, B., Li, L., Qin, S., Kang, Y., Ma, M., Liu,
  G., Lin, Q., Rajmohan, S., Zhang, D., Zhang, Q.: {Large Language
  Model-Brained} {GUI} agents: A survey. Transactions on Machine Learning
  Research ({TMLR})  (2025)

\bibitem{zhang2025appagent}
Zhang, C., Yang, Z., Liu, J., Li, Y., Han, Y., Chen, X., Huang, Z., Fu, B., Yu,
  G.: {AppAgent}: Multimodal agents as smartphone users. In: {ACM} Conference
  on Human Factors in Computing Systems ({CHI}). pp. 1--20. ACM (Apr 2025).
  \doi{10.1145/3706598.3713600}

\bibitem{zhang2024android}
Zhang, J., Wu, J., Yihua, T., Liao, M., Xu, N., Xiao, X., Wei, Z., Tang, D.:
  {Android} in the zoo: {Chain-of-Action-Thought} for {GUI} agents. In:
  Findings of the Association for Computational Linguistics ({EMNLP}). pp.
  12016--12031. Association for Computational Linguistics, Miami, Florida, USA
  (Nov 2024). \doi{10.18653/v1/2024.findings-emnlp.702}

\bibitem{zhao2024gui}
Zhao, K., Song, J., Sha, L., Shen, H., Chen, Z., Zhao, T., Liang, X., Yin, J.:
  {GUI} testing arena: A unified benchmark for advancing autonomous {GUI}
  testing agent. arXiv preprint arXiv:2412.18426  (2024)

\end{thebibliography}
\clearpage
\appendix
\setcounter{figure}{0}
\setcounter{table}{0}
\setcounter{equation}{0}
\renewcommand{\thefigure}{A\arabic{figure}}
\renewcommand{\thetable}{A\arabic{table}}
\renewcommand{\theequation}{A\arabic{equation}}
\renewcommand{\theHfigure}{appendix.\arabic{figure}}
\renewcommand{\theHtable}{appendix.\arabic{table}}
\renewcommand{\theHequation}{appendix.\arabic{equation}}
\section{Supplementary Overview}
\label{sec:supp_overview}

We organize the supplementary material as a roadmap for quick reference.

\begin{tcolorbox}[
  breakable,
  colback=eccvblue!3,
  colframe=eccvblue!55!black,
  boxrule=0.5pt,
  arc=1.2mm,
  left=1.5mm,
  right=1.5mm,
  top=1mm,
  bottom=1mm
]
\small
\noindent\textbf{Implementation Details (Sec.~\ref{sec:supp_implementation}).}
Complete experimental settings, including hyperparameters, prompt templates, and practical details of the shared VLM backbone.

\medskip
\noindent\textbf{Algorithmic Details (Sec.~\ref{sec:supp_method}).}
Formal pseudocode for single-trial inference, failure-driven cross-trial distillation, and the overall multi-trial workflow.

\medskip
\noindent\textbf{Benchmark and Evaluation Protocols (Sec.~\ref{sec:supp_benchmark}).}
Benchmark definitions, action spaces, metrics, and evaluation procedures for both offline and online experiments.

\medskip
\noindent\textbf{Baseline Descriptions (Sec.~\ref{sec:supp_baselines}).}
Concise summaries of the compared baselines, with emphasis on their history-management and reflection mechanisms.

\medskip
\noindent\textbf{Evaluator Reliability (Sec.~\ref{sec:supp_reliability}).}
Manual annotation protocol and reliability analysis for the evaluator's binary verdicts.

\medskip
\noindent\textbf{Qualitative Case Studies (Sec.~\ref{sec:supp_cases}).}
Trajectory-level visualizations showing how AnchorGUI differs from standard accumulation and how cross-trial distillation improves retries.

\medskip
\noindent\textbf{Future Directions (Sec.~\ref{sec:supp_future}).}
Potential extensions beyond the current study.
\end{tcolorbox}

\section{Implementation Details}
\label{sec:supp_implementation}

\subsection{Hyperparameters}
\label{sec:supp_hyperparameters}

Unless otherwise stated, all main experiments use Qwen3-VL-8B as the shared backbone for the policy, evaluator, and distiller, with different roles instantiated only through different system prompts. Table~\ref{tab:supp_hyperparameters} summarizes the core hyperparameters used in our experiments.

We use an asymmetric sliding window of size $k=2$ for intra-trial history management, allow up to $K=3$ attempts in the cross-trial setting, and cap each AndroidWorld trial at $T=30$ steps. For decoding, we use stochastic sampling with temperature $=0.7$, top-$p=0.8$, top-$k=20$, and a repetition penalty of $1.0$. We report the mean performance over three random seeds.

\begin{table}[th!]
\centering
\caption{Core hyperparameter configuration.}
\label{tab:supp_hyperparameters}
\begin{tabular}{lc}
\toprule
\textbf{Hyperparameter} & \textbf{Value} \\
\midrule
Primary VLM backbone & Qwen3-VL-8B \\
Intra-trial sliding window size ($k$) & 2 \\
Maximum cross-trial attempts ($K$) & 3 \\
Maximum episode length ($T$) & 30 \\
Decoding temperature & 0.7 \\
Top-$p$ & 0.8 \\
Top-$k$ & 20 \\
Repetition penalty & 1.0 \\
Number of random seeds & 3 \\
\bottomrule
\end{tabular}
\end{table}

\subsection{Prompt Templates}
\label{sec:supp_prompts}

AnchorGUI uses one shared VLM with three role-specific prompts. The exact prompt instances differ slightly across benchmarks because the available action schema and environment metadata are not identical, but all experiments follow the same template structure described below. Runtime placeholders are filled with the current instruction, trajectory context, and benchmark-specific tool description.

\subsubsection{Policy Prompt}
\label{sec:supp_policy_prompt}

The policy prompt is responsible for next-action prediction together with an explicit expected transition. Its input contains six components: (1) the user instruction, (2) the compressed long-range history, (3) the role and tool specification, (4) the recent asymmetric window serialized as \texttt{\{\{ recent\_steps \}\}}, (5) the current screenshot, and (6) the cross-trial experience bank when available. The policy is required to output a valid environment action and to describe what it expects to happen after executing that action.

\begin{promptlisting}{Policy Prompt Template}
You are a helpful assistant.

# Tools

You may call one or more functions to assist with the user query.

You are provided with function signatures within <tools></tools> XML tags:
<tools>
{
  "type": "function",
  "function": {
    "name": "mobile_use",
    "description": "Use a touchscreen to interact with a mobile device, and take screenshots.\n* This is an interface to a mobile device with touchscreen. You can perform actions like clicking, typing, swiping, etc.\n* Some applications may take time to start or process actions, so you may need to wait and take successive screenshots to see the results of your actions.\n* The screen's resolution is {{ w }}x{{ h }}.\n* Make sure to click any buttons, links, icons, etc with the cursor tip in the center of the element. Don't click boxes on their edges unless asked.",
    "parameters": {
      "properties": {
        "action": {
          "description": "The action to perform. The available actions are:\n* `click`: Click the point on the screen with coordinate (x, y).\n* `long_press`: Press the point on the screen with coordinate (x, y) for specified seconds.\n* `swipe`: Swipe from the starting point with coordinate (x, y) to the end point with coordinates2 (x2, y2).\n* `type`: Input the specified text into the activated input box.\n* `answer`: Output the answer.\n* `system_button`: Press the system button.\n* `wait`: Wait specified seconds for the change to happen.\n* `terminate`: Terminate the current task and report its completion status.",
          "enum": [
            "click",
            "long_press",
            "swipe",
            "type",
            "answer",
            "system_button",
            "wait",
            "terminate"
          ],
          "type": "string"
        },
        "coordinate": {
          "description": "(x, y): The x (pixels from the left edge) and y (pixels from the top edge) coordinates to move the mouse to. Required only by `action=click`, `action=long_press`, and `action=swipe`.",
          "type": "array"
        },
        "coordinate2": {
          "description": "(x, y): The x (pixels from the left edge) and y (pixels from the top edge) coordinates to move the mouse to. Required only by `action=swipe`.",
          "type": "array"
        },
        "text": {
          "description": "Required only by `action=type` and `action=answer`.",
          "type": "string"
        },
        "time": {
          "description": "The seconds to wait. Required only by `action=long_press` and `action=wait`.",
          "type": "number"
        },
        "button": {
          "description": "Back means returning to the previous interface, Home means returning to the desktop, Menu means opening the application background menu, and Enter means pressing the enter. Required only by `action=system_button`",
          "enum": [
            "Back",
            "Home",
            "Menu",
            "Enter"
          ],
          "type": "string"
        },
        "status": {
          "description": "The status of the task. Required only by `action=terminate`.",
          "type": "string",
          "enum": [
            "success",
            "failure"
          ]
        }
      },
      "required": [
        "action"
      ],
      "type": "object"
    }
  }
}
</tools>

For each function call, return a json object with function name and arguments within <tool_call></tool_call> XML tags:
<tool_call>
{"name": <function-name>, "arguments": <args-json-object>}
</tool_call>

The user query: {{ instruction }}
{{ experience }}
Task progress (You have done the following operation on the current device): {{ history }}
Before answering, explain your reasoning step-by-step in <thinking></thinking> tags, and insert them before the <tool_call></tool_call> XML tags.
After answering, summarize your action in <conclusion></conclusion> tags, and insert them after the <tool_call></tool_call> XML tags.
{{ recent_steps }}
<image>
\end{promptlisting}

The placeholder \texttt{\{\{ recent\_steps \}\}} injects the short-horizon asymmetric memory used for intra-trial correction. Concretely, it serializes the most recent interaction records as an interleaved sequence of retained screenshots, historical policy outputs, and evaluator feedback. Steps with negative verdicts retain their original image as causal evidence, whereas expectation-consistent steps can be represented more compactly. This format allows the policy to recover local causal context directly from the prompt without reloading the entire trajectory. At a structural level, \texttt{\{\{ recent\_steps \}\}} is instantiated as a repeated sequence of visual evidence, policy record, and judge record, as shown below.

\begin{promptlisting}{Template of \texttt{\{\{ recent\_steps \}\}}}
<image_{t-k}>              % retained only if this step preserves visual evidence
<thinking>
<reasoning over state s_{t-k} and the intended next move>
</thinking>
<tool_call>
{"name": "mobile_use", "arguments": {<action_{t-k}>}}
</tool_call>
<conclusion>
<expected transition \hat{\Delta}_{t-k}>
</conclusion>
[JUDGE]
Verdict: <POSITIVE_or_NEGATIVE>
Summary: <observed transition \Delta_{t-k}>

...

<image_{t-1}>              % optional for expectation-consistent steps
<thinking>
<reasoning over state s_{t-1} and the intended next move>
</thinking>
<tool_call>
{"name": "mobile_use", "arguments": {<action_{t-1}>}}
</tool_call>
<conclusion>
<expected transition \hat{\Delta}_{t-1}>
</conclusion>
[JUDGE]
Verdict: <POSITIVE_or_NEGATIVE>
Summary: <observed transition \Delta_{t-1}>
\end{promptlisting}

In practice, the policy output is parsed into two fields: the executable action $a_t$ and the textual expected transition $\hat{\Delta}_t$. This design forces the model to externalize its short-term causal belief before observing the next screen, which is necessary for the subsequent evaluator stage.

\subsubsection{Evaluator Prompt}
\label{sec:supp_evaluator_prompt}

The evaluator prompt judges whether the executed action produced the intended GUI transition. It takes as input the pre-action screenshot, the post-action screenshot, and the expected transition produced by the policy. The evaluator then returns an observed transition description and a binary verdict indicating whether the expected and observed transitions are aligned.

\begin{promptlisting}{Evaluator Prompt Template}
You are a strict evaluator of GUI actions.
Your job: decide if the assistant's intent is achieved by the observed screen transition.

Output rules:
- Output EXACTLY two lines and nothing else.
- Line 1: "Verdict: POSITIVE" or "Verdict: NEGATIVE"
- Line 2: "Summary: <one sentence>" where the sentence combines intent + what actually changed.

Intent (from conclusion): {{ intent }}

Given the BEFORE and AFTER screenshots, decide whether the intent is achieved.
If the transition does not match the intent, Verdict must be NEGATIVE.

Remember: output EXACTLY two lines (Verdict + Summary).
\end{promptlisting}

As described in the main paper, the evaluator is the key mechanism that converts passive image sequences into Cognitive State Anchors. Its output forms the tuple $(\Delta_t, V_t)$, where $\Delta_t$ is the observed transition and $V_t\in\{0,1\}$ is the verdict used for asymmetric retention.

\subsubsection{Distiller Prompt}
\label{sec:supp_distiller_prompt}

The distiller prompt is only invoked after an unsuccessful trial. Its input is the episode-level asymmetric memory, which contains compact text summaries for expectation-consistent steps and preserves original screenshots only for expectation-mismatched steps. The distiller must diagnose the failure, avoid uninformative loops, and produce a concise experience bank that can be prepended to the next attempt.

\begin{promptlisting}{Distiller Prompt Template}
You are a GUI task reflection assistant specialized in failure diagnosis.

You must analyze failed attempts rigorously, identify root causes and error loops, and produce a complete corrected plan that avoids previously ineffective actions.

The user query: {{ instruction }}
Task progress from the previous failed attempt:
{{ progress_schema_note }}
{{ history }}

The task is NOT successfully completed because {{ failure_reason }}.
Now it is your turn to reflect on the past experience and come up with a new plan of action.

Requirements:
- Diagnosis must explicitly identify:
  1) root cause of failure,
  2) whether there was a repeated ineffective loop,
  3) where the strategy diverged from the task intent.
- Improved plan must cover the entire task path to completion, avoid ineffective steps from the previous attempt, and include decision points plus fallback branches when key actions fail.
- Avoid rules must be scenario-triggered and actionable. Write them as:
  "IF <situation where the current path is uncertain or repeatedly ineffective> THEN DO NOT <repeated bad pattern>; INSTEAD <targeted exploration or alternative solution attempt>".

Then output the final reflection strictly in this format:
<remark>
Diagnosis:
- Root cause: ...
- Loop pattern: ...
- Divergence from intent: ...
Improved plan:
1) Start from the initial page/state and specify the first required operation.
2) ...
3) ...
...
Avoid rules:
- ...
- ...
</remark>
\end{promptlisting}

The distiller output is stored as a short natural-language experience bank rather than a full free-form reflection transcript. This keeps the cross-trial memory compact and makes the retrieved guidance directly actionable in subsequent policy calls. Consistent with our asymmetric design, the distiller reasons over full visual evidence only for the subset of mismatched steps, which substantially reduces multimodal credit assignment complexity.

In our implementation, \texttt{\{\{progress\_schema\_note\}\}} is filled with the following fixed instruction:
\begin{promptlisting}{\texttt{progress\_schema\_note}}
Task-progress schema:
- screenshot_i (BEFORE action_i)
- JSON record for step i: {"Action": ..., "Intent": ..., "Judge": ...}
- screenshot_{i+1} (AFTER action_i)
Judge semantics: Verdict POSITIVE means the transition from BEFORE to AFTER
satisfies the intent; Verdict NEGATIVE means it does not. Prioritize NEGATIVE
verdicts and repeated ineffective loops.
\end{promptlisting}

We also use an explicit failure-reason string whose content depends on the failure mode. Concretely, \texttt{\{\{failure\_reason\}\}} is instantiated as one of the following:
\begin{promptlisting}{\texttt{failure\_reason}: Premature Completion}
the previous attempt incorrectly concluded task completion, but the task outcome was still incomplete or partially incorrect
\end{promptlisting}

\begin{promptlisting}{\texttt{failure\_reason}: Max-Step Termination}
the previous attempt reached the max-step limit without completion, likely due to ineffective repetition or a wrong action loop
\end{promptlisting}

Thus, our distiller explicitly distinguishes two unsuccessful-trial cases: incorrect self-termination despite an incomplete outcome, and forced termination after exhausting the step budget. We inject the corresponding failure-reason text to disambiguate these two situations during reflection.

\section{Algorithmic Details}
\label{sec:supp_method}

\subsection{Algorithmic Pseudocode}
\label{sec:supp_pseudocode}

For completeness, we provide formal algorithms for the single-trial inference routine, the failure-driven cross-trial distillation step, and the overall multi-trial inference workflow.

\begin{algorithm}[th!]
\caption{Single-Trial Inference}
\label{alg:single_trial}
\begin{algorithmic}[1]
\Require task instruction $g$, initial observation $o_1$, window size $k$, step budget $T$, optional experience bank $\mathcal{X}^{(k-1)}$
\State $\mathcal{B}_0 \gets \emptyset$
\State $\mathcal{H}_{asym}^{(k)} \gets \emptyset$
\For{$t = 1$ to $T$}
    \State $(a_t, \hat{z}_t) \gets \pi_\theta(g, o_t, \mathcal{B}_{t-1}, \mathcal{X}^{(k-1)})$
    \State execute $a_t$ in the environment and observe $o_{t+1}$
    \State $(z_t, V_t) \gets \phi_\theta(o_t, a_t, \hat{z}_t, o_{t+1})$
    \If{$V_t = 1$}
        \State $r_t \gets \mathrm{text}(o_t, a_t, \hat{z}_t, z_t)$
    \Else
        \State $r_t \gets (o_t, \mathrm{text}(a_t, \hat{z}_t, z_t))$
    \EndIf
    \State append $r_t$ to $\mathcal{H}_{asym}^{(k)}$
    \State $\mathcal{B}_t \gets \mathrm{Tail}_k(\mathcal{B}_{t-1} \cup \{r_t\})$
    \If{success checker returns true}
        \State \Return success, $\mathcal{H}_{asym}^{(k)}$
    \EndIf
    \If{$a_t$ is STOP or FINISH}
        \State \Return failure, $\mathcal{H}_{asym}^{(k)}$
    \EndIf
\EndFor
\State \Return failure, $\mathcal{H}_{asym}^{(k)}$
\end{algorithmic}
\end{algorithm}

\begin{algorithm}[th!]
\caption{Failure-Driven Cross-Trial Distillation}
\label{alg:cross_trial}
\begin{algorithmic}[1]
\Require failed trial memory $\mathcal{H}_{asym}^{(k)}$, failure reason $f^{(k)}$, previous experience bank $\mathcal{X}^{(k-1)}$
\State retain text summaries for all matched steps
\State retain original screenshots only for mismatched steps
\State $\mathcal{X}^{(k)} \gets \psi_\theta(\mathcal{H}_{asym}^{(k)}, f^{(k)}, \mathcal{X}^{(k-1)})$
\State convert $\mathcal{X}^{(k)}$ into concise reusable rules
\State remove redundant or overly trajectory-specific statements
\State \Return updated experience bank $\mathcal{X}^{(k)}$
\end{algorithmic}
\end{algorithm}

\begin{algorithm}[th!]
\caption{Overall Multi-Trial Inference Workflow}
\label{alg:full_workflow}
\begin{algorithmic}[1]
\Require task instruction $g$, maximum attempts $K$, step budget $T$, window size $k$
\State $\mathcal{X}^{(0)} \gets \emptyset$
\For{attempt $j = 1$ to $K$}
    \State reset environment to the task-specific initial state
    \State observe initial screenshot $o_1^{(j)}$
    \State status, $\mathcal{H}_{asym}^{(j)} \gets \Call{SingleTrialInference}{g, o_1^{(j)}, k, T, \mathcal{X}^{(j-1)}}$
    \If{status is success}
        \State \Return success at attempt $j$
    \EndIf
    \If{$j < K$}
        \State determine failure reason $f^{(j)}$
        \State $\mathcal{X}^{(j)} \gets \Call{FailureDrivenDistillation}{\mathcal{H}_{asym}^{(j)}, f^{(j)}, \mathcal{X}^{(j-1)}}$
    \EndIf
\EndFor
\State \Return failure after $K$ attempts
\end{algorithmic}
\end{algorithm}

\section{Benchmark and Evaluation Protocol Details}
\label{sec:supp_benchmark}

\paragraph{Unified offline evaluation protocol.}
For the three offline benchmarks (AITZ, Android-Control-High, and GUI-Odyssey), we adopt a unified step-level evaluation to avoid benchmark-specific reporting discrepancies. Let $\hat{a}_t=(\hat{c}_t,\hat{u}_t)$ and $a_t=(c_t,u_t)$ denote the predicted and gold action at step $t$, where $c_t$ is the action category and $u_t$ contains its arguments. We report
\begin{equation}
\mathrm{TM}=\mathbb{I}[\hat{c}_t=c_t], \qquad
\mathrm{EM}=\mathbb{I}[(\hat{c}_t,\hat{u}_t)\sim_d(c_t,u_t)],
\end{equation}
where $\sim_d$ denotes the dataset-specific full-match rule. TM evaluates category correctness only, while EM additionally checks whether the action arguments satisfy the benchmark's official matching criterion. Concretely, point-based actions must hit the correct target region or UI element, directional actions must match the gold direction, text actions must match the gold text (or satisfy the dataset's official text-similarity rule), and system / termination actions must match exactly. For all offline datasets, we use the released test split and evaluate all compared methods on the same examples without method-specific re-sampling.

\paragraph{AITZ.}
AITZ is a single-app Android GUI benchmark. We evaluate on its released test split (506 episodes / 4,724 screens). The action space contains five categories:
\texttt{CLICK},
\texttt{SCROLL},
\texttt{TYPE},
\texttt{PRESS}, and
\texttt{STOP}.
Click positions use normalized coordinates in $[0,1]\times[0,1]$. TM checks action-category agreement. EM follows the official screen-wise matching rule and additionally requires the predicted action arguments to match the gold action.

\paragraph{Android-Control-High.}
For Android-Control-High, we use the official processed high-level task and evaluate on the released high-level test split (1,543 episodes / 7,897 steps). The action schema includes four interaction actions:
\texttt{click},
\texttt{long\_press},
\texttt{input\_text}, and
\texttt{scroll};
and four navigation / control actions:
\texttt{navigate\_home},
\texttt{navigate\_back},
\texttt{open\_app}, and
\texttt{wait}.
Element-based actions are grounded by pixel coordinates. TM measures action-type correctness after normalizing predictions into the benchmark schema. EM follows the official relaxed step-wise rule: for element-based actions, the predicted point must lie inside the gold target element box; \texttt{navigate\_back} may also match an on-screen Back button; and \texttt{open\_app} may match a UI element whose text matches the app name. We use the benchmark's official processed examples and filtering rules.

\paragraph{GUI-Odyssey.}
GUI-Odyssey is a cross-app mobile GUI benchmark. We follow its released evaluation protocol. The action space includes
\texttt{CLICK},
\texttt{LONG PRESS},
\texttt{SCROLL},
\texttt{TYPE},
\texttt{COMPLETE},
\texttt{IMPOSSIBLE},
\texttt{HOME},
\texttt{BACK}, and
\texttt{RECENT}.
TM checks action-category agreement. EM follows the benchmark's official action-matching rule (reported as AMS): point-based actions must hit the official target region, including the provided masks when applicable; \texttt{SCROLL} must match direction; \texttt{TYPE} must satisfy the official text-similarity rule; and system / completion actions must match exactly.

\paragraph{AndroidWorld.}
AndroidWorld is our online benchmark for cross-trial evaluation. It contains 116 hand-crafted tasks spanning 20 real-world Android apps, and each task is executed in a live emulator rather than evaluated from a static screenshot--action corpus. We follow the official AndroidWorld environment and its task definitions. In particular, each task comes with dedicated \emph{initialization}, \emph{success-checking}, and \emph{tear-down} procedures that manipulate and inspect the emulator state. A trial is counted as successful only if the official task-specific success checker returns success before the trial terminates; otherwise, including premature termination or exhausting the step budget, the trial is counted as failure.

For reporting, let $y_i^{(k)}\in\{0,1\}$ indicate whether task $i$ is solved on attempt $k$. We define
\begin{equation}
\mathrm{SR}=\mathrm{SR}@1=\frac{1}{N}\sum_{i=1}^{N} y_i^{(1)},
\end{equation}
and
\begin{equation}
\mathrm{SR}@K=\frac{1}{N}\sum_{i=1}^{N}\mathbb{I}\!\left[\max_{1\leq j\leq K} y_i^{(j)}=1\right].
\end{equation}
Thus, $\mathrm{SR}@K$ measures whether a task is solved within the first $K$ attempts.

AndroidWorld supports randomized task instantiation, which is important for benchmark diversity but can confound cross-trial comparison if different retries observe different starting states. To isolate the effect of experience accumulation from environment randomness, each retry of the same evaluation item is reset through the benchmark's official initialization and tear-down procedures, so all attempts start from the same task-specific initial state rather than a newly sampled one. Put differently, randomness is introduced across benchmark items, but not across retries of the same item. This protocol ensures that gains in $\mathrm{SR}@K$ reflect improved cross-trial learning instead of simply benefiting from easier task re-instantiations on later attempts.

\section{Detailed Baseline Descriptions}
\label{sec:supp_baselines}

\subsection{Intra-Trial Baselines}
\label{sec:supp_intra_baselines}

\subsubsection{ReAct}
\label{sec:supp_baseline_react}
ReAct interleaves explicit reasoning and action execution in a closed loop. At each step, the model outputs a thought and a GUI action from the current screenshot, then appends the returned observation together with the preceding thought and action to history. The resulting \emph{Thought--Action--Observation} trace is kept in full and reused for later decisions.

\subsubsection{ReSum}
\label{sec:supp_baseline_resum}
ReSum replaces raw trajectory accumulation with incremental textual summarization. Each step is compressed into a short \emph{action summary}, and subsequent decisions condition on the running list of summaries rather than the original multimodal history.

\subsubsection{Truncation}
\label{sec:supp_baseline_truncation}
Truncation addresses context growth with a fixed recency window. It keeps only the most recent $k$ steps in the prompt and discards all earlier interactions without summarization. This strategy minimizes prompt length and preserves the exact format of recent multimodal context, but sacrifices long-range historical information once the window is exceeded.

\subsubsection{Self-Reflection}
\label{sec:supp_baseline_self_reflection}
Self-Reflection augments ReAct with immediate post-step reflection. After each action, the model writes a short reflective comment about whether the action worked, why it succeeded or failed, and what should be tried next; this reflection is then appended to history before the next decision.

\subsubsection{M3A}
\label{sec:supp_baseline_m3a}
M3A is a zero-shot mobile agent built around an \emph{observe--decide--act--reflect} loop. Each step conditions on the user instruction, the raw screenshot, detected UI elements, and a Set-of-Mark (SoM) rendering with indexed boxes. The model outputs a rationale and a structured action, then performs a post-action self-assessment from the before/after GUI states and appends a concise summary to memory.

\subsubsection{Mobile-Agent-v3}
\label{sec:supp_baseline_mobile_agent_v3}
Mobile-Agent-v3 is a multi-agent GUI automation framework that decomposes control into specialized roles. A \emph{Manager} converts the user instruction into an ordered list of sub-goals and dynamically updates this plan during execution. A \emph{Worker} selects the most actionable current sub-goal according to the present GUI state, prior feedback, and accumulated notes, then outputs the next operation. A \emph{Reflector} compares the Worker's intention with the observed state transition, judges success or failure, and returns diagnostic feedback to support replanning. When useful information appears on screen, a \emph{Notetaker} stores it as persistent notes for later reuse.

\subsubsection{Chain-of-Memory}
\label{sec:supp_baseline_chain_of_memory}
Chain-of-Memory (CoM) is an explicit memory framework for long-horizon, cross-app GUI navigation. After each action, it first performs information perception by comparing the recent action history and the screen change between consecutive observations, converting the transition into a textual \emph{action result}. Recent action-result pairs are stored in an ordered short-term memory (STM) to track the current task state. In parallel, the agent extracts task-relevant screen information from the current GUI conditioned on the instruction and STM, distills potentially reusable facts, and updates a long-term memory (LTM) that carries forward useful information across later steps.

\subsection{Cross-Trial Baselines}
\label{sec:supp_cross_baselines}

\subsubsection{Episodic Memory}
\label{sec:supp_baseline_episodic_memory}
Episodic Memory directly reuses the complete trajectory from the previous failed attempt. After trial $k$, the full sequence of observed screens, actions, and outcomes is serialized and prepended to the context of trial $k{+}1$.

\subsubsection{Reflexion}
\label{sec:supp_baseline_reflexion}
Reflexion performs cross-trial learning through verbal self-reflection. After a failed trial, the model reviews the complete trajectory, writes a compact textual reflection about what went wrong and what should change, and injects this reflection into the next trial as guidance.

\subsubsection{Memoryless Retry}
\label{sec:supp_baseline_memoryless_retry}
Memoryless Retry is a retry-only control baseline with no cross-trial transfer. After a failed trial, the environment is reset and the agent starts again from scratch, without carrying over any trajectory, summary, reflection, or memory from earlier attempts.

\section{Evaluator Reliability and Human Annotation Details}
\label{sec:supp_reliability}

\subsection{Sampling Protocol and Annotation Task}
\label{sec:supp_reliability_protocol}

To verify whether the evaluator can serve as a reliable source of binary verdicts for asymmetric retention, we conducted a manual audit of 535 evaluator decisions. Each audited item corresponds to one interaction step and was \textbf{randomly sampled} from the full pool of evaluator invocations collected during our experimental runs. The sampling unit is therefore an individual post-action verification event, rather than an entire episode.

For each sampled step, the annotator was presented with four fields: (1) the pre-action screenshot, (2) the executed action together with the policy's expected transition, (3) the post-action screenshot, and (4) the evaluator's predicted verdict. The human task was to determine whether the \emph{actual} GUI transition was semantically consistent with the policy's stated expectation.

We define the two labels as follows:
\begin{itemize}[leftmargin=1.5em]
    \item \textbf{Positive class} ($V_t=1$): the expected transition is achieved, \ie, the post-action GUI state is semantically aligned with what the policy intended.
    \item \textbf{Negative class} ($V_t=0$): the expected transition is not achieved, including wrong pages, no-op transitions, unexpected pop-ups, failed text entry, or partial changes that do not satisfy the intended sub-goal.
\end{itemize}

\paragraph{Annotation Instructions.}
The human annotation document followed the operational rules below.
\begin{itemize}[leftmargin=1.5em]
    \item Label a step as \emph{consistent} only when the main intended UI effect is clearly realized in the post-action screenshot.
    \item Ignore purely cosmetic differences (such as small animations or harmless layout shifts) unless they change task semantics.
    \item Label a step as \emph{mismatch} if the action lands on an unintended page, triggers an unexpected dialog, leaves the interface effectively unchanged, or only partially completes the intended transition.
    \item When uncertainty exists, prioritize task semantics over pixel-level similarity: the question is whether the action outcome matches the intended state transition, not whether the two screenshots look visually similar.
\end{itemize}

\subsection{Label Distribution and Reliability Metrics}
\label{sec:supp_reliability_metrics}

Table~\ref{tab:verdict_distribution} reports the ground-truth class distribution. The audited set is moderately imbalanced, with 406 positive samples (expectation-consistent transitions) and 129 negative samples (mismatched transitions). This mirrors the real usage pattern of the evaluator, where most routine interaction steps are expected to succeed and only a minority correspond to true mismatches.

\begin{table}[t]
    \centering
    \caption{Ground-truth label distribution in the 535 randomly sampled audited steps.}
    \label{tab:verdict_distribution}
    \begin{tabular}{lcc}
        \toprule
        Class & Count & Ratio \\
        \midrule
        Positive ($V_t=1$, expectation-consistent) & 406 & 75.89\% \\
        Negative ($V_t=0$, mismatch) & 129 & 24.11\% \\
        Total & 535 & 100.00\% \\
        \bottomrule
    \end{tabular}
\end{table}

Table~\ref{tab:verdict_confusion} shows the confusion matrix using the positive class definition above. The evaluator attains an F1 score of 96.6\%, complementing the precision and recall values reported in the main paper.

\begin{table}[t]
    \centering
    \caption{Confusion matrix and derived reliability metrics for the evaluator. Positive means $V_t=1$ (expectation-consistent); negative means $V_t=0$ (mismatch).}
    \label{tab:verdict_confusion}
    \begin{tabular}{lc}
        \toprule
        Quantity & Value \\
        \midrule
        True Positive (TP) & 382 \\
        True Negative (TN) & 126 \\
        False Positive (FP) & 3 \\
        False Negative (FN) & 24 \\
        Precision & 99.22\% \\
        Recall & 94.09\% \\
        F1 & 96.59\% \\
        \bottomrule
    \end{tabular}
\end{table}

These numbers explain the system behavior discussed in the main paper. Because the evaluator almost never produces false positives (only 3 out of 535 audited steps), it rarely marks a truly mismatched step as expectation-consistent; therefore, failure evidence is seldom discarded by mistake. The remaining errors are dominated by false negatives, which merely keep a small number of redundant successful frames and thus introduce minor token overhead without corrupting causal evidence.

\subsection{Annotation Platform Interface}
\label{sec:supp_reliability_platform}

\begin{figure*}[th!]
    \centering
    \setlength{\fboxsep}{0pt}
    \setlength{\tabcolsep}{6pt}
    \renewcommand{\arraystretch}{0}
    \begin{tabular}{cc}
        \fbox{\includegraphics[width=0.46\textwidth]{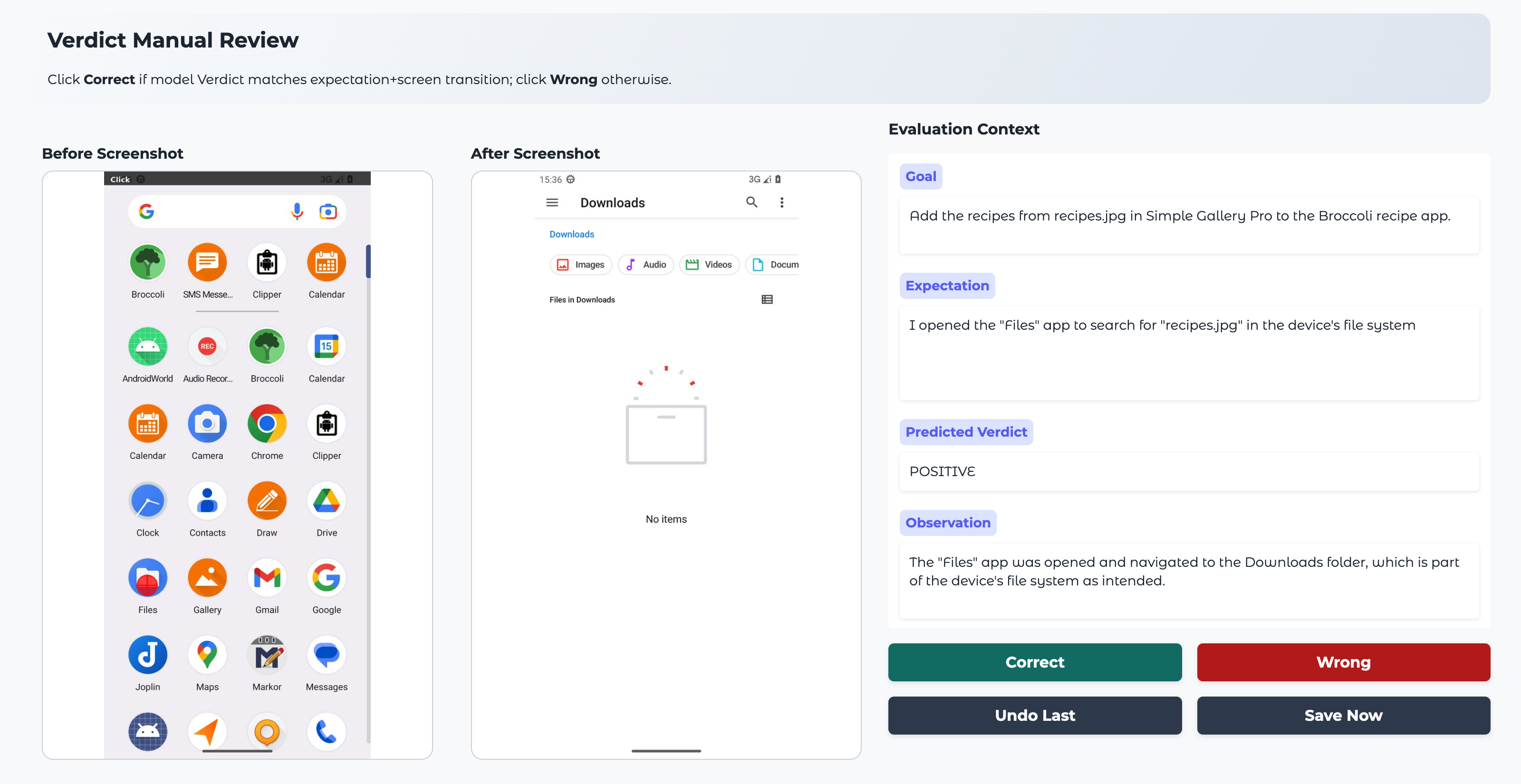}} &
        \fbox{\includegraphics[width=0.46\textwidth]{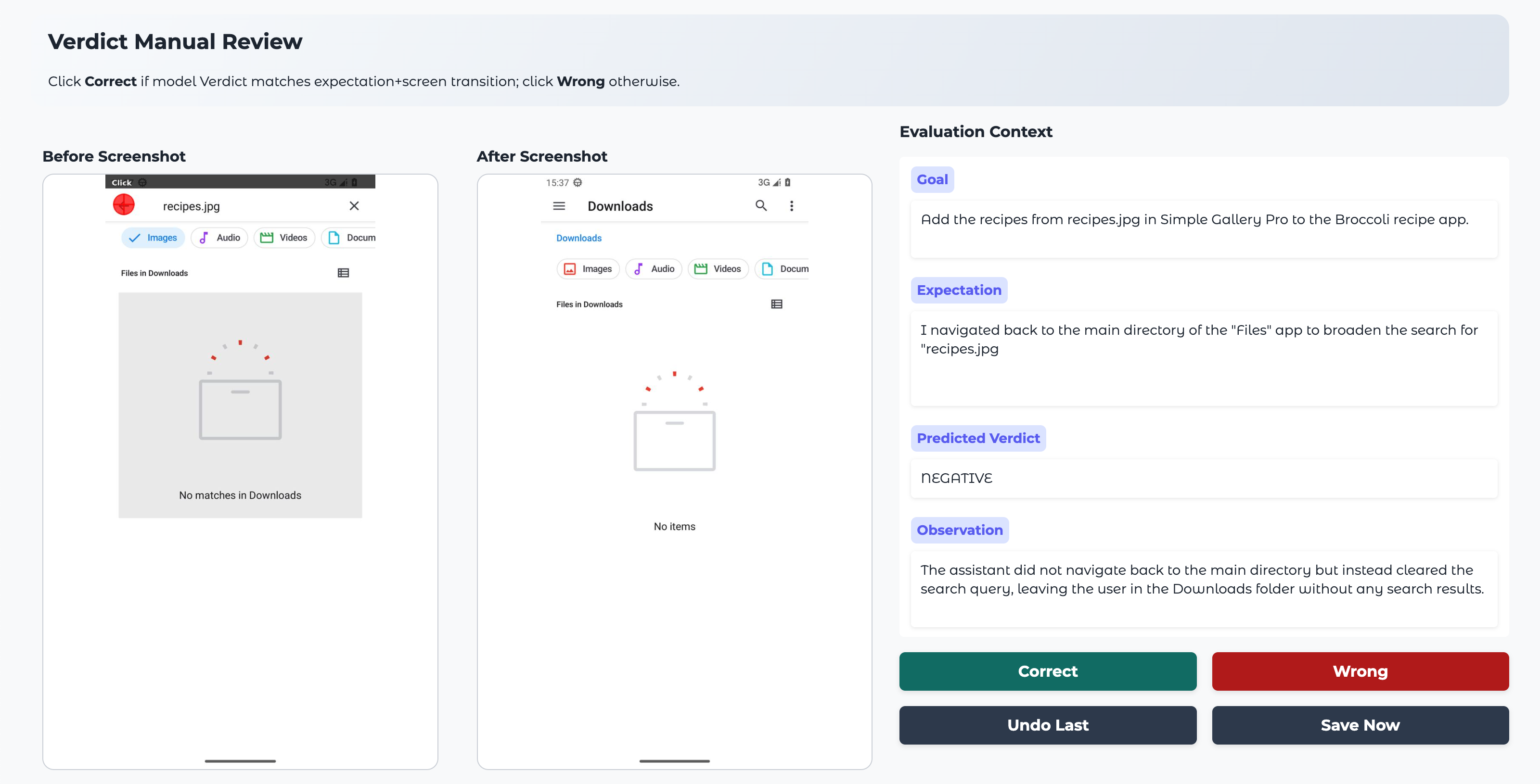}} \\
        \fbox{\includegraphics[width=0.46\textwidth]{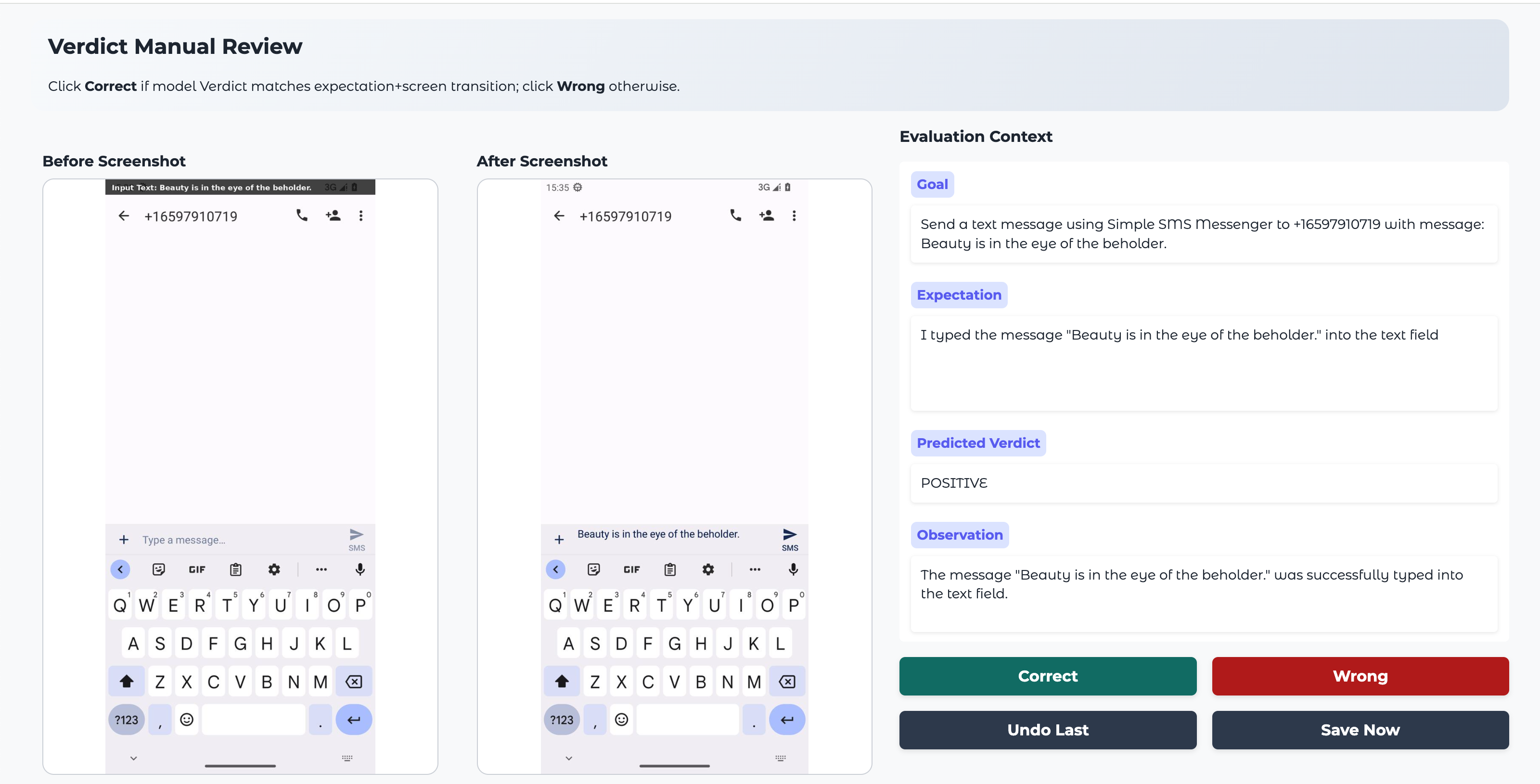}} &
        \fbox{\includegraphics[width=0.46\textwidth]{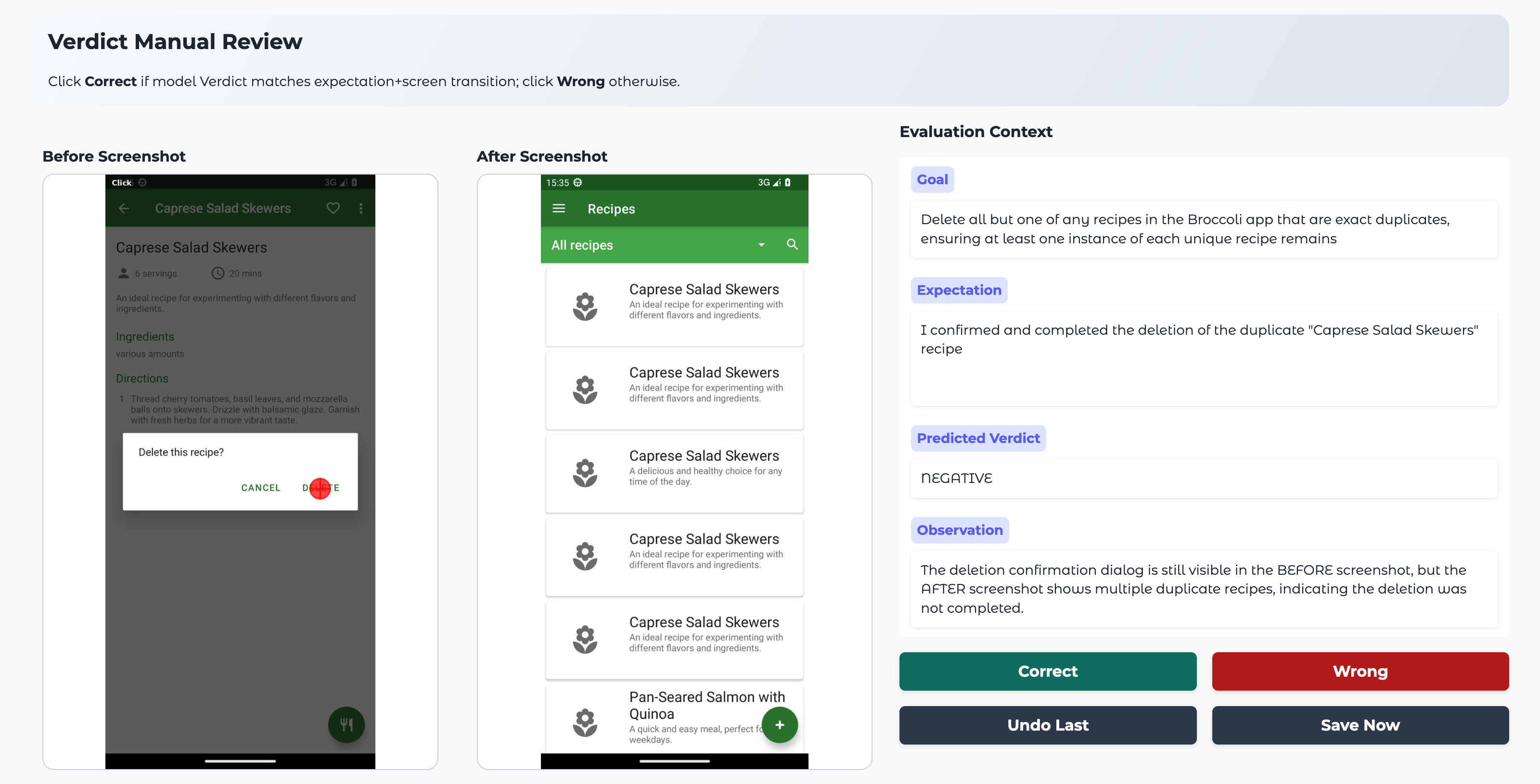}} \\
    \end{tabular}
    \caption{Screenshots of our custom annotation platform for evaluator reliability review. We show four example pages, each framed for clearer visual separation. For each sampled interaction step, the interface displays the pre-action screenshot, executed action, expected transition, post-action screenshot, evaluator verdict, and manual annotation panel.}
    \label{fig:verdict_manual_review}
\end{figure*}

Figure~\ref{fig:verdict_manual_review} shows the custom web-based annotation interface used for manual review. The platform presents the pre-action state, post-action state, model-produced expectation, evaluator verdict, and the annotator's final decision in a single page, enabling fast step-level verification with minimal context switching.

\section{Qualitative Case Studies}
\label{sec:supp_cases}

\subsection{AnchorGUI vs.\ ReAct}
\label{sec:supp_case_browsermaze}

We visualize synchronized trajectories on the same BrowserMaze instance to contrast how the two agents react after the first unexpected transition. For ReAct, we retain the first three executed actions, collapse the repetitive middle loop, and show the final stuck state. This condensation is faithful to the original rollout: after the third issued action fails to open \texttt{task.html}, ReAct keeps repeating the same click from Step~4 onward until the maximum horizon is exhausted. For AnchorGUI, we show the full trajectory because the policy changes substantially after each mismatch, and each panel includes the executed action together with the CSA verdict recorded after execution.

\begin{figure*}[ph!]
    \centering
    \small
    \casepanel{Step 1}{
        \caseimage{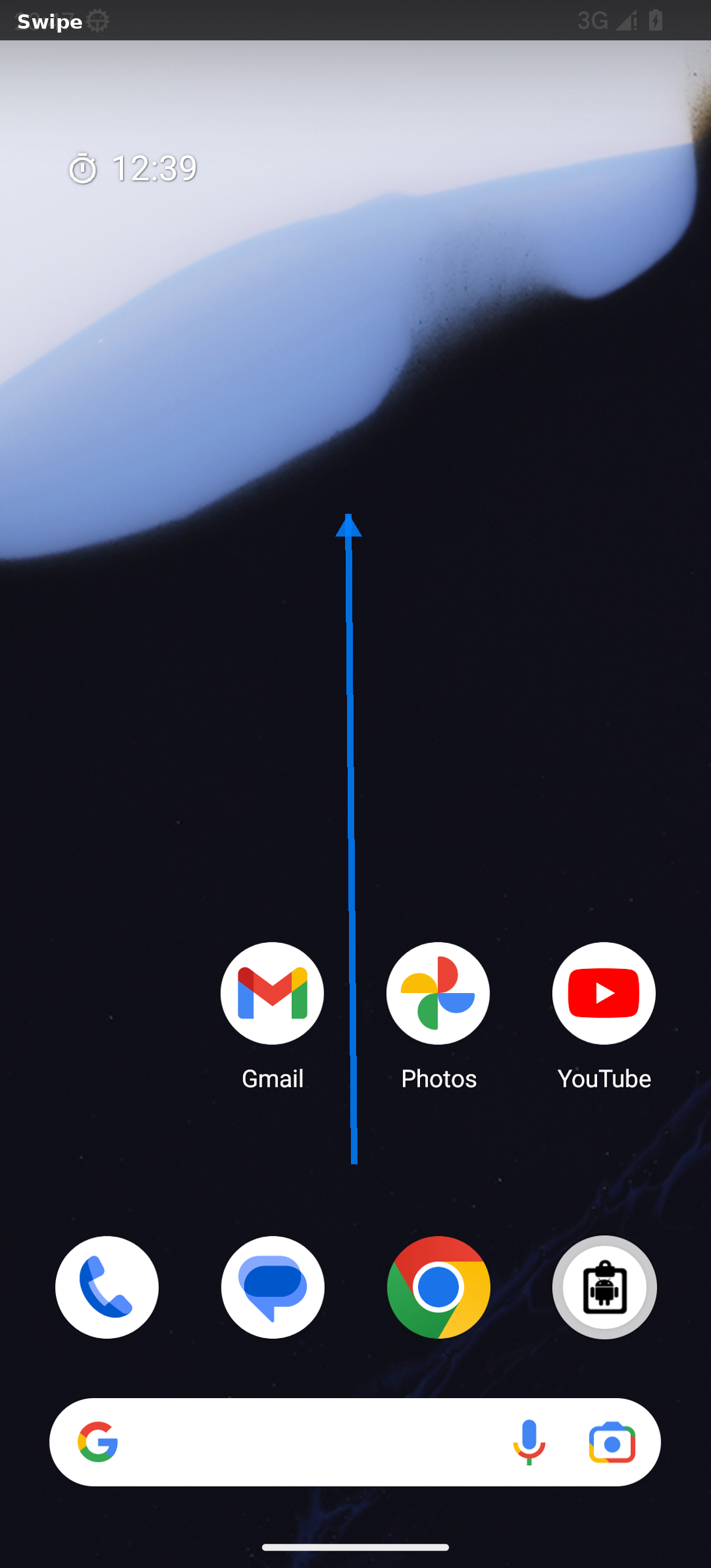}
    }{
        \caseinfo{0.17\textwidth}{
            \casejson{{"action": "swipe", "coordinate": [499, 742], "coordinate2": [491, 328]}}
        }
    }\hfill
    \casepanel{Step 2}{
        \caseimage{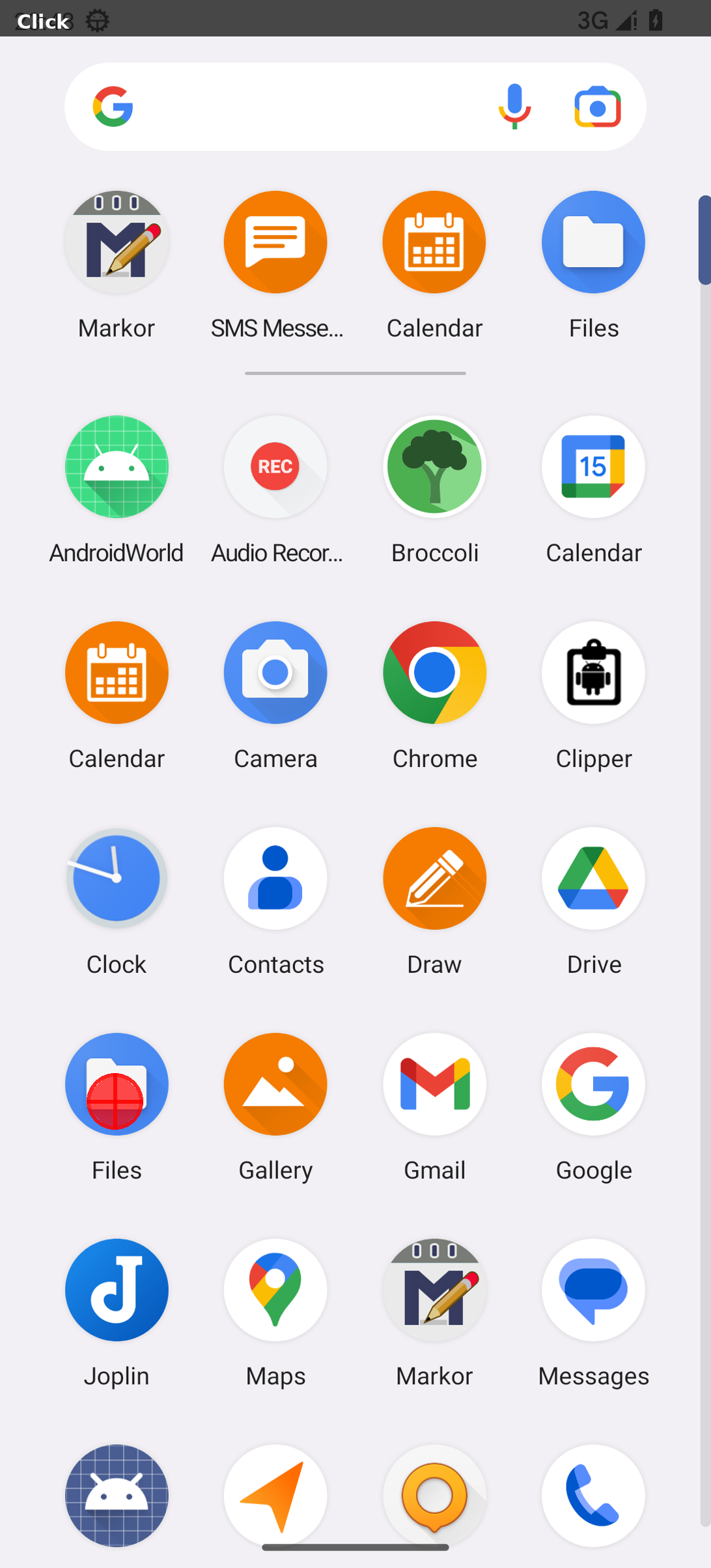}
    }{
        \caseinfo{0.17\textwidth}{
            \casejson{{"action": "click", "coordinate": [162, 702]}}
        }
    }\hfill
    \casepanel{Step 3}{
        \caseimage{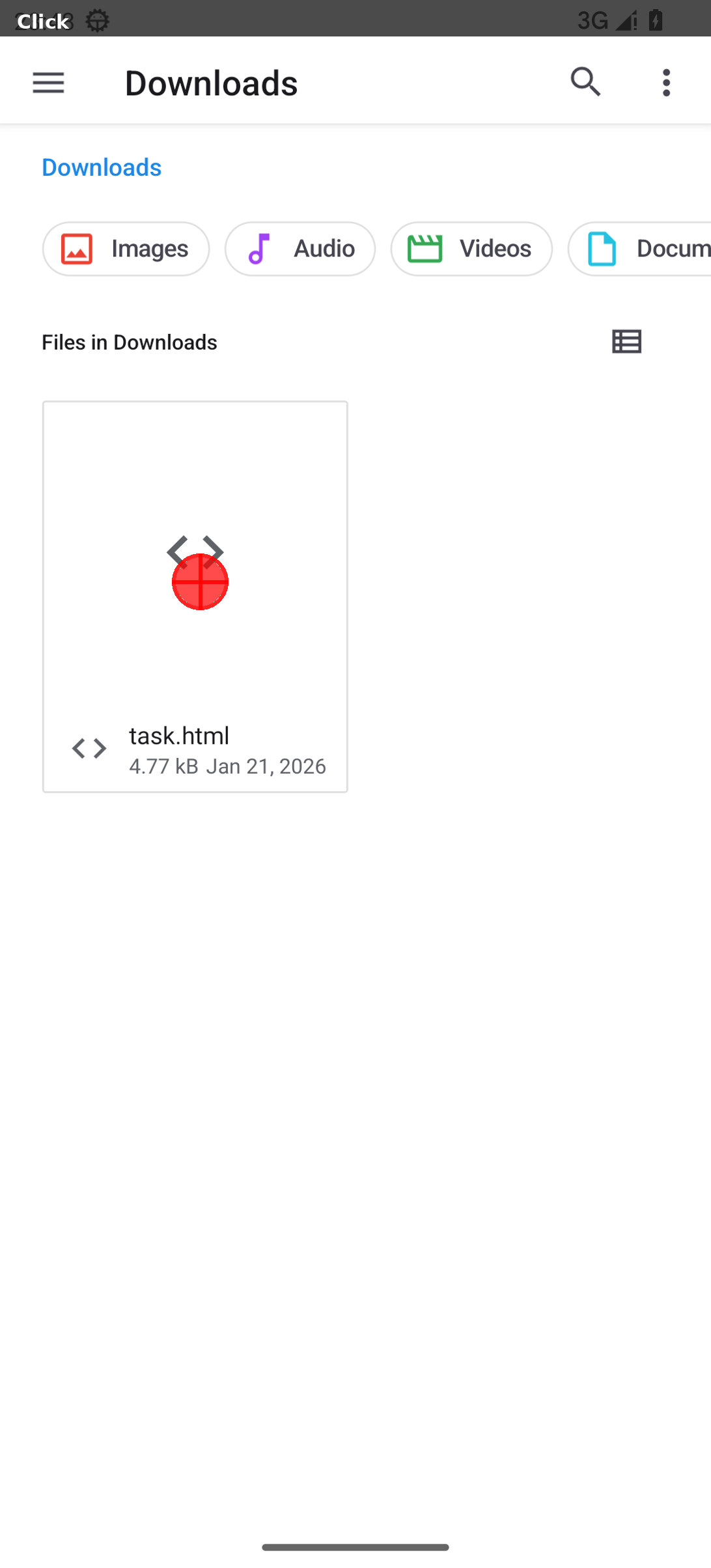}
    }{
        \caseinfo{0.17\textwidth}{
            \casejson{{"action": "click", "coordinate": [282, 371]}}
        }
    }\hfill
    \casepanel{Steps 4--19}{
        \caseimage{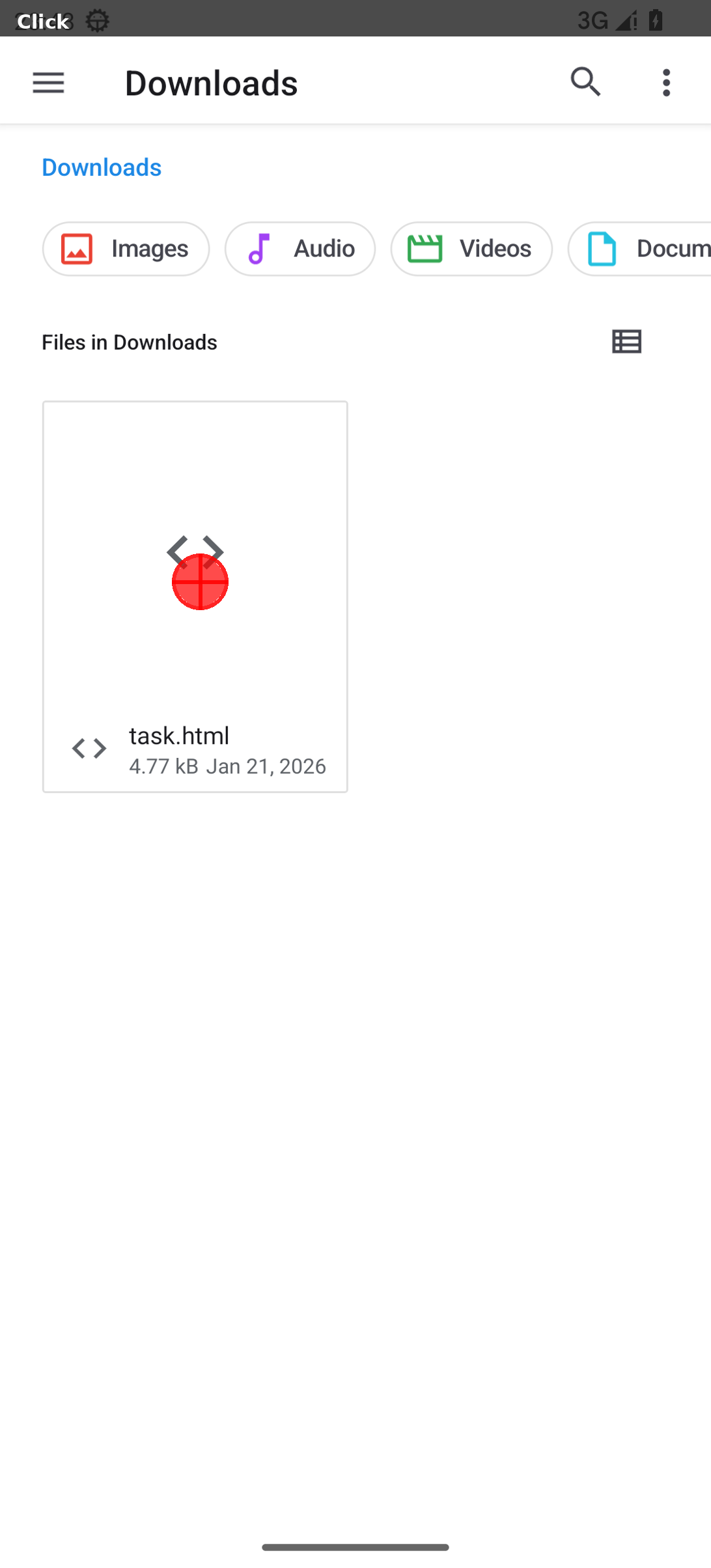}
    }{
        \caseinfo{0.17\textwidth}{
            \casejson{{"action": "click", "coordinate": [282, 371]}}
        }
    }\hfill
    \casepanel{Step 20}{
        \caseimage{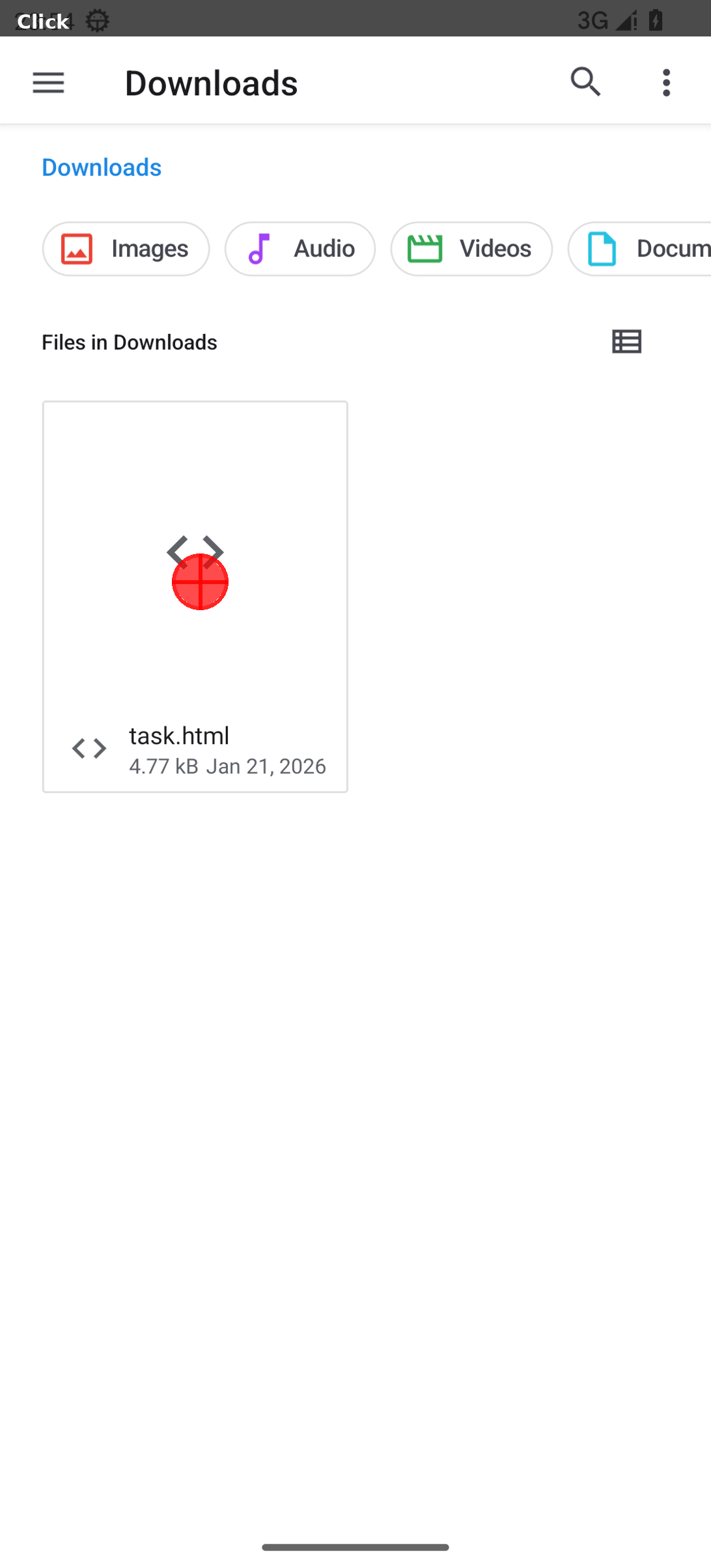}
    }{
        \caseinfo{0.17\textwidth}{
            \casejson{{"action": "click", "coordinate": [282, 371]}}
        }
    }
    \caption{\textbf{ReAct trajectory on BrowserMaze.} We keep the first three actions, collapse the repeated middle loop, and show the final stuck state. After the third issued action fails to open \texttt{task.html}, ReAct never revises its hypothesis and keeps clicking the same coordinates from Step~4 until the maximum horizon is exhausted.}
    \label{fig:case_react_browsermaze}
\end{figure*}

\begin{figure*}[ph!]
    \centering
    \small
    \casepanel{Step 1}{
        \caseimage{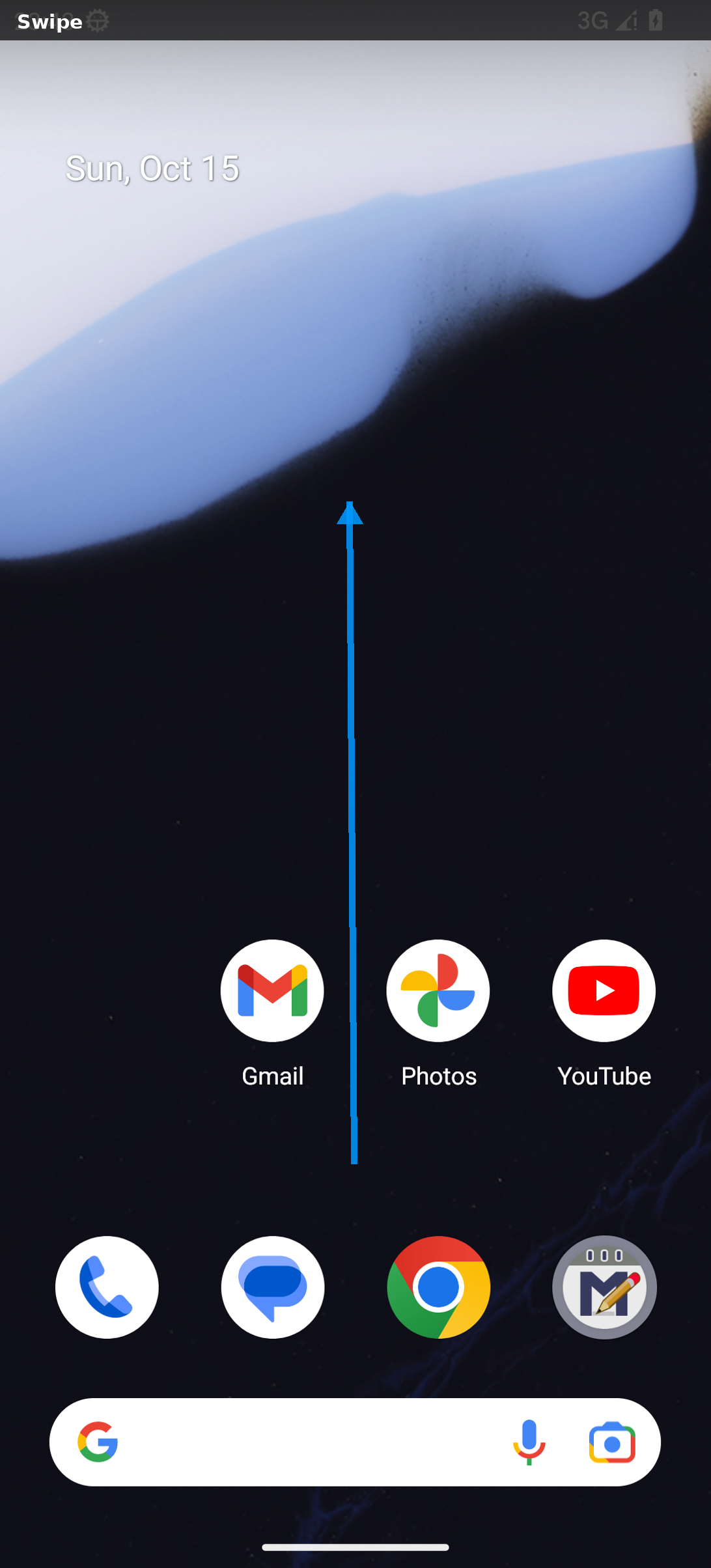}
    }{
        \caseinfo{0.235\textwidth}{
            \casejson{{"action": "swipe", "coordinate": [499, 742], "coordinate2": [493, 320]}}
            \casefield{Verdict}{\casepositive}
        }
    }\hfill
    \casepanel{Step 2}{
        \caseimage{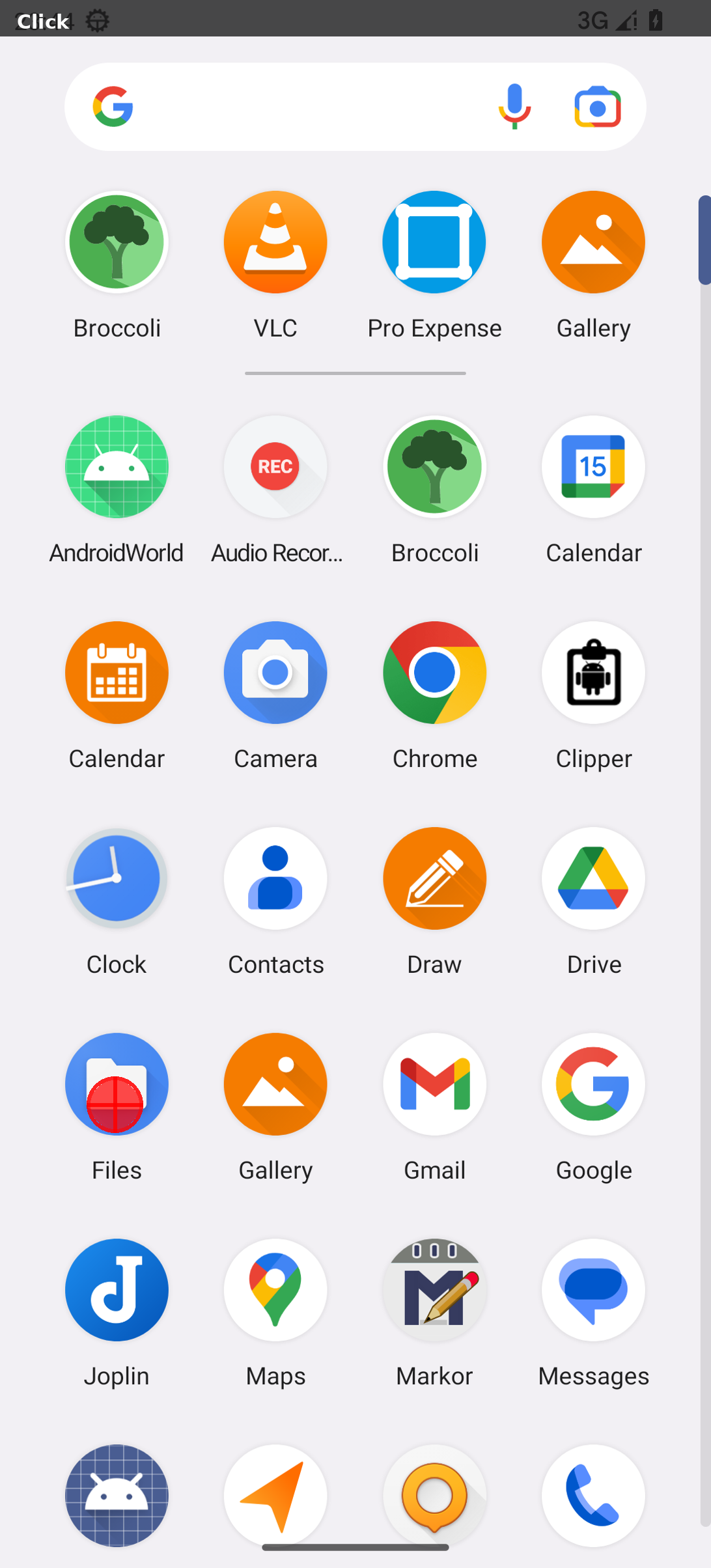}
    }{
        \caseinfo{0.235\textwidth}{
            \casejson{{"action": "click", "coordinate": [162, 704]}}
            \casefield{Verdict}{\casepositive}
        }
    }\hfill
    \casepanel{Step 3}{
        \caseimage{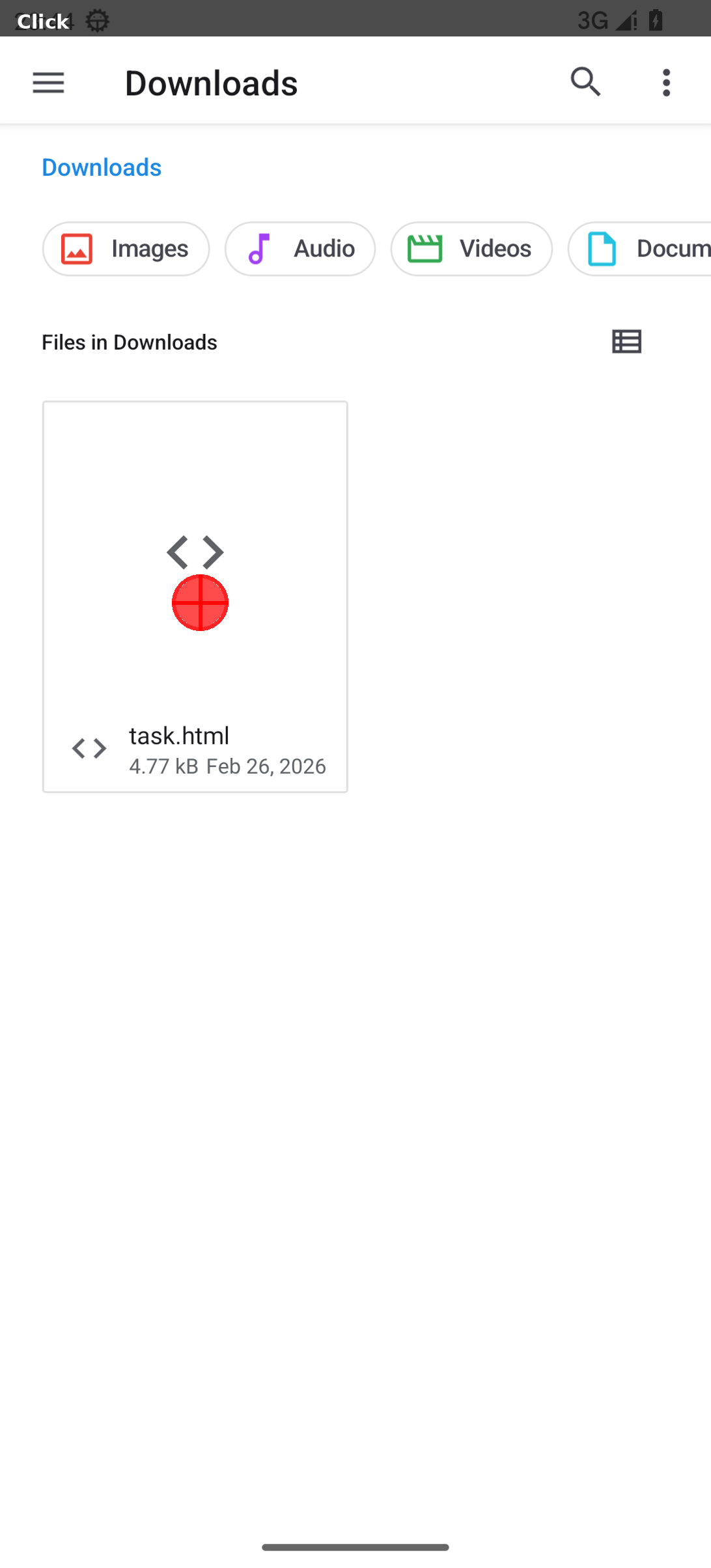}
    }{
        \caseinfo{0.235\textwidth}{
            \casejson{{"action": "click", "coordinate": [282, 384]}}
            \casefield{Verdict}{\casenegative}
        }
    }\hfill
    \casepanel{Step 4}{
        \caseimage{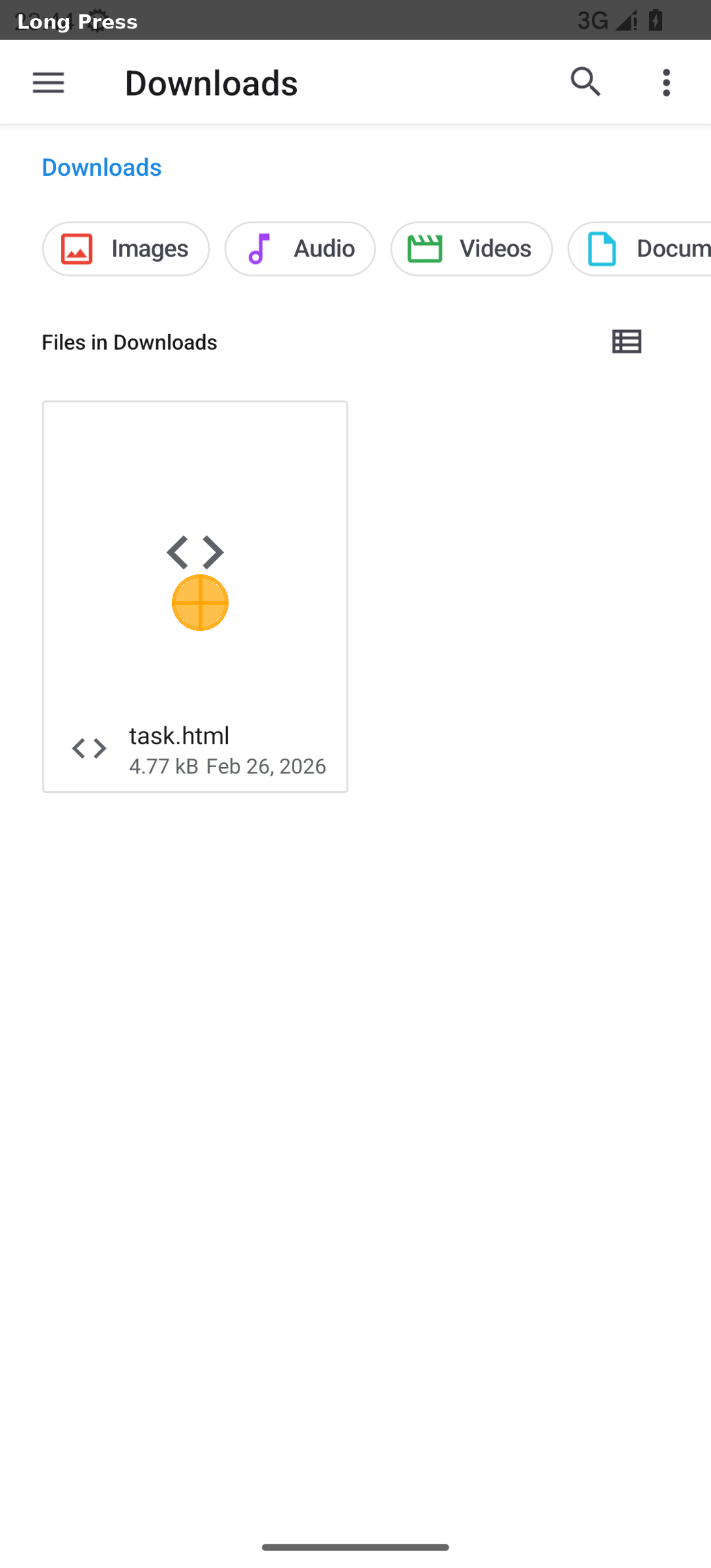}
    }{
        \caseinfo{0.235\textwidth}{
            \casejson{{"action": "long_press", "coordinate": [282, 384]}}
            \casefield{Verdict}{\casenegative}
        }
    }\hfill
    \casepanel{Step 5}{
        \caseimage{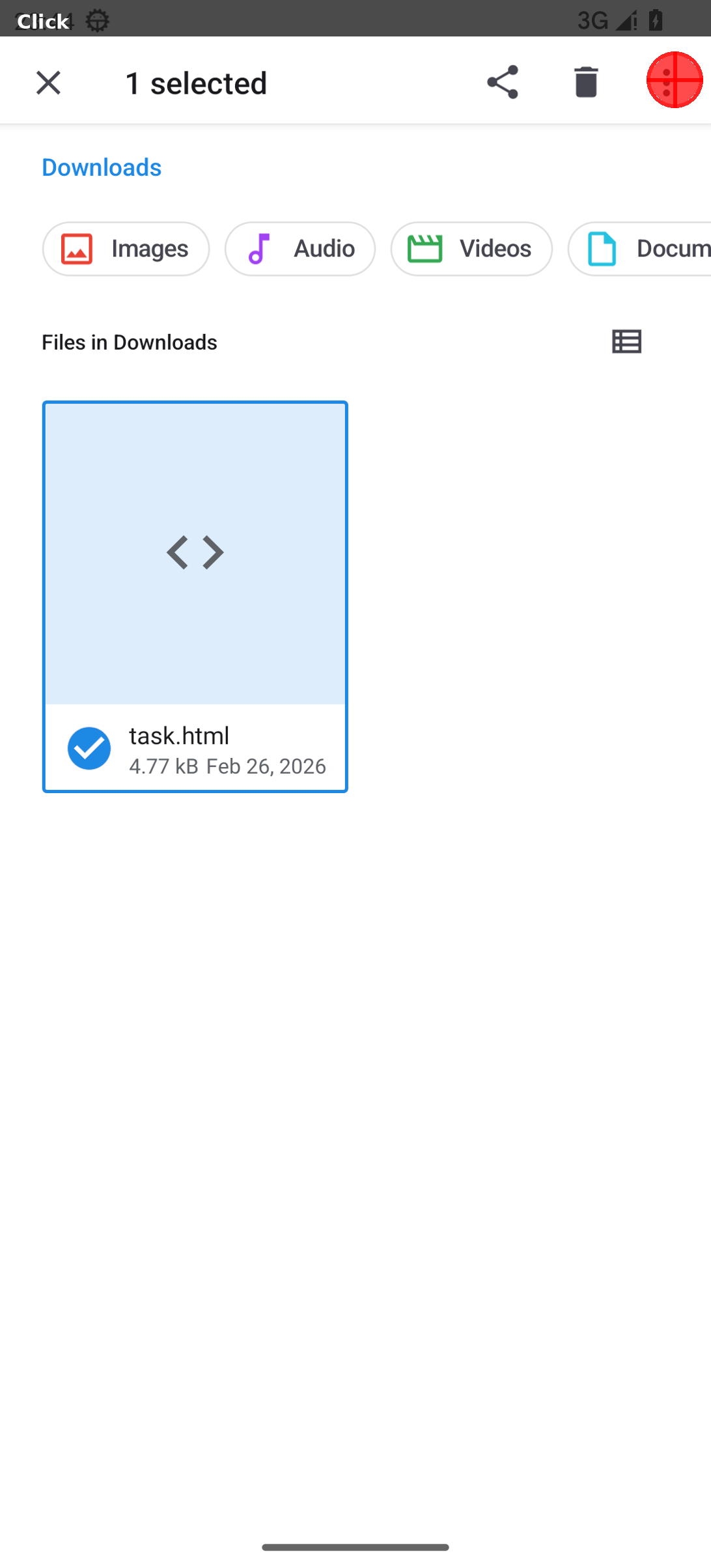}
    }{
        \caseinfo{0.235\textwidth}{
            \casejson{{"action": "click", "coordinate": [949, 51]}}
            \casefield{Verdict}{\casepositive}
        }
    }
    \caption{\textbf{AnchorGUI trajectory on BrowserMaze, steps 1--5.} Unlike ReAct, AnchorGUI turns the failed direct tap and long press into explicit negative CSA evidence. These mismatch records redirect the policy toward the menu-based fallback, which then opens the HTML file successfully.}
    \label{fig:case_anchor_browsermaze}
\end{figure*}

\begin{figure*}[ph!]\ContinuedFloat
    \centering
    \small
    \casepanel{Step 6}{
        \caseimage{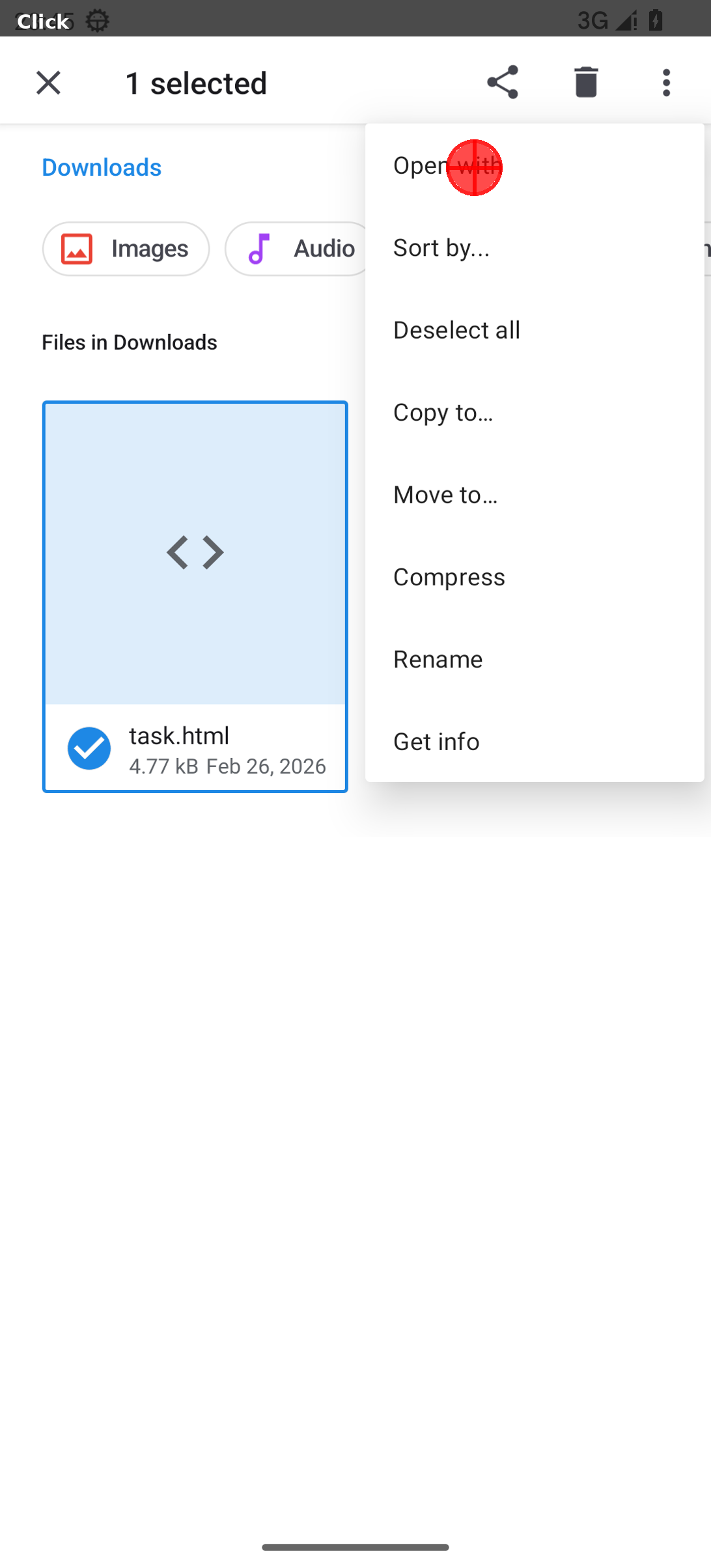}
    }{
        \caseinfo{0.235\textwidth}{
            \casejson{{"action": "click", "coordinate": [666, 107]}}
            \casefield{Verdict}{\casepositive}
        }
    }\hfill
    \casepanel{Step 7}{
        \caseimage{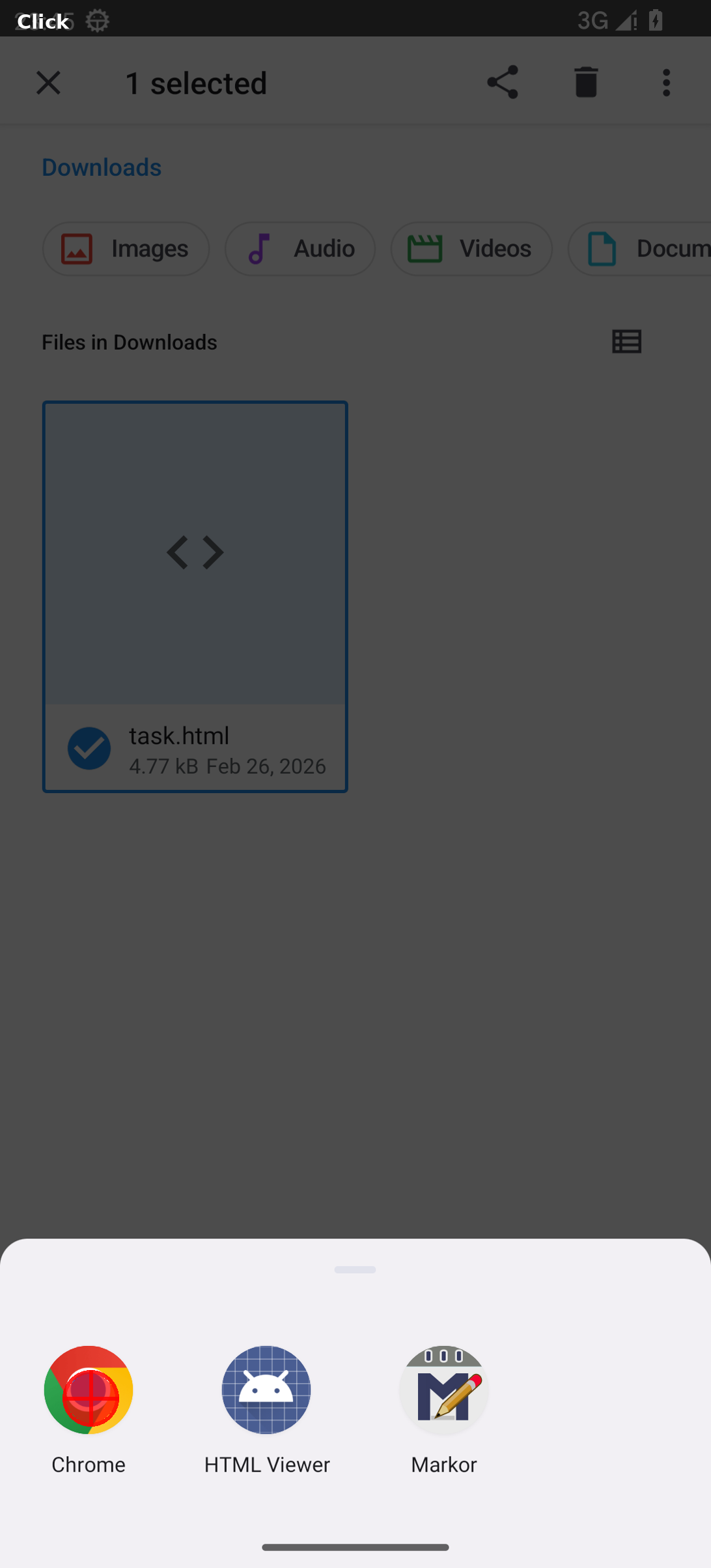}
    }{
        \caseinfo{0.235\textwidth}{
            \casejson{{"action": "click", "coordinate": [128, 891]}}
            \casefield{Verdict}{\casepositive}
        }
    }\hfill
    \casepanel{Step 8}{
        \caseimage{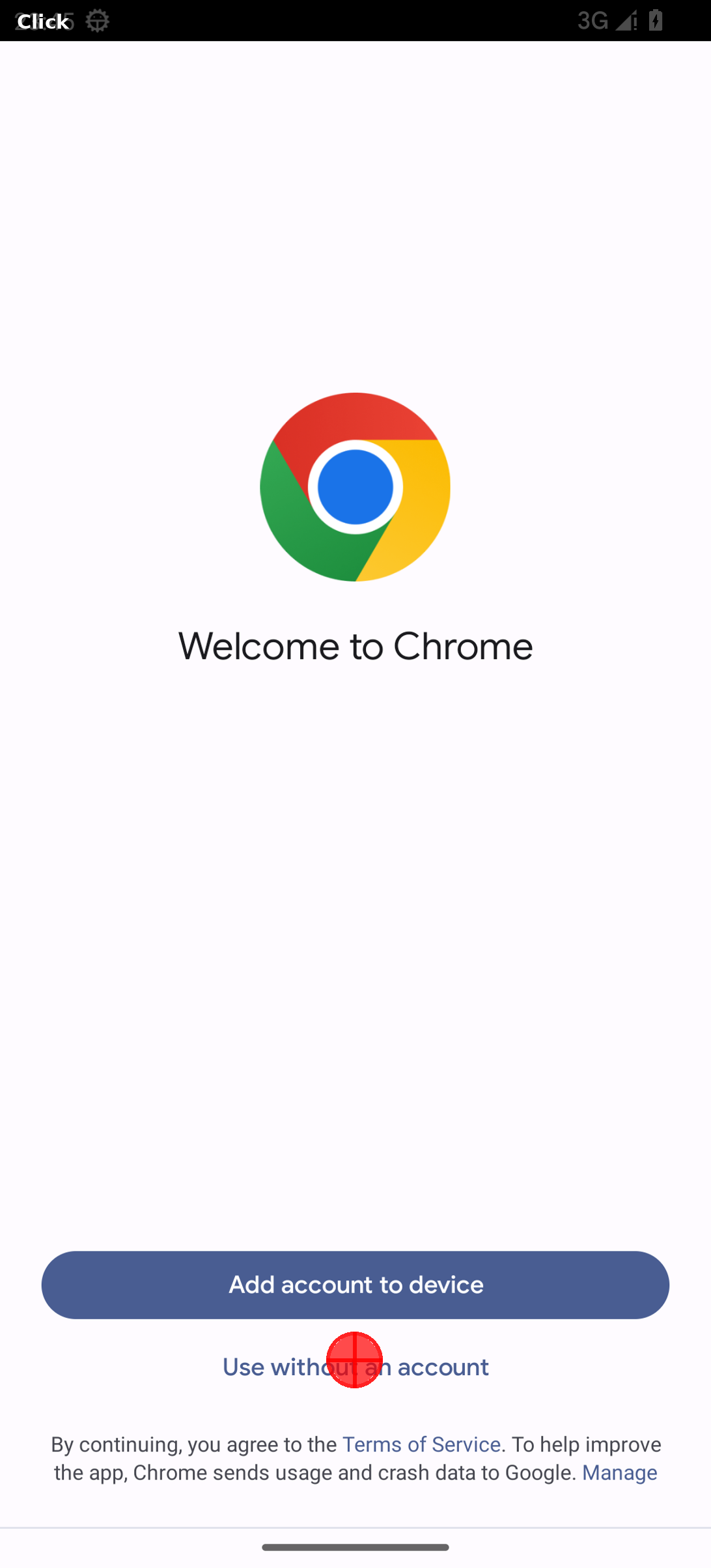}
    }{
        \caseinfo{0.235\textwidth}{
            \casejson{{"action": "click", "coordinate": [499, 867]}}
            \casefield{Verdict}{\casenegative}
        }
    }\hfill
    \casepanel{Step 9}{
        \caseimage{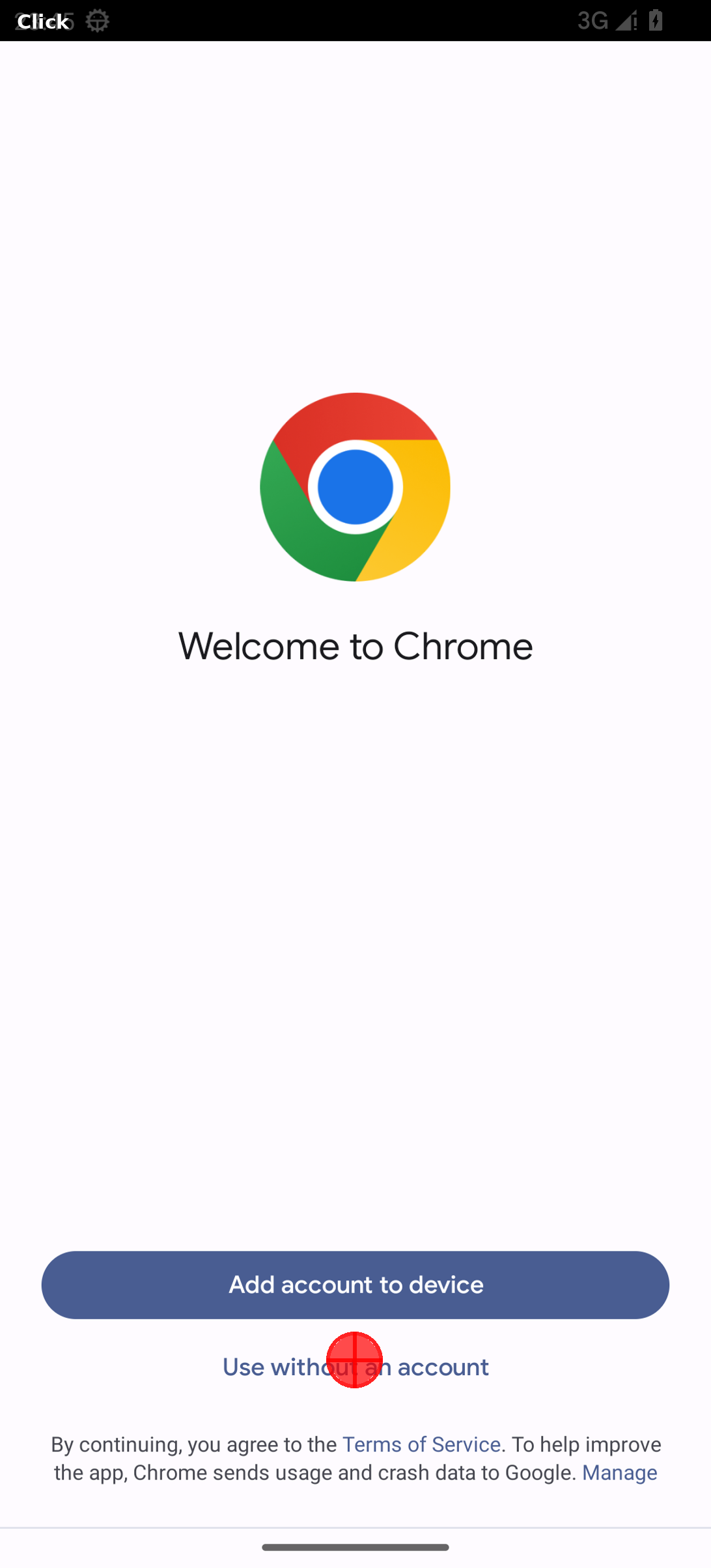}
    }{
        \caseinfo{0.235\textwidth}{
            \casejson{{"action": "click", "coordinate": [499, 867]}}
            \casefield{Verdict}{\casenegative}
        }
    }\hfill
    \casepanel{Step 10}{
        \caseimage{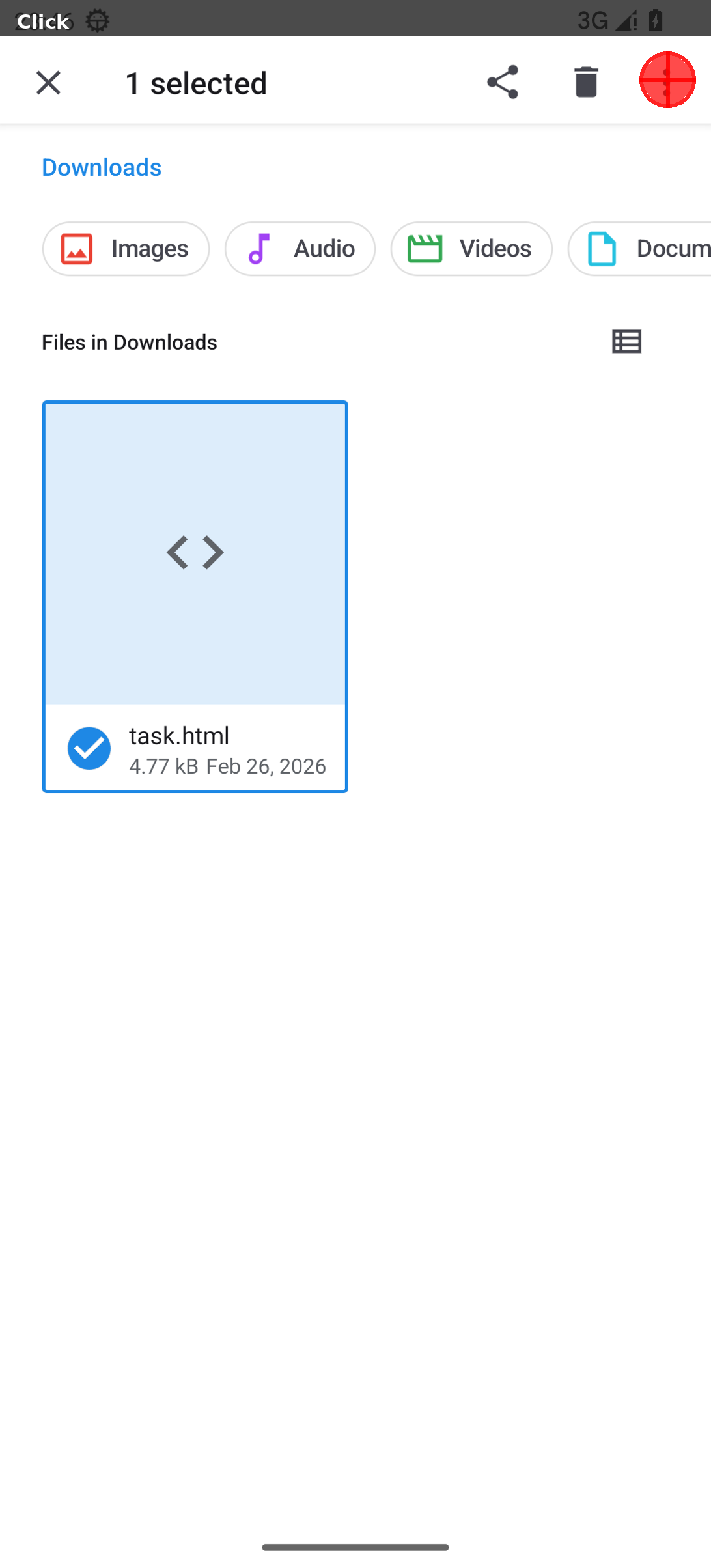}
    }{
        \caseinfo{0.235\textwidth}{
            \casejson{{"action": "click", "coordinate": [938, 51]}}
            \casefield{Verdict}{\casepositive}
        }
    }
    \caption[]{\textbf{AnchorGUI trajectory on BrowserMaze (continued), steps 6--10.}}
\end{figure*}

\begin{figure*}[ph!]\ContinuedFloat
    \centering
    \small
    \casepanel{Step 11}{
        \caseimage{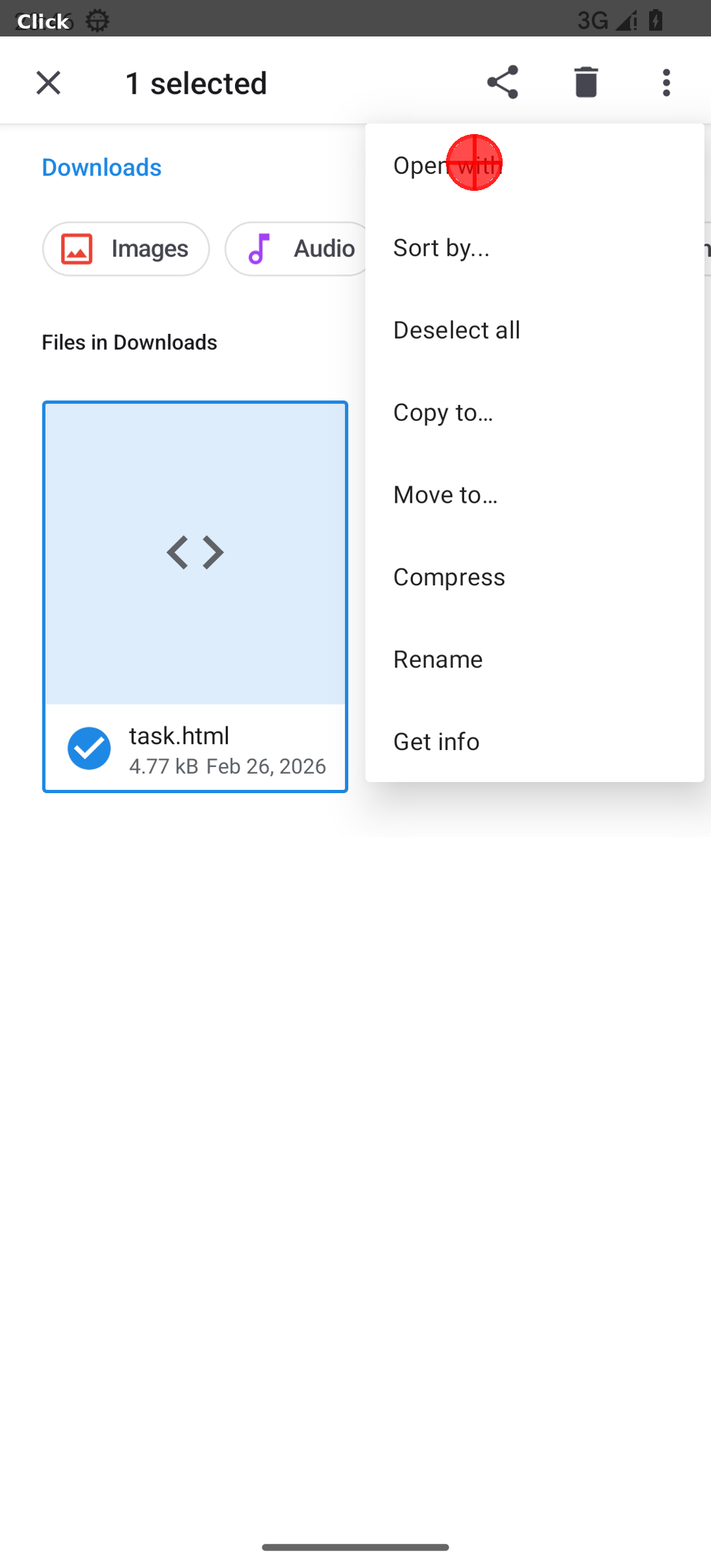}
    }{
        \caseinfo{0.235\textwidth}{
            \casejson{{"action": "click", "coordinate": [666, 104]}}
            \casefield{Verdict}{\casepositive}
        }
    }\hfill
    \casepanel{Step 12}{
        \caseimage{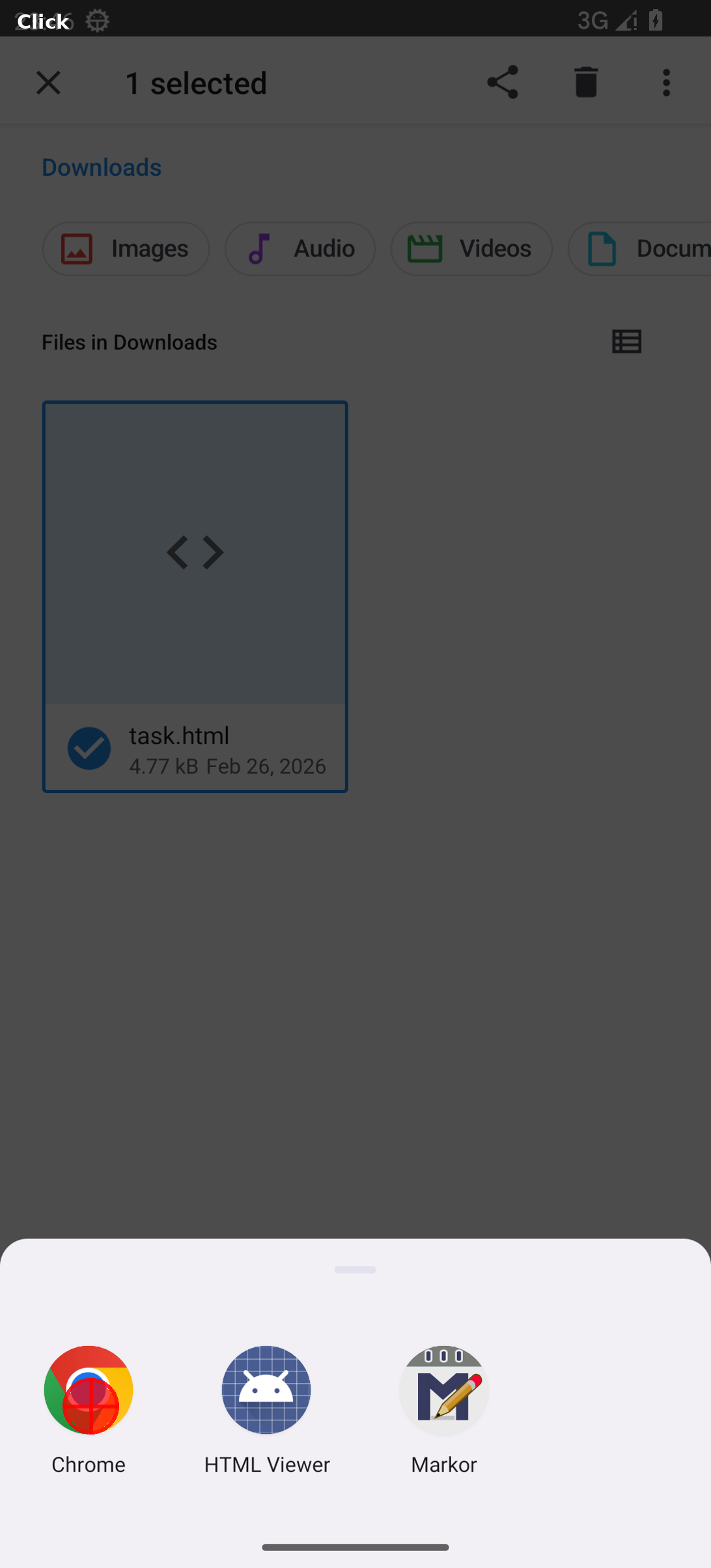}
    }{
        \caseinfo{0.235\textwidth}{
            \casejson{{"action": "click", "coordinate": [128, 896]}}
            \casefield{Verdict}{\casepositive}
        }
    }\hfill
    \casepanel{Step 13}{
        \caseimage{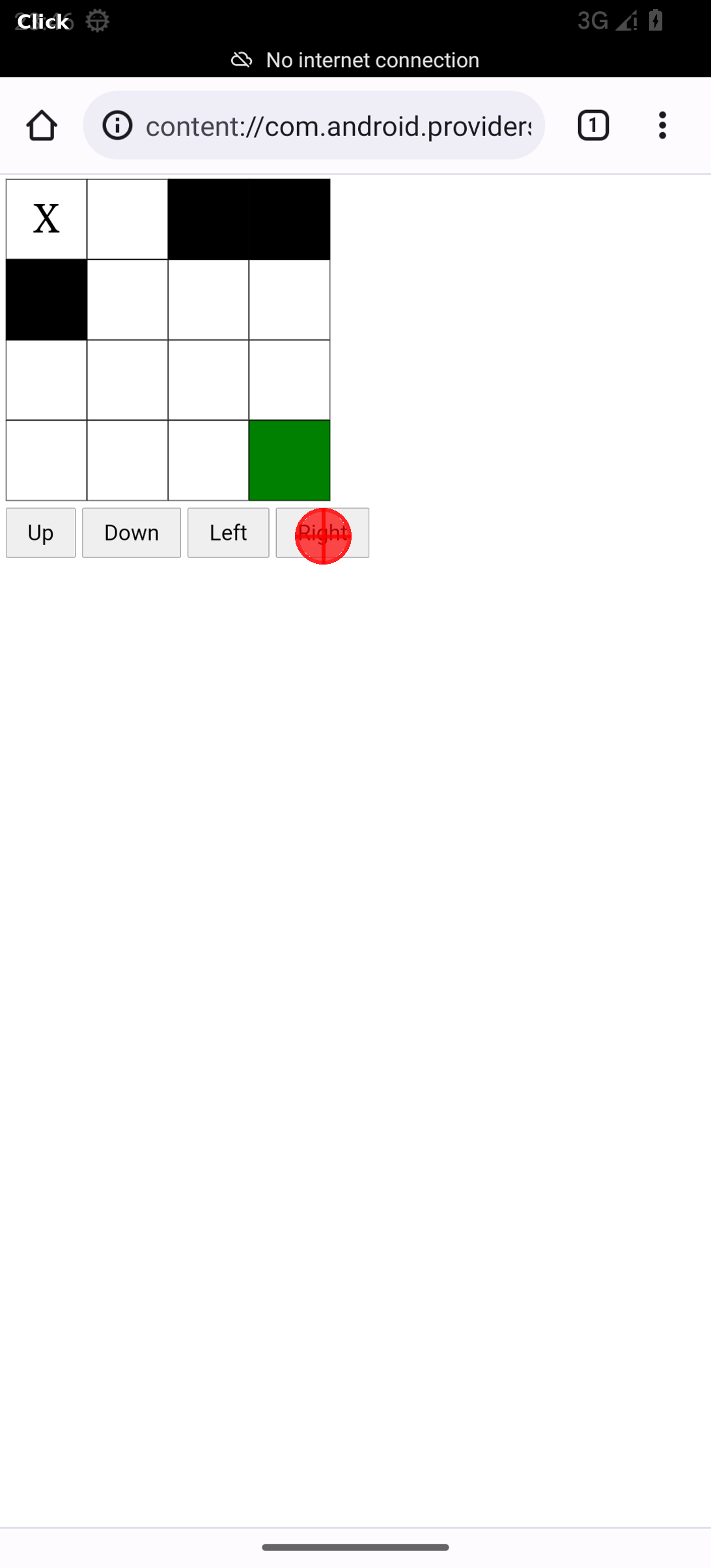}
    }{
        \caseinfo{0.235\textwidth}{
            \casejson{{"action": "click", "coordinate": [455, 342]}}
            \casefield{Verdict}{\casepositive}
        }
    }\hfill
    \casepanel{Step 14}{
        \caseimage{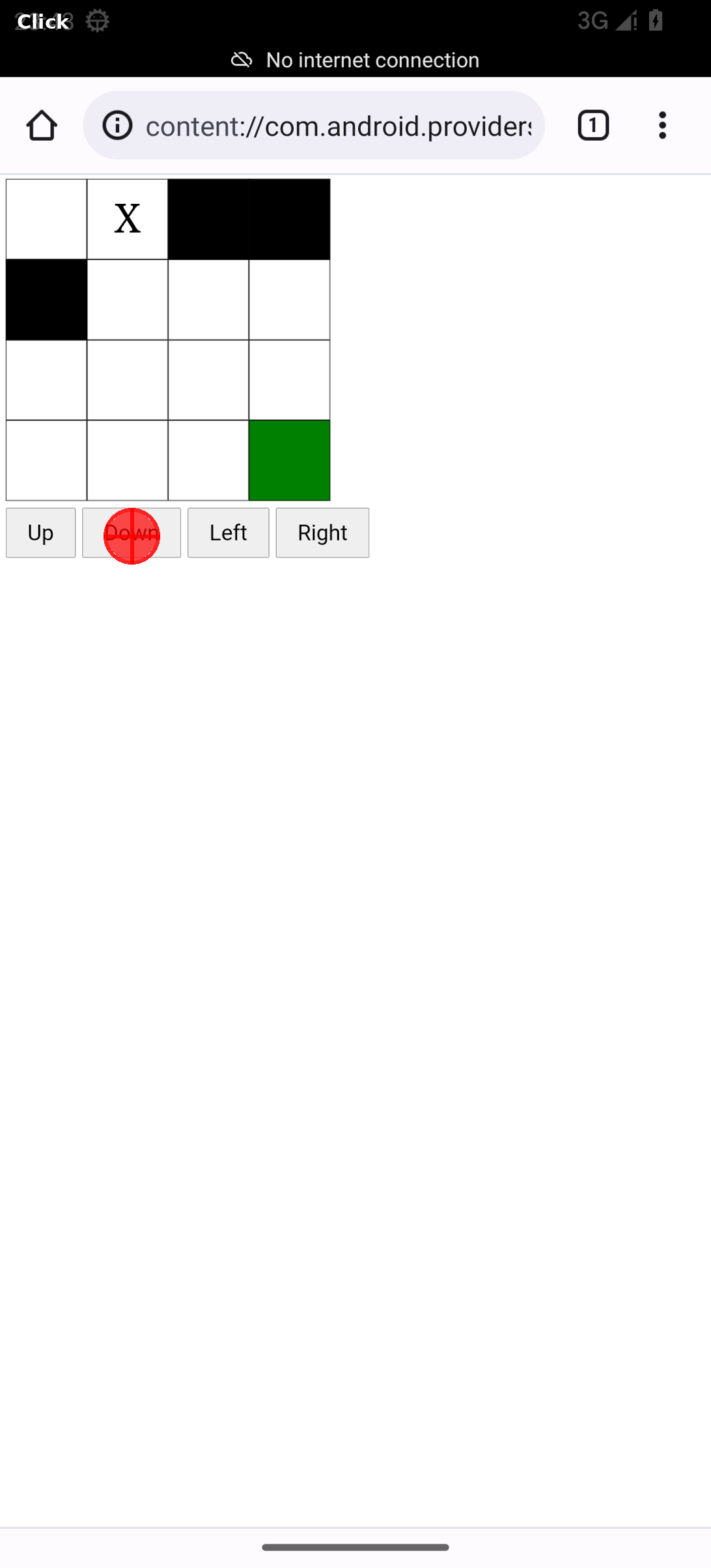}
    }{
        \caseinfo{0.235\textwidth}{
            \casejson{{"action": "click", "coordinate": [185, 342]}}
            \casefield{Verdict}{\casepositive}
        }
    }\hfill
    \casepanel{Step 15}{
        \caseimage{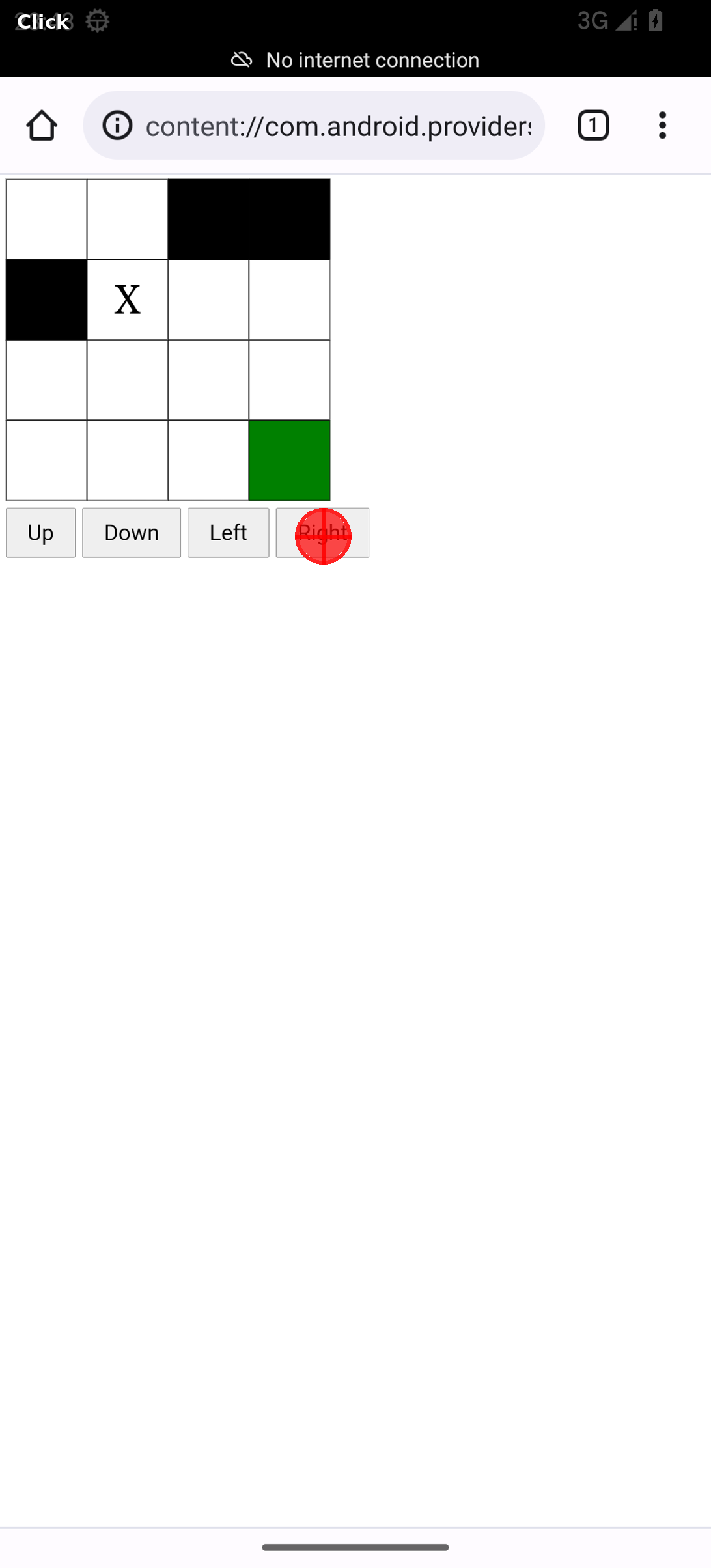}
    }{
        \caseinfo{0.235\textwidth}{
            \casejson{{"action": "click", "coordinate": [455, 342]}}
            \casefield{Verdict}{\casepositive}
        }
    }
    \caption[]{\textbf{AnchorGUI trajectory on BrowserMaze (continued), steps 11--15.}}
\end{figure*}

\begin{figure*}[ph!]\ContinuedFloat
    \centering
    \small
    \casepanel{Step 16}{
        \caseimage{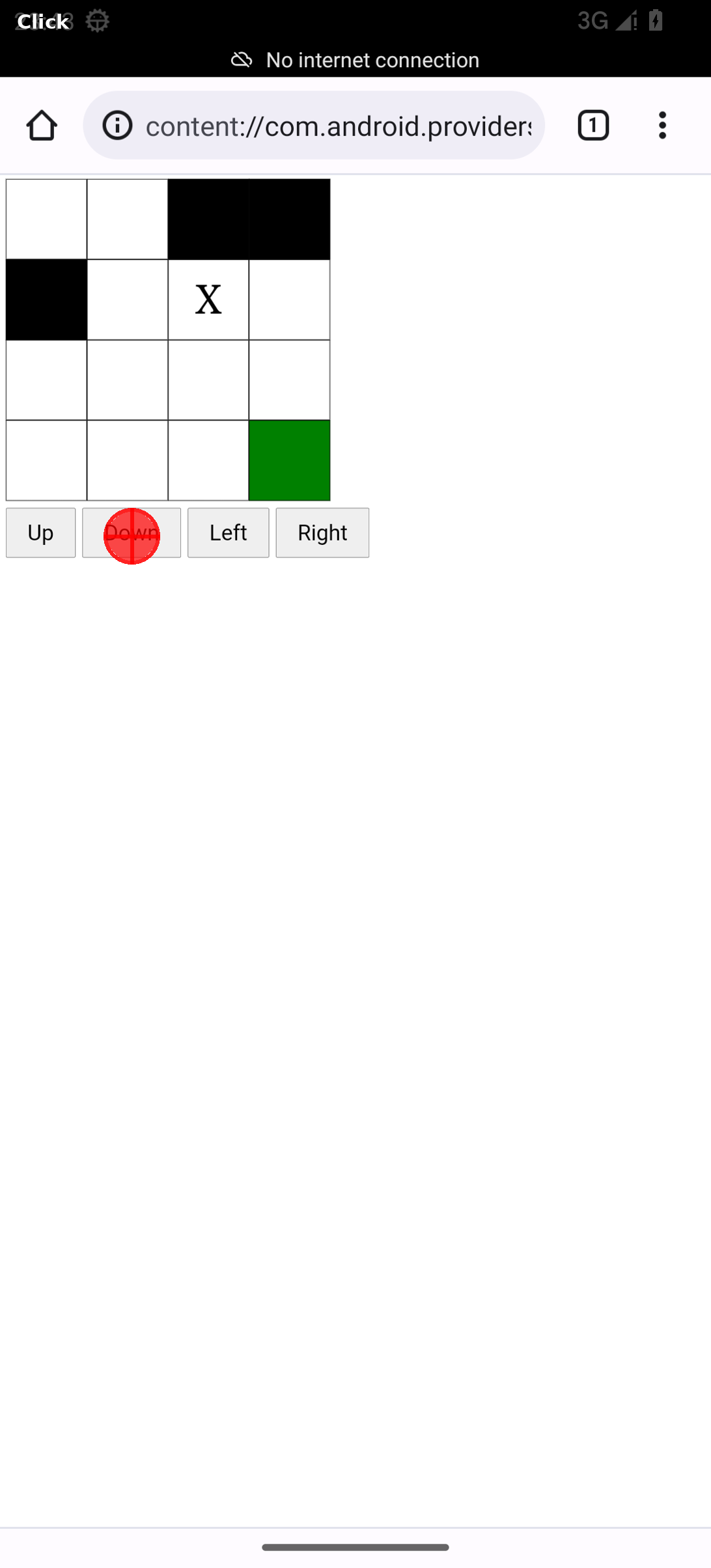}
    }{
        \caseinfo{0.235\textwidth}{
            \casejson{{"action": "click", "coordinate": [185, 342]}}
            \casefield{Verdict}{\casepositive}
        }
    }\hfill
    \casepanel{Step 17}{
        \caseimage{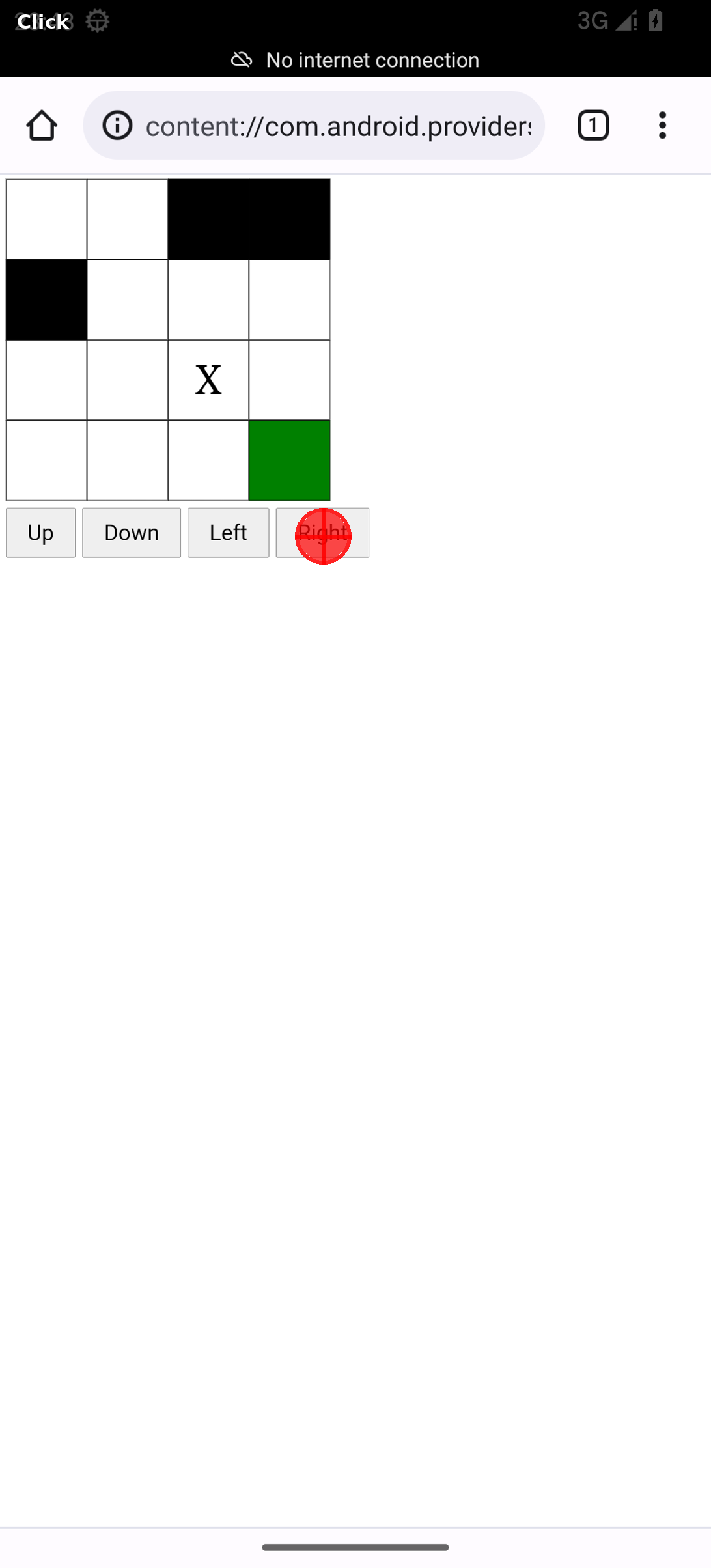}
    }{
        \caseinfo{0.235\textwidth}{
            \casejson{{"action": "click", "coordinate": [455, 342]}}
            \casefield{Verdict}{\casepositive}
        }
    }\hfill
    \casepanel{Step 18}{
        \caseimage{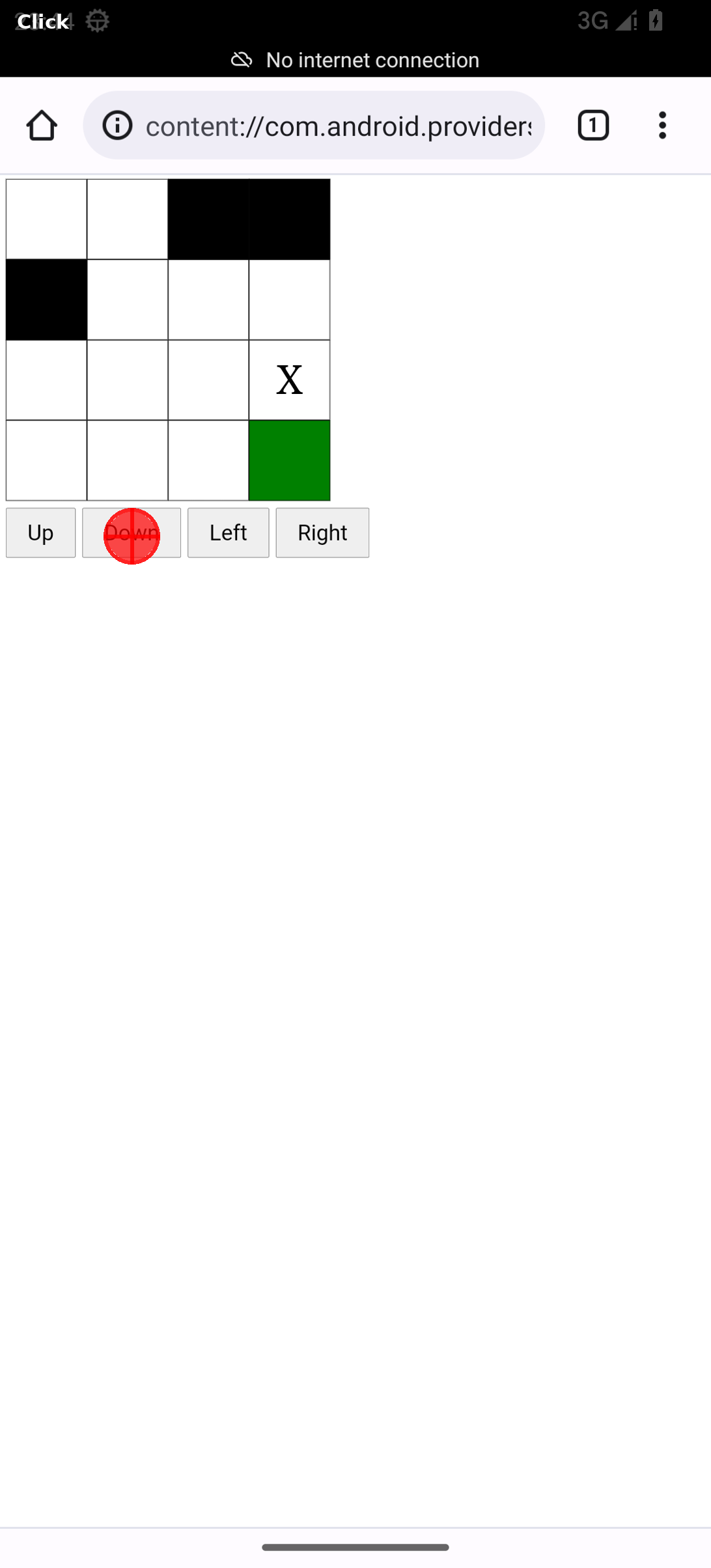}
    }{
        \caseinfo{0.235\textwidth}{
            \casejson{{"action": "click", "coordinate": [185, 342]}}
            \casefield{Verdict}{\casepositive}
        }
    }\hfill
    \casepanel{Step 19}{
        \caseimage{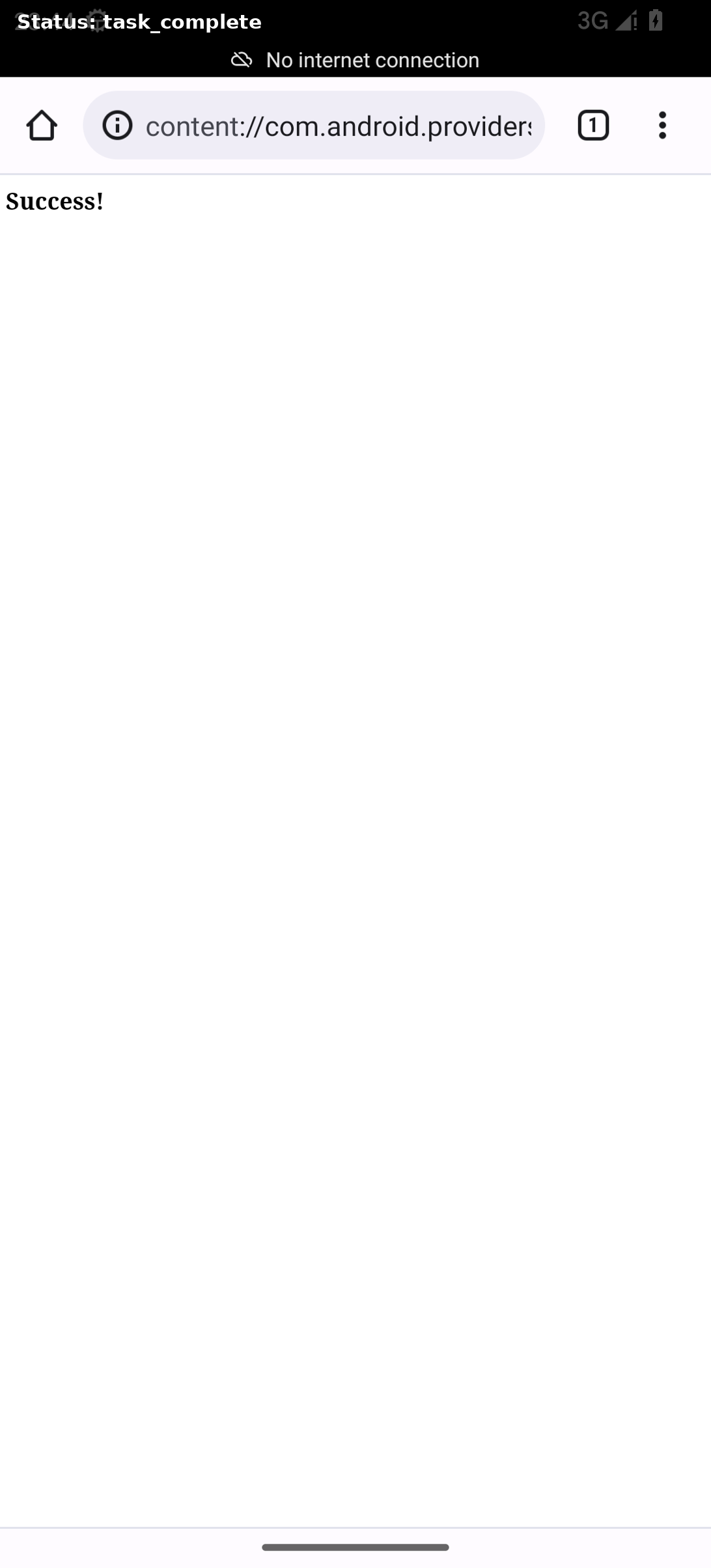}
    }{
        \caseinfo{0.235\textwidth}{
            \casejson{{"action": "terminate", "status": "success"}}
        }
    }\hfill
    \casepanel{End}{
        \caseblankplaceholder
    }{
        \caseinfo{0.162\textwidth}{
        }
    }
    \caption[]{\textbf{AnchorGUI trajectory on BrowserMaze (continued), steps 16--19 and termination.} Across the full episode, negative CSA records act as explicit correction anchors: they first redirect file opening from direct tap to \texttt{Open with}, and later help the policy recover from local maze-control mismatches instead of repeating them blindly.}
\end{figure*}

The contrast between Figures~\ref{fig:case_react_browsermaze} and~\ref{fig:case_anchor_browsermaze} makes the role of explicit mismatch supervision visible at the trajectory level. ReAct starts from a plausible plan, but once the direct tap fails to open the file, it never updates that hypothesis and falls into a degenerate repetition loop. AnchorGUI instead converts each mismatch into explicit failure evidence. After observing that direct click does not work, it tries a long press; after that also fails, it switches to the context-menu route, selects \emph{Open with}, and successfully enters the HTML file.

\subsection{AnchorGUI First Trial vs.\ Cross-Trial Retry}
\label{sec:supp_case_expense}

We next examine a failure mode that is harder to expose with step-level execution traces alone: \emph{semantic drift under locally successful actions}. In the first trial, every interaction is individually executable and receives positive local feedback, yet the trajectory still misses the user intent. As Figure~\ref{fig:case_anchor_first_expense} shows, the agent types the wrong amount (\$100 instead of \$307.01), settles for the immediately visible \emph{Social} category rather than swiping to reveal \emph{Health Care}, and writes a generic note unrelated to the request. Because the final save operation also succeeds mechanically, the episode terminates despite producing an incorrect expense record.

\begin{figure*}[ph!]
    \centering
    \small
    \casepanel{Step 1}{
        \caseimage{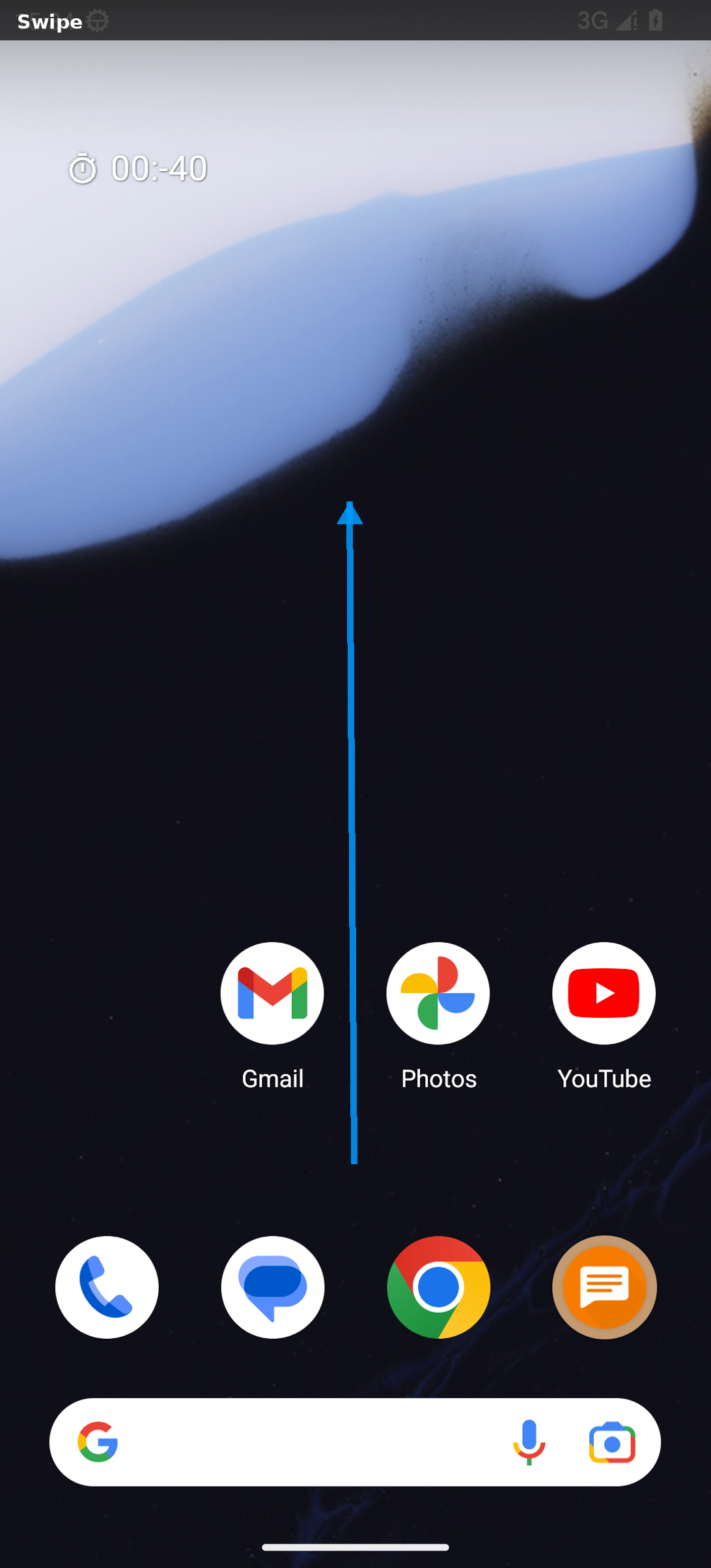}
    }{
        \caseinfo{0.235\textwidth}{
            \casejson{{"action": "swipe", "coordinate": [499, 742], "coordinate2": [493, 320]}}
            \casefield{Verdict}{\casepositive}
        }
    }\hfill
    \casepanel{Step 2}{
        \caseimage{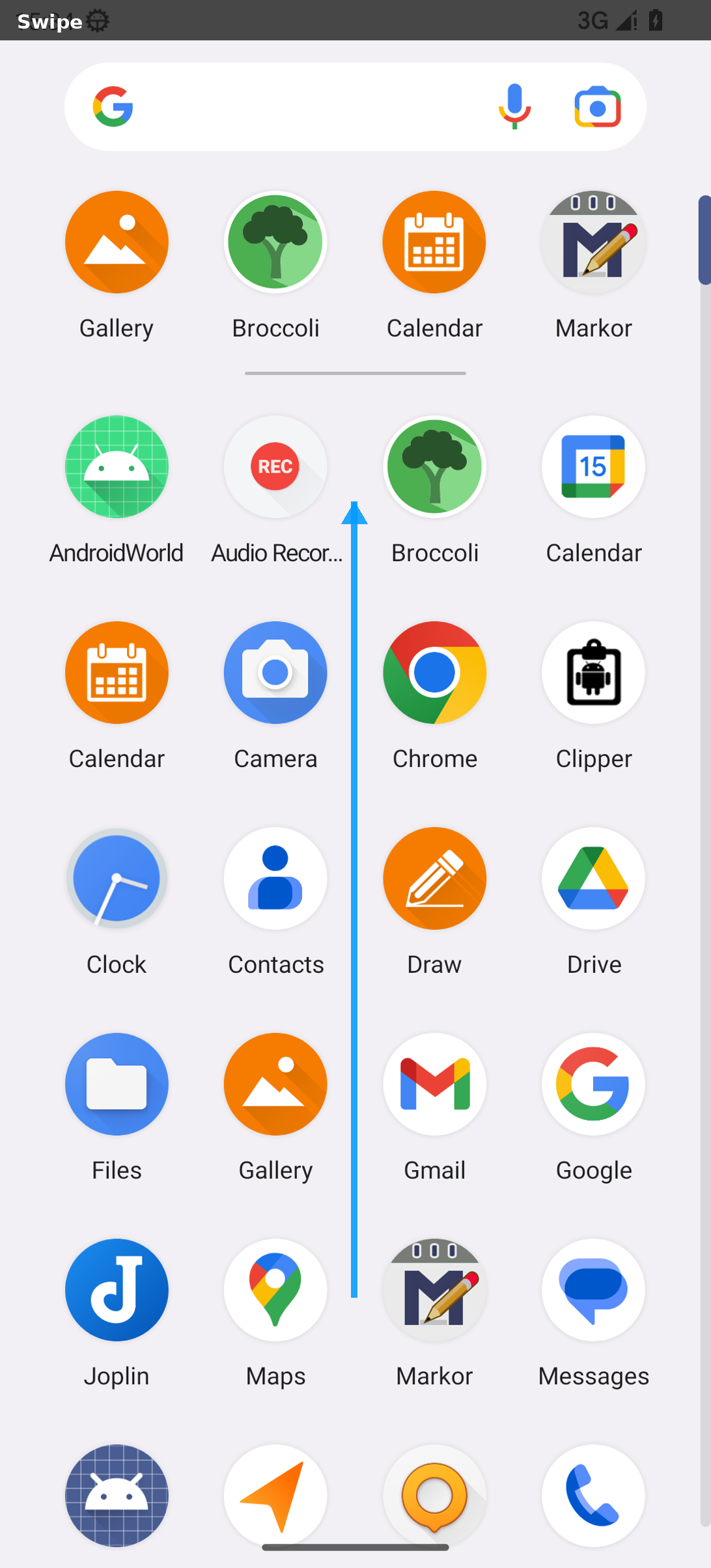}
    }{
        \caseinfo{0.235\textwidth}{
            \casejson{{"action": "swipe", "coordinate": [499, 827], "coordinate2": [499, 320]}}
            \casefield{Verdict}{\casepositive}
        }
    }\hfill
    \casepanel{Step 3}{
        \caseimage{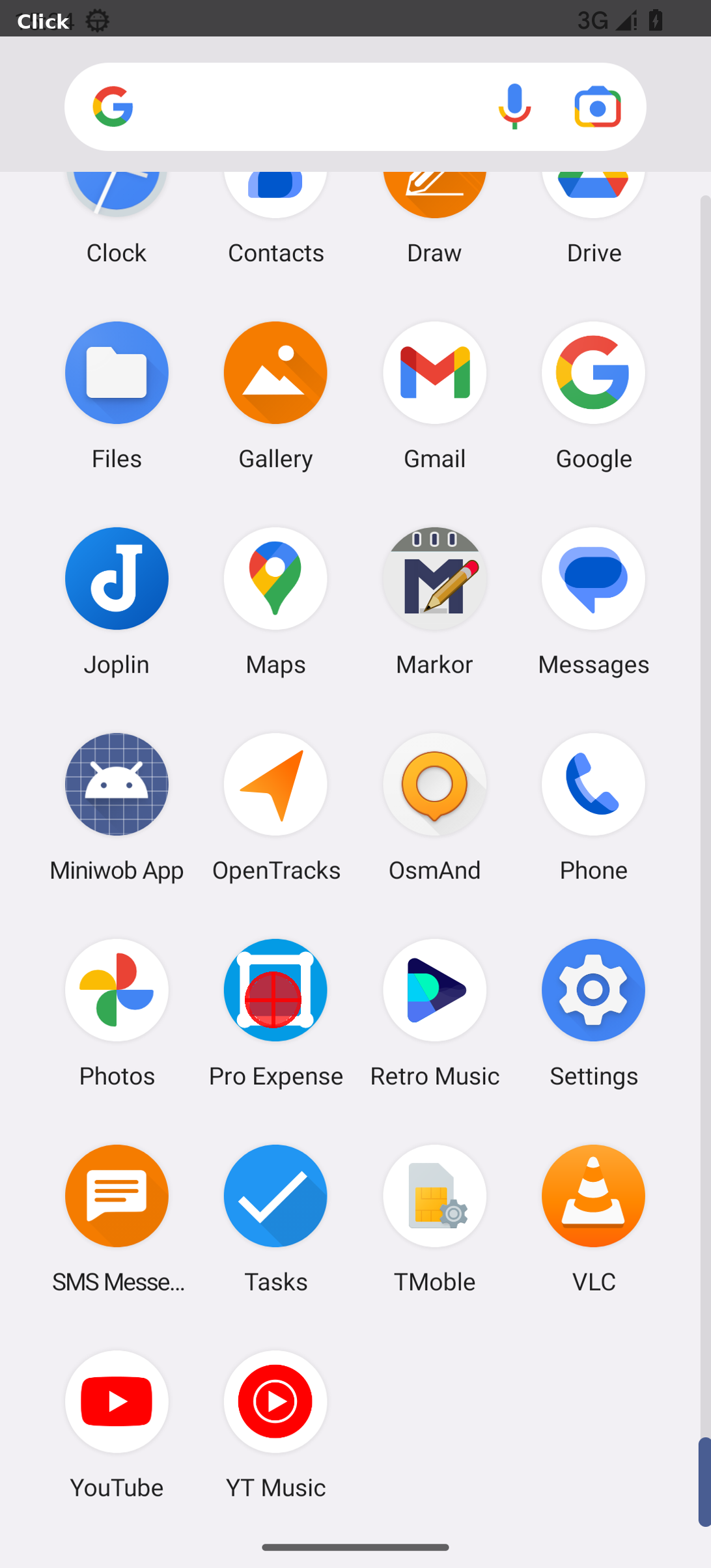}
    }{
        \caseinfo{0.235\textwidth}{
            \casejson{{"action": "click", "coordinate": [384, 637]}}
            \casefield{Verdict}{\casepositive}
        }
    }\hfill
    \casepanel{Step 4}{
        \caseimage{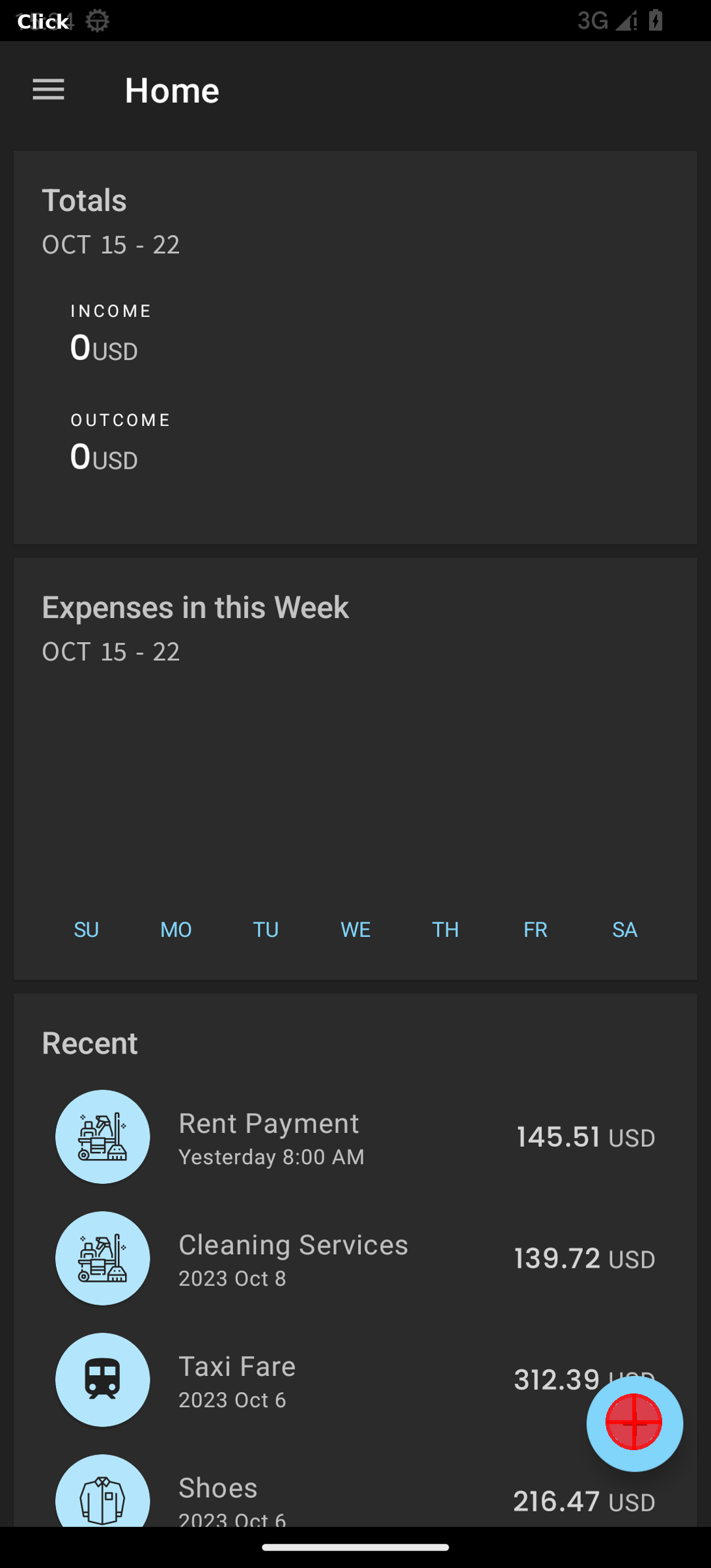}
    }{
        \caseinfo{0.235\textwidth}{
            \casejson{{"action": "click", "coordinate": [891, 906]}}
            \casefield{Verdict}{\casepositive}
        }
    }\hfill
    \casepanel{Step 5}{
        \caseimage{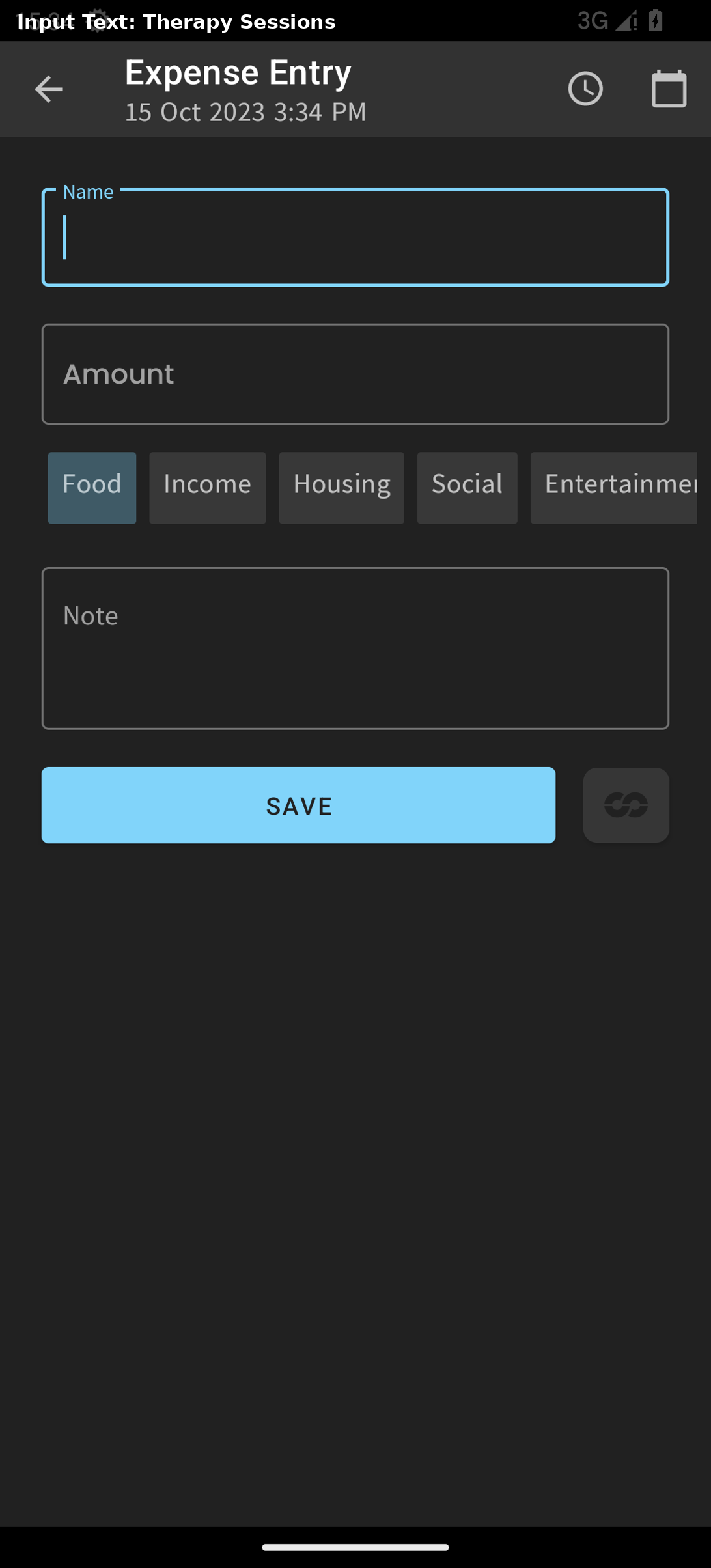}
    }{
        \caseinfo{0.235\textwidth}{
            \casejson{{"action": "type", "text": "Therapy Sessions"}}
            \casefield{Verdict}{\casepositive}
        }
    }
    \caption{\textbf{AnchorGUI first trial on Pro Expense, steps 1--5.} The rollout is locally smooth: every interaction receives a positive evaluator verdict while the agent opens the app and starts filling the form.}
    \label{fig:case_anchor_first_expense}
\end{figure*}

\begin{figure*}[ph!]\ContinuedFloat
    \centering
    \small
    \casepanel{Step 6}{
        \caseimage{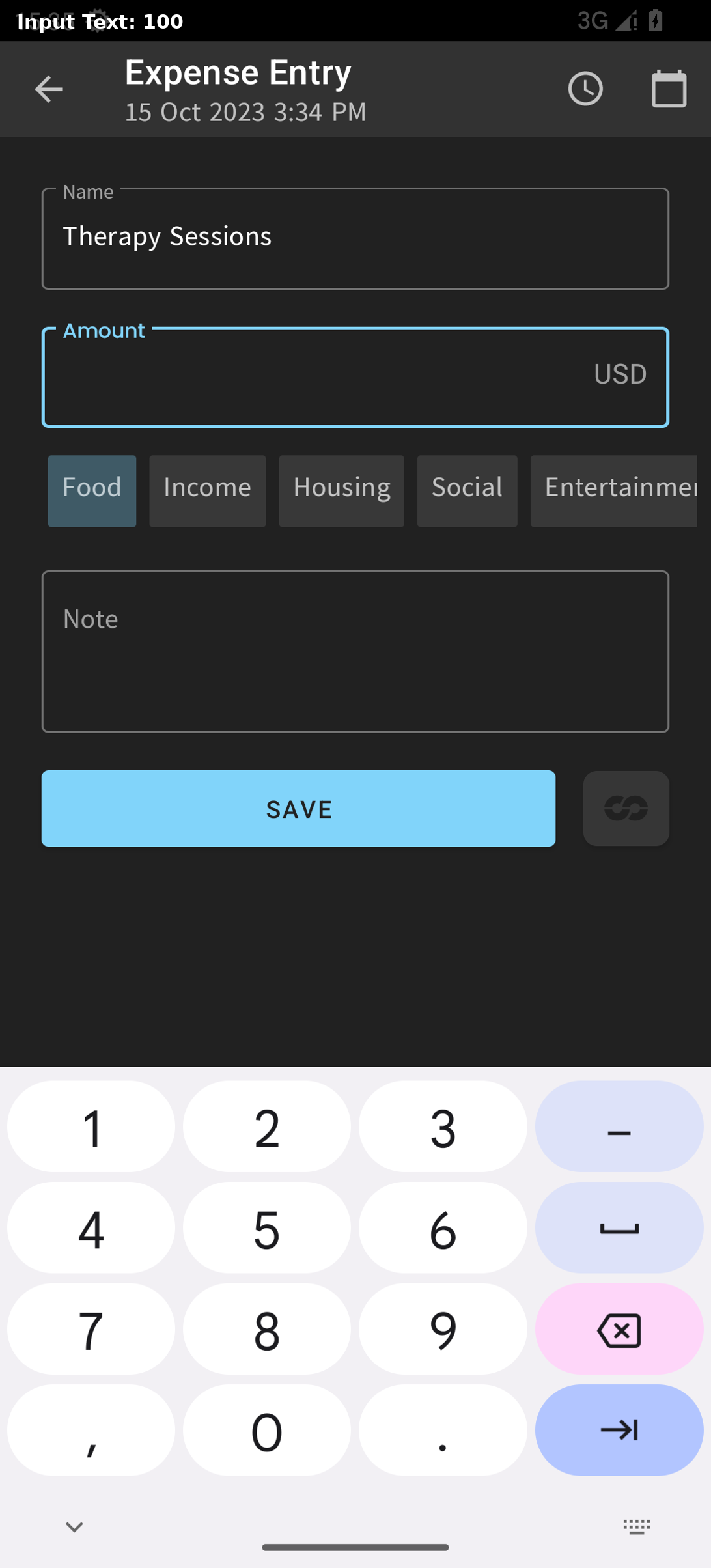}
    }{
        \caseinfo{0.235\textwidth}{
            \casejson{{"action": "type", "text": "100"}}
            \casefield{Verdict}{\casepositive}
        }
    }\hfill
    \casepanel{Step 7}{
        \caseimage{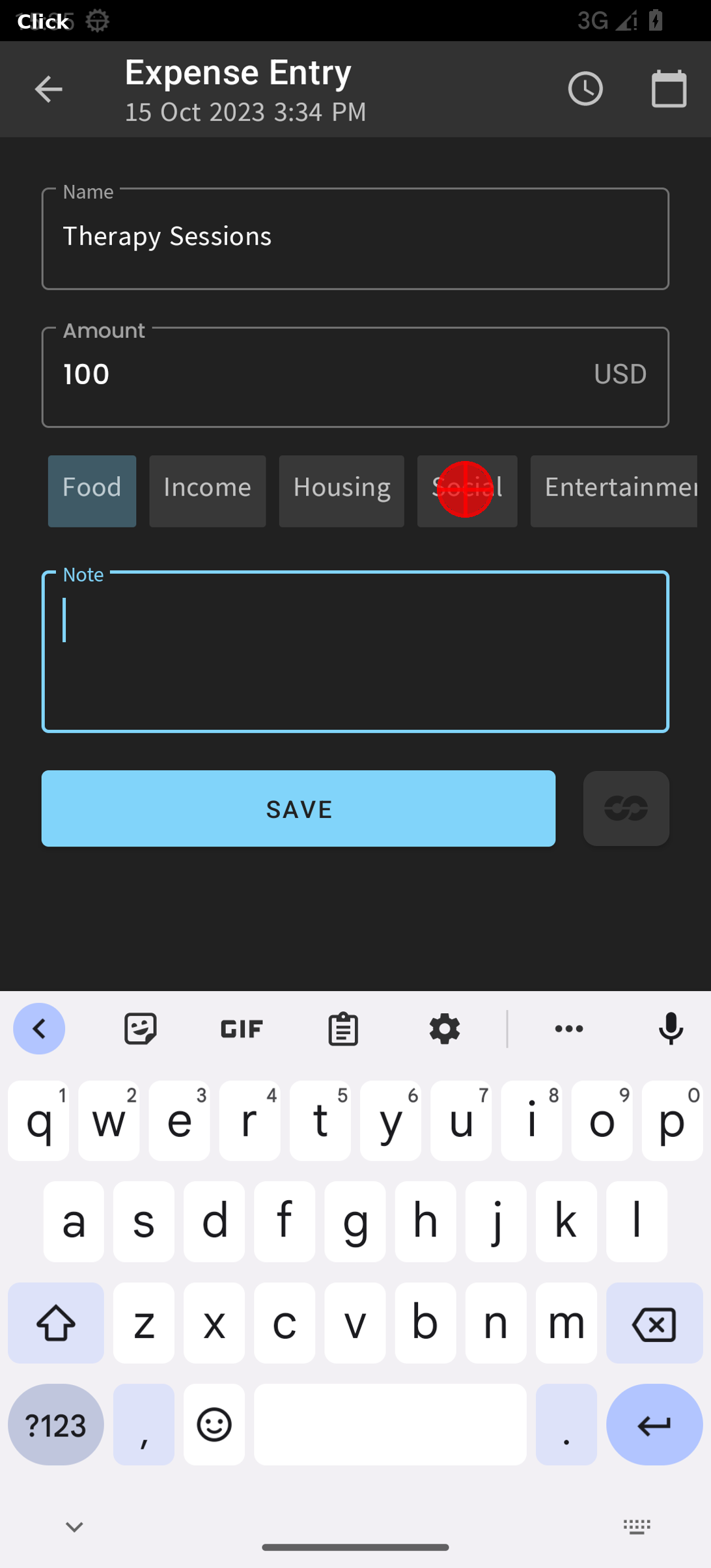}
    }{
        \caseinfo{0.235\textwidth}{
            \casejson{{"action": "click", "coordinate": [654, 312]}}
            \casefield{Verdict}{\casepositive}
        }
    }\hfill
    \casepanel{Step 8}{
        \caseimage{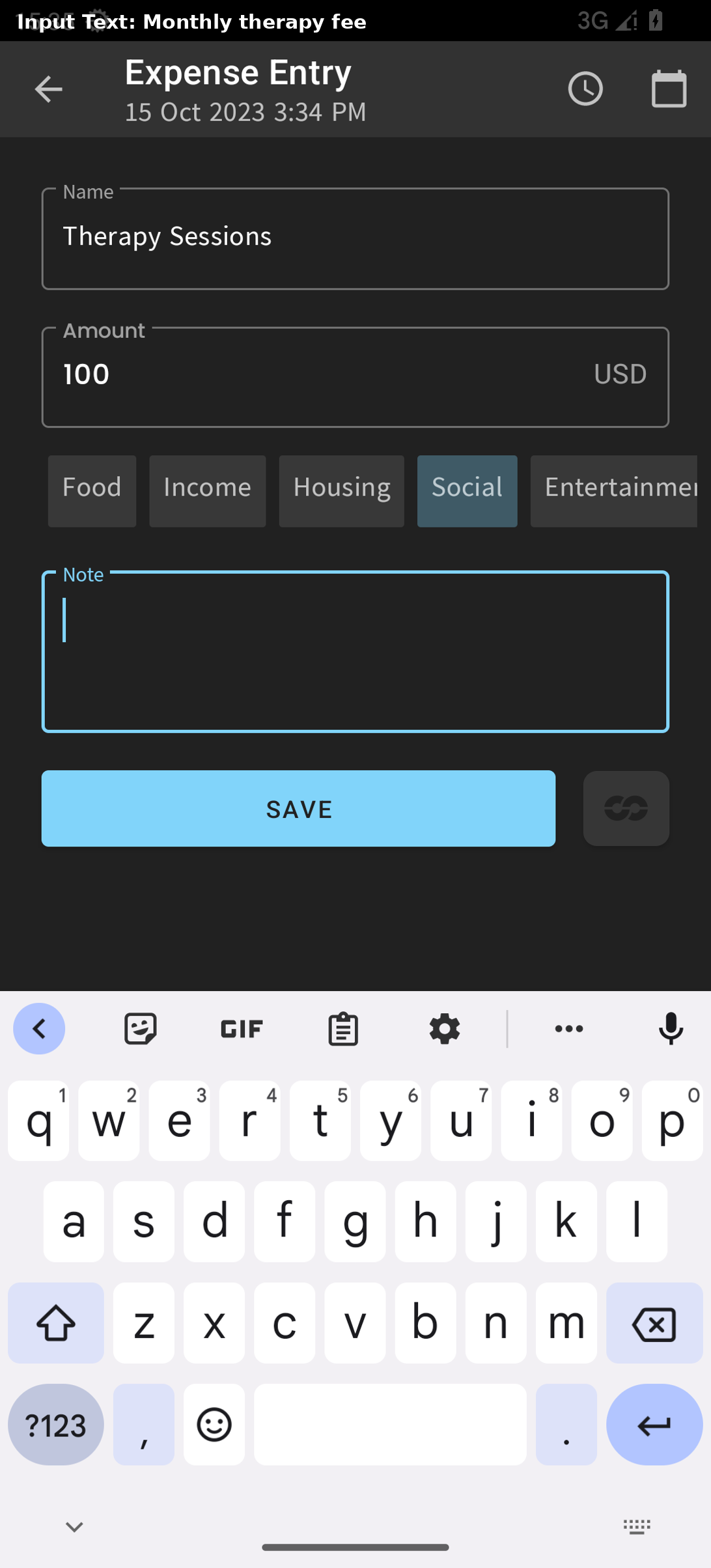}
    }{
        \caseinfo{0.235\textwidth}{
            \casejson{{"action": "type", "text": "Monthly therapy fee"}}
            \casefield{Verdict}{\casepositive}
        }
    }\hfill
    \casepanel{Step 9}{
        \caseimage{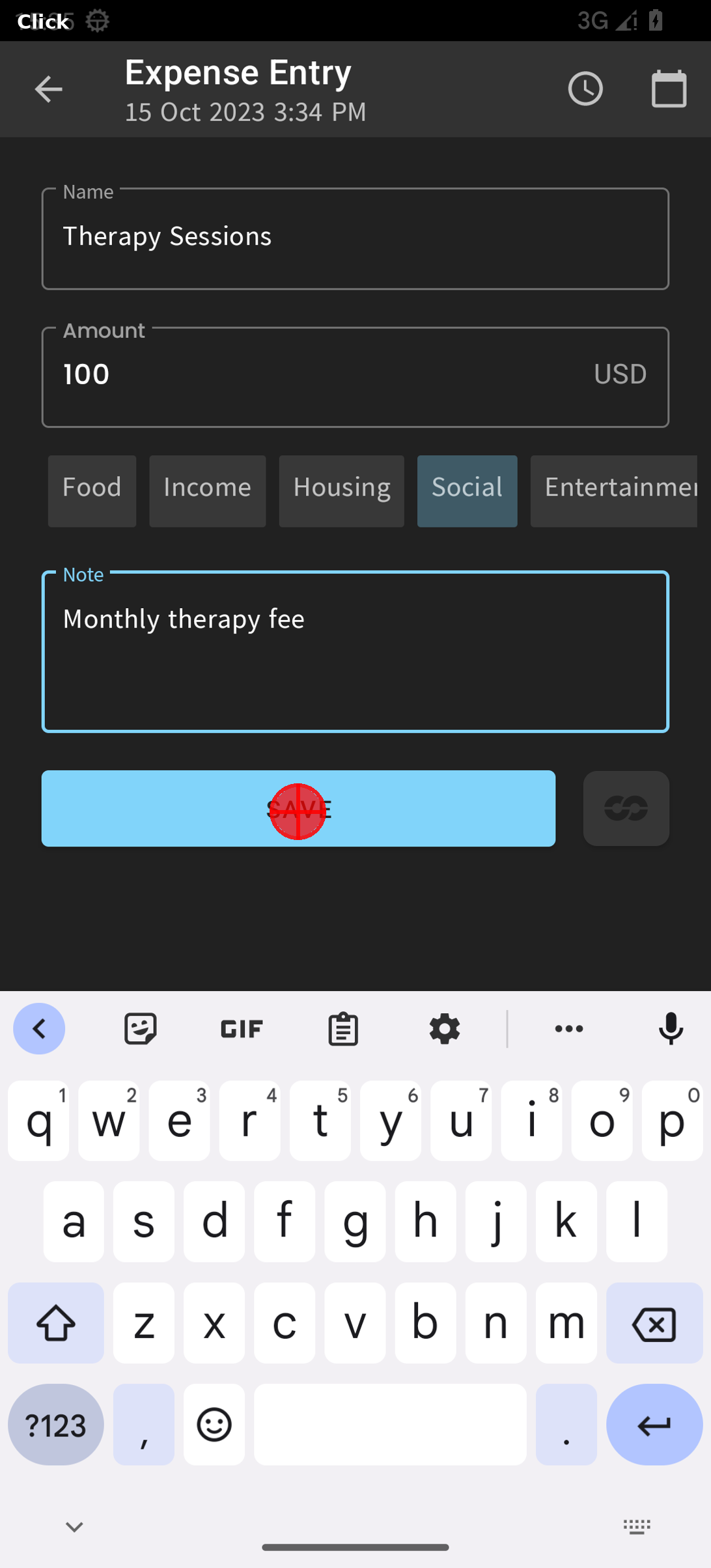}
    }{
        \caseinfo{0.235\textwidth}{
            \casejson{{"action": "click", "coordinate": [419, 517]}}
            \casefield{Verdict}{\casepositive}
        }
    }\hfill
    \casepanel{Step 10}{
        \caseimage{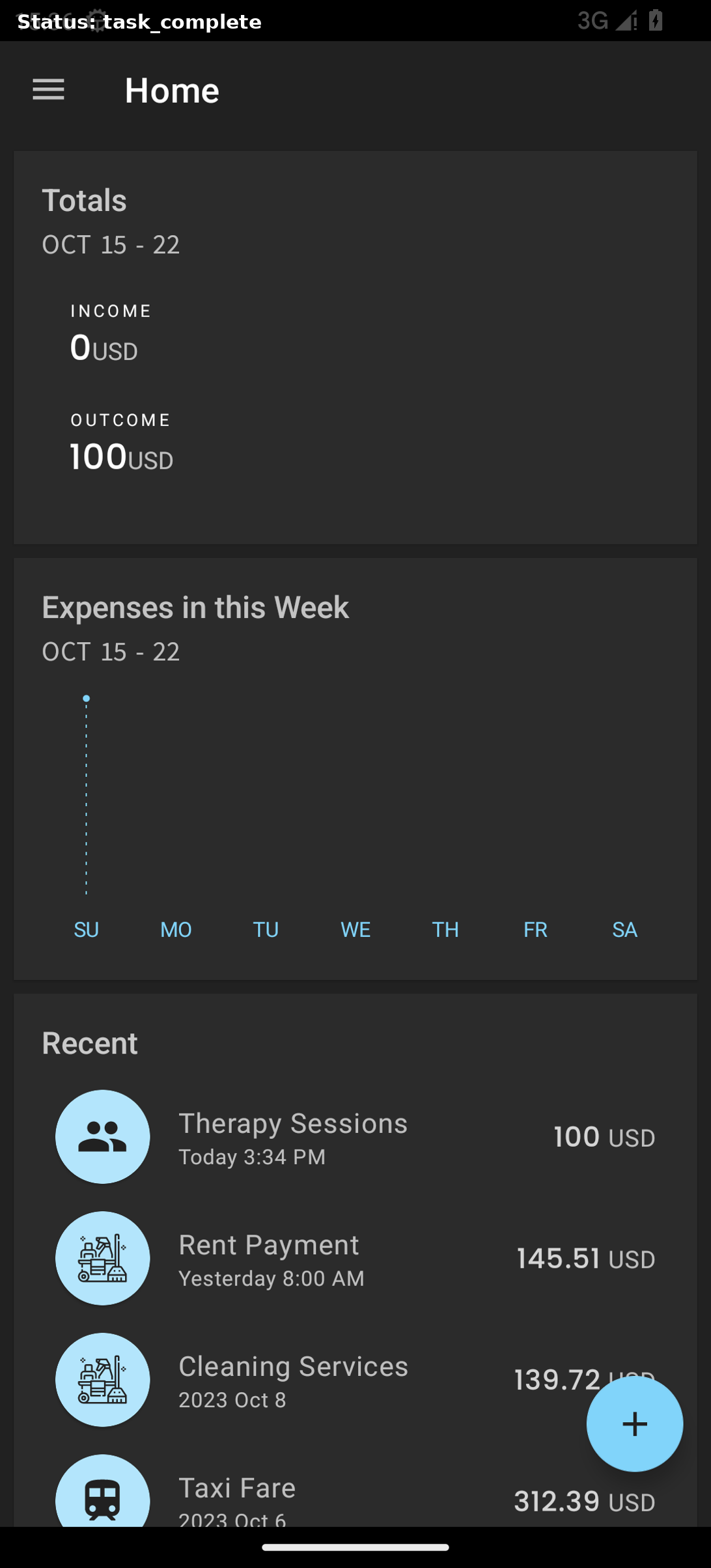}
    }{
        \caseinfo{0.235\textwidth}{
            \casejson{{"action": "terminate", "status": "success"}}
        }
    }
    \caption[]{\textbf{AnchorGUI first trial on Pro Expense (continued), steps 6--10.} The failure is therefore not caused by visible execution breakdowns. Instead, the agent drifts semantically while still collecting only positive local verdicts: it enters the wrong amount, selects \emph{Social}, writes the wrong note, and terminates after a mechanically successful save.}
\end{figure*}

This case is important precisely because the per-step evaluator does \emph{not} emit negative verdicts in the first trial. The screenshots and logs show a harder failure mode: local action intent is satisfied, while the global task semantics are still wrong. Cross-trial learning therefore has to operate above step-level executability, using the final failed outcome to recover missing constraints and feed them back into the next attempt as an explicit corrected plan.

\begin{promptlisting}{Cross-Trial Improved Plan for Pro Expense}
Improved plan:
1) Start from the initial page/state and specify the first required operation.
   - Open the app drawer and locate the "Pro Expense" app (as previously done, but validate app is correctly identified and opened).
2) Tap the "+" button to open the expense entry form.
3) Type "Therapy Sessions" into the "Name" field.
4) Type "307.01" into the "Amount" field --- ensure decimal point and correct number of digits are entered.
5) Swipe left on category buttons to reveal and select "Health" category (if not immediately visible).
6) Tap the "Health" category to confirm selection.
7) Locate and tap the "Note" field (if not auto-focused) and type "I may repeat this".
8) Tap the "Save" button to finalize the entry.
9) Verify the entry appears correctly in the Recent expenses list with all fields matching: Name, Amount, Category, and Note.
10) If any step fails (e.g., field not visible, incorrect input format, or save fails), trigger fallback:
   - IF the "Amount" field does not accept decimal input or shows "150" despite typing "307.01" THEN DO NOT assume the field is editable; INSTEAD tap the field to ensure focus, clear it, and retype "307.01" with explicit decimal formatting.
   - IF the "Note" field is missing or not visible THEN DO NOT skip it; INSTEAD swipe to reveal it or tap the "More Options" button to expand fields, then enter the note.
\end{promptlisting}

The second attempt demonstrates what cross-trial distillation changes. The distilled experience explicitly identifies the omitted constraints from the failed rollout and converts them into a corrected plan: validate the exact decimal amount, reveal the hidden category by swiping, and fill the missing note before saving. Figure~\ref{fig:case_anchor_cross_expense} shows the full retry trajectory with verdict annotations. The result is a qualitatively different episode: rather than merely repeating successful low-level operations, AnchorGUI uses past failure evidence to repair task-level semantics and produce the correct entry.

\begin{figure*}[ph!]
    \centering
    \small
    \casepanel{Step 1}{
        \caseimage{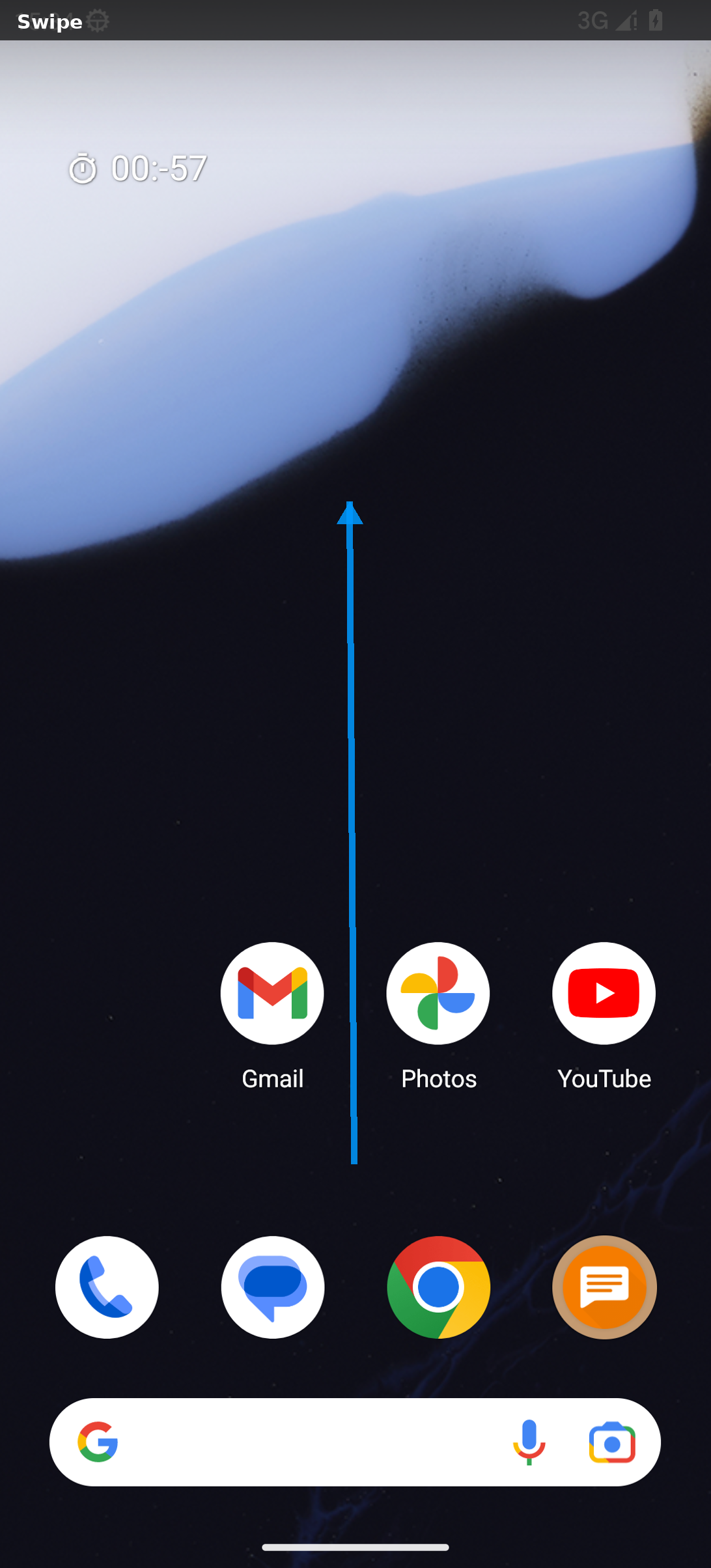}
    }{
        \caseinfo{0.235\textwidth}{
            \casejson{{"action": "swipe", "coordinate": [499, 742], "coordinate2": [493, 320]}}
            \casefield{Verdict}{\casepositive}
        }
    }\hfill
    \casepanel{Step 2}{
        \caseimage{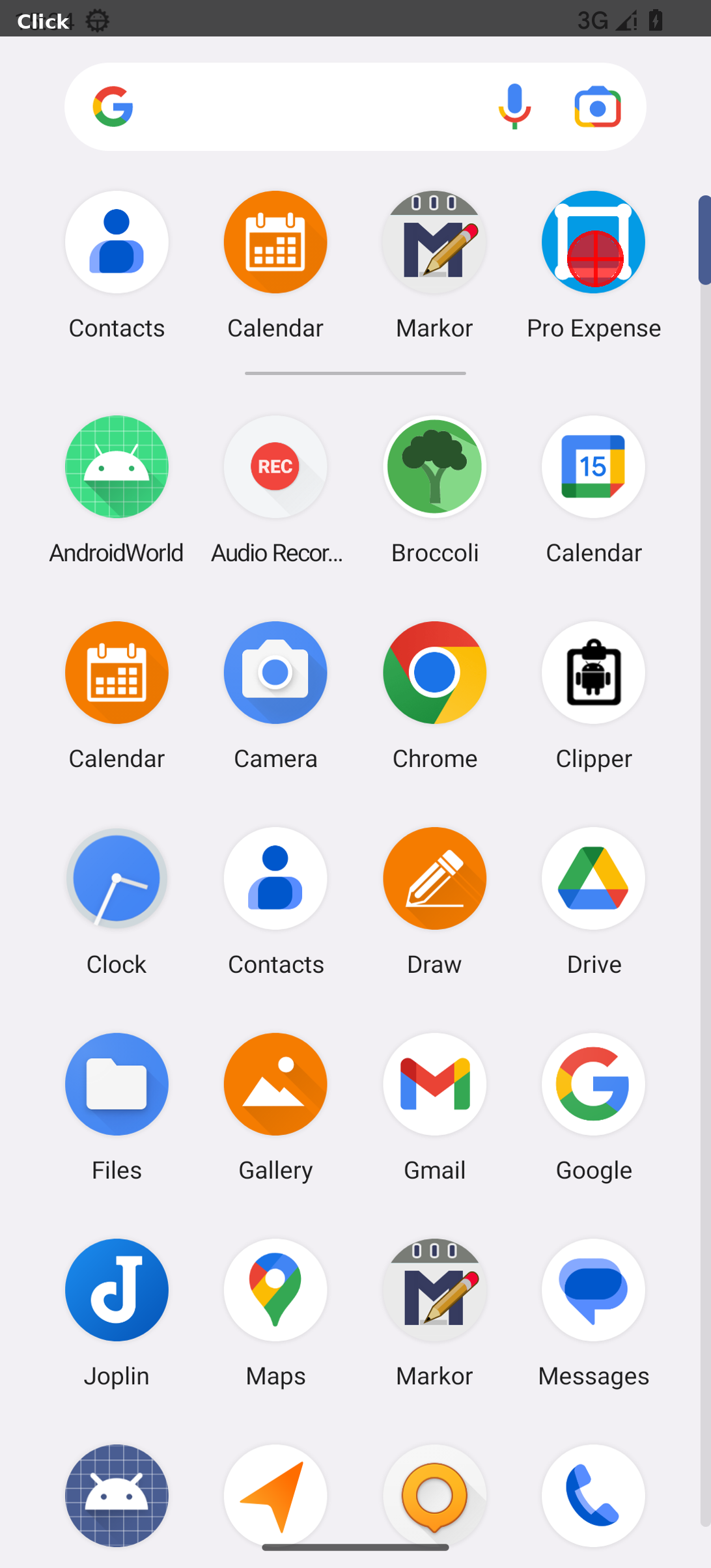}
    }{
        \caseinfo{0.235\textwidth}{
            \casejson{{"action": "click", "coordinate": [837, 165]}}
            \casefield{Verdict}{\casepositive}
        }
    }\hfill
    \casepanel{Step 3}{
        \caseimage{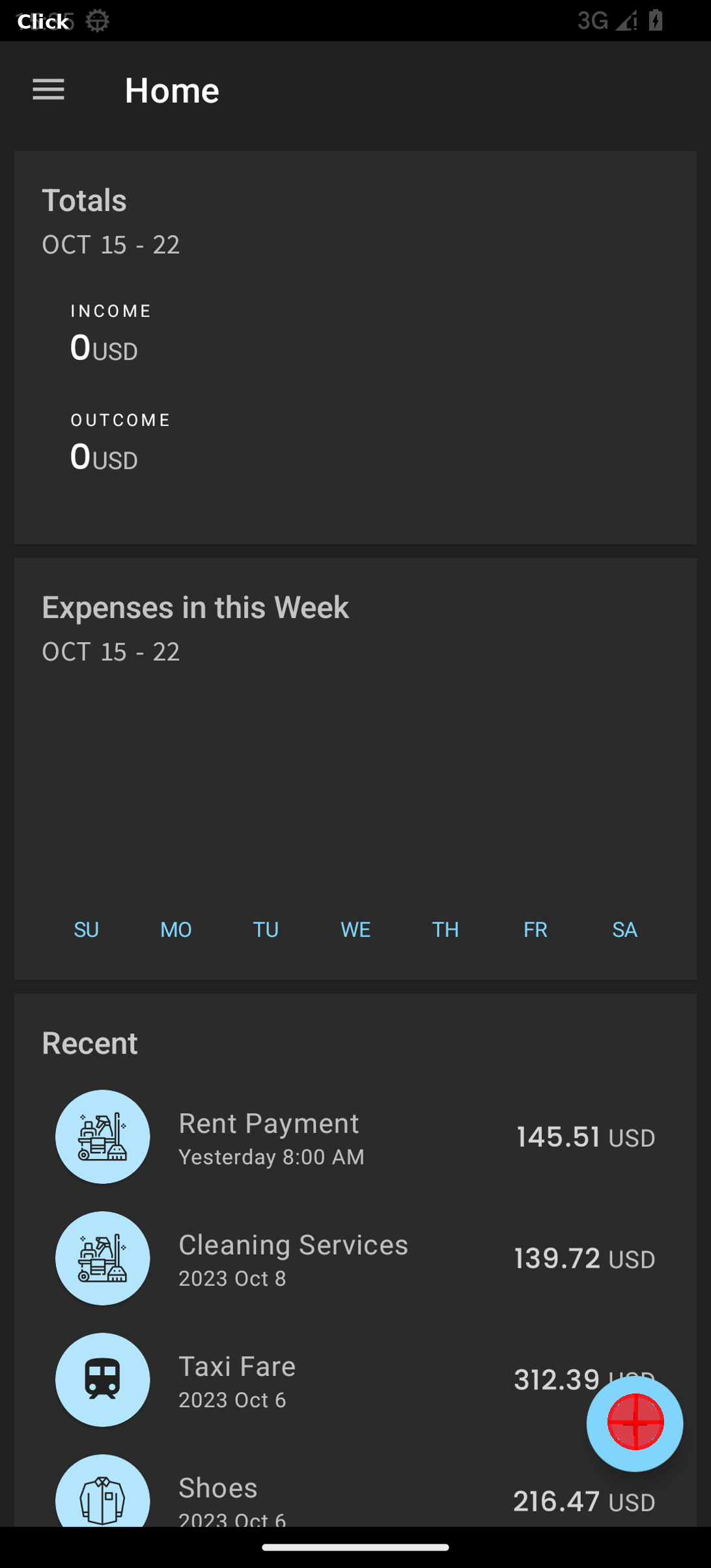}
    }{
        \caseinfo{0.235\textwidth}{
            \casejson{{"action": "click", "coordinate": [894, 906]}}
            \casefield{Verdict}{\casepositive}
        }
    }\hfill
    \casepanel{Step 4}{
        \caseimage{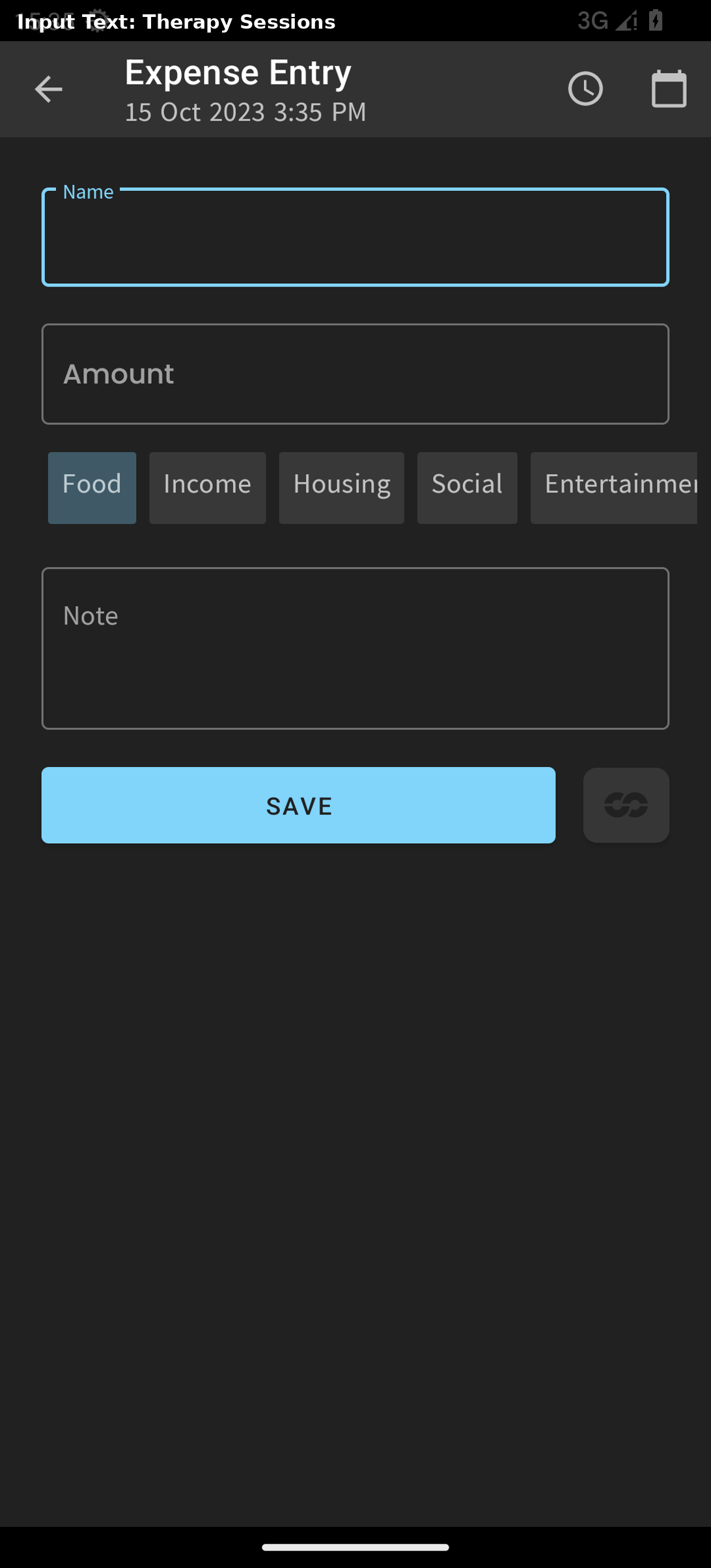}
    }{
        \caseinfo{0.235\textwidth}{
            \casejson{{"action": "type", "text": "Therapy Sessions"}}
            \casefield{Verdict}{\casepositive}
        }
    }\hfill
    \casepanel{Step 5}{
        \caseimage{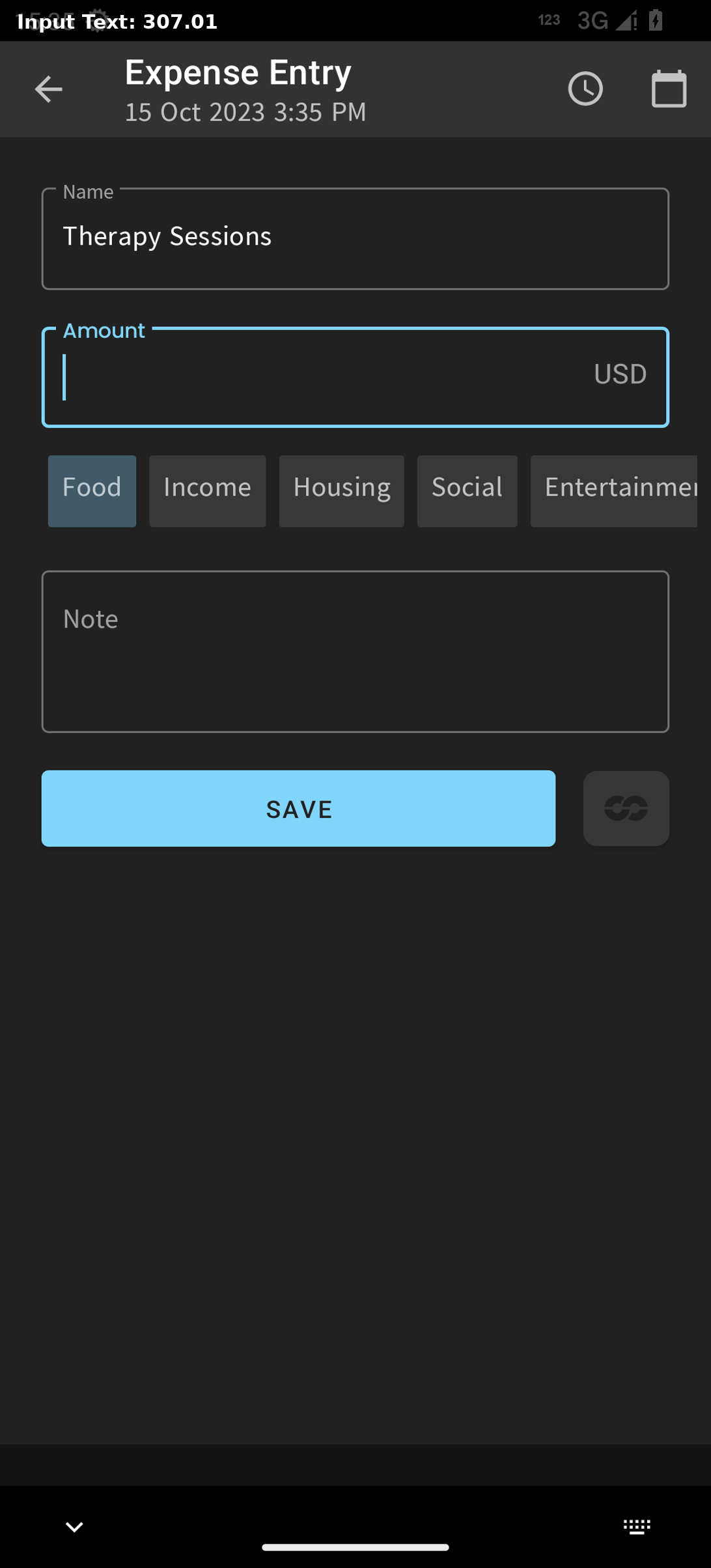}
    }{
        \caseinfo{0.235\textwidth}{
            \casejson{{"action": "type", "text": "307.01"}}
            \casefield{Verdict}{\casepositive}
        }
    }
    \caption{\textbf{AnchorGUI second attempt with cross-trial experience, steps 1--5.} The retry begins with the same setup operations, but now follows a corrected high-level plan distilled from the previous failure.}
    \label{fig:case_anchor_cross_expense}
\end{figure*}

\begin{figure*}[ph!]\ContinuedFloat
    \centering
    \small
    \casepanel{Step 6}{
        \caseimage{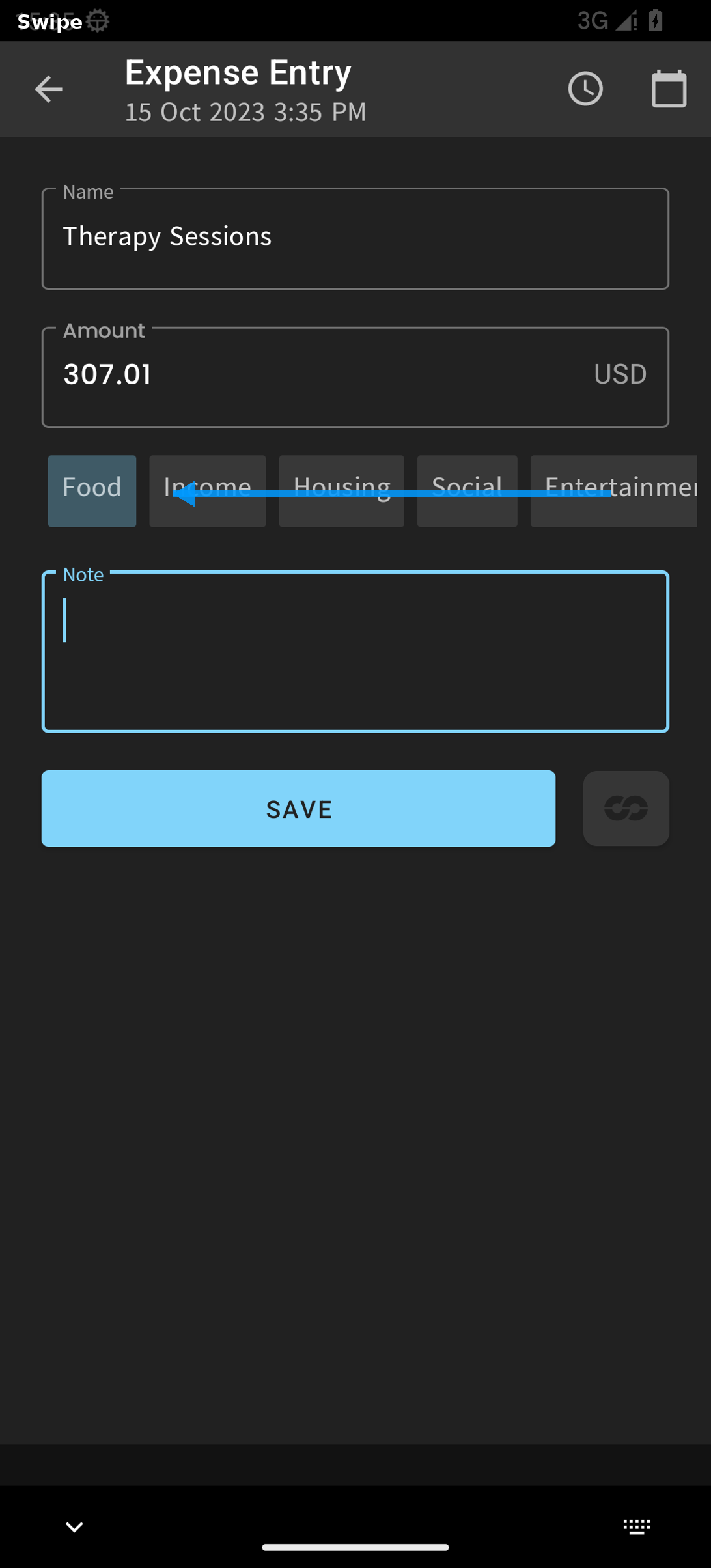}
    }{
        \caseinfo{0.235\textwidth}{
            \casejson{{"action": "swipe", "coordinate": [859, 315], "coordinate2": [244, 315]}}
            \casefield{Verdict}{\casepositive}
        }
    }\hfill
    \casepanel{Step 7}{
        \caseimage{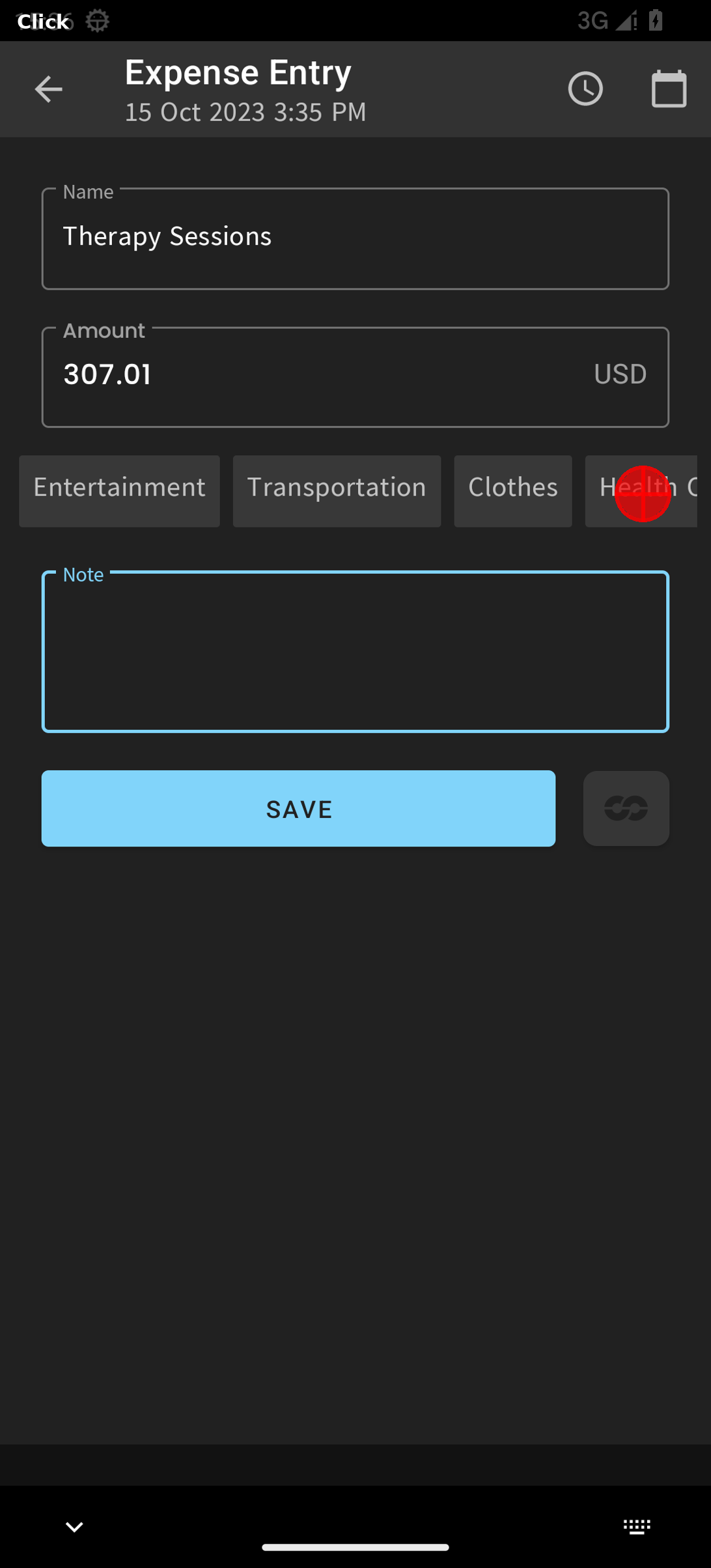}
    }{
        \caseinfo{0.235\textwidth}{
            \casejson{{"action": "click", "coordinate": [904, 315]}}
            \casefield{Verdict}{\casepositive}
        }
    }\hfill
    \casepanel{Step 8}{
        \caseimage{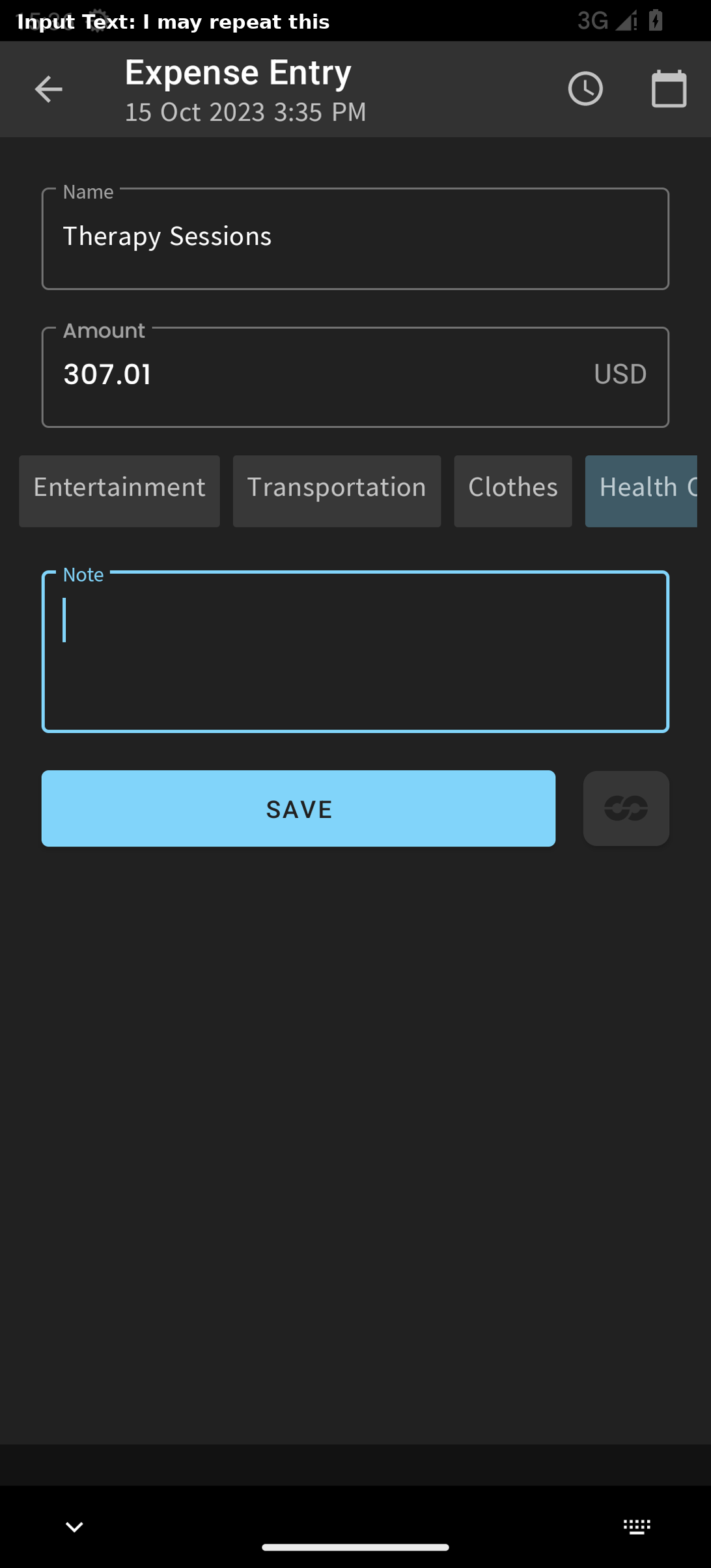}
    }{
        \caseinfo{0.235\textwidth}{
            \casejson{{"action": "type", "text": "I may repeat this"}}
            \casefield{Verdict}{\casepositive}
        }
    }\hfill
    \casepanel{Step 9}{
        \caseimage{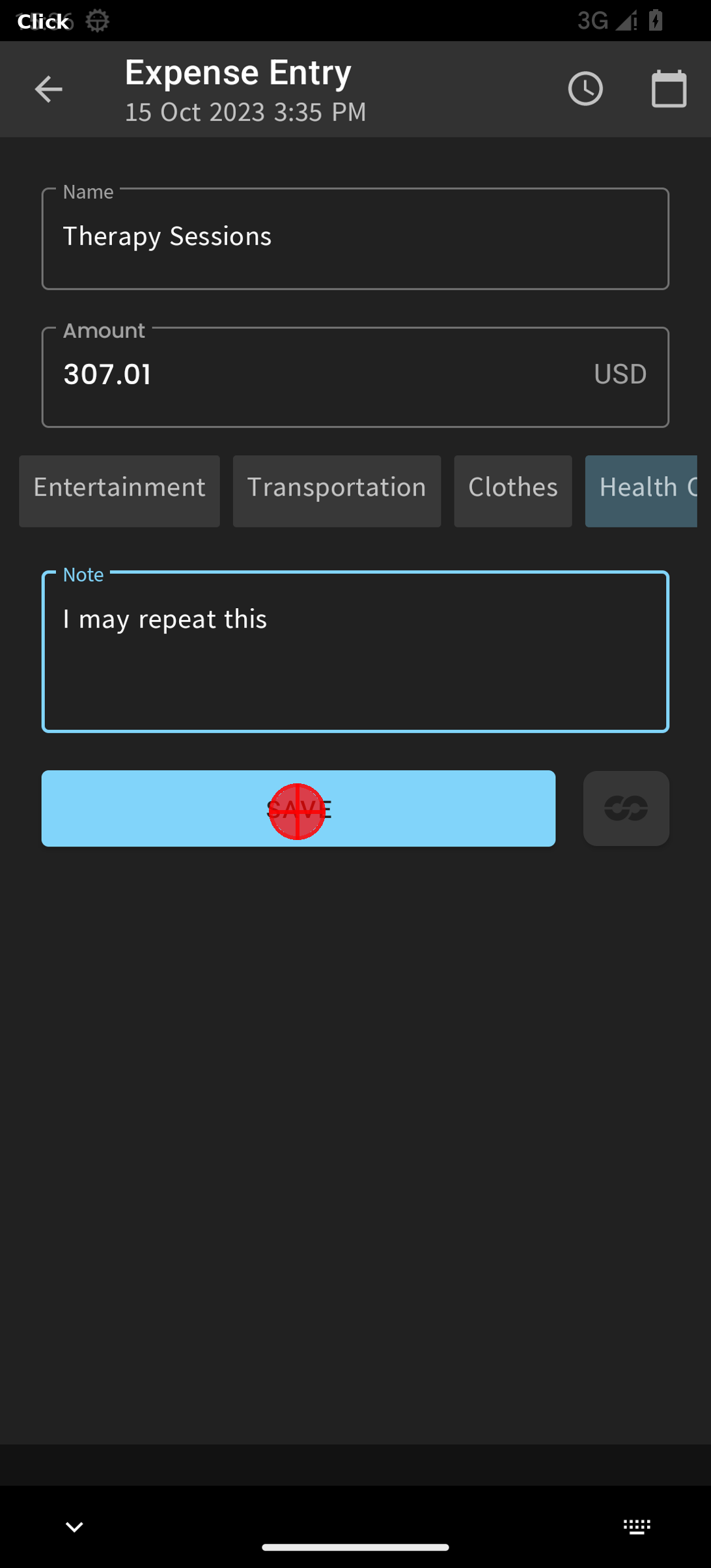}
    }{
        \caseinfo{0.235\textwidth}{
            \casejson{{"action": "click", "coordinate": [418, 517]}}
            \casefield{Verdict}{\casepositive}
        }
    }\hfill
    \casepanel{Step 10}{
        \caseimage{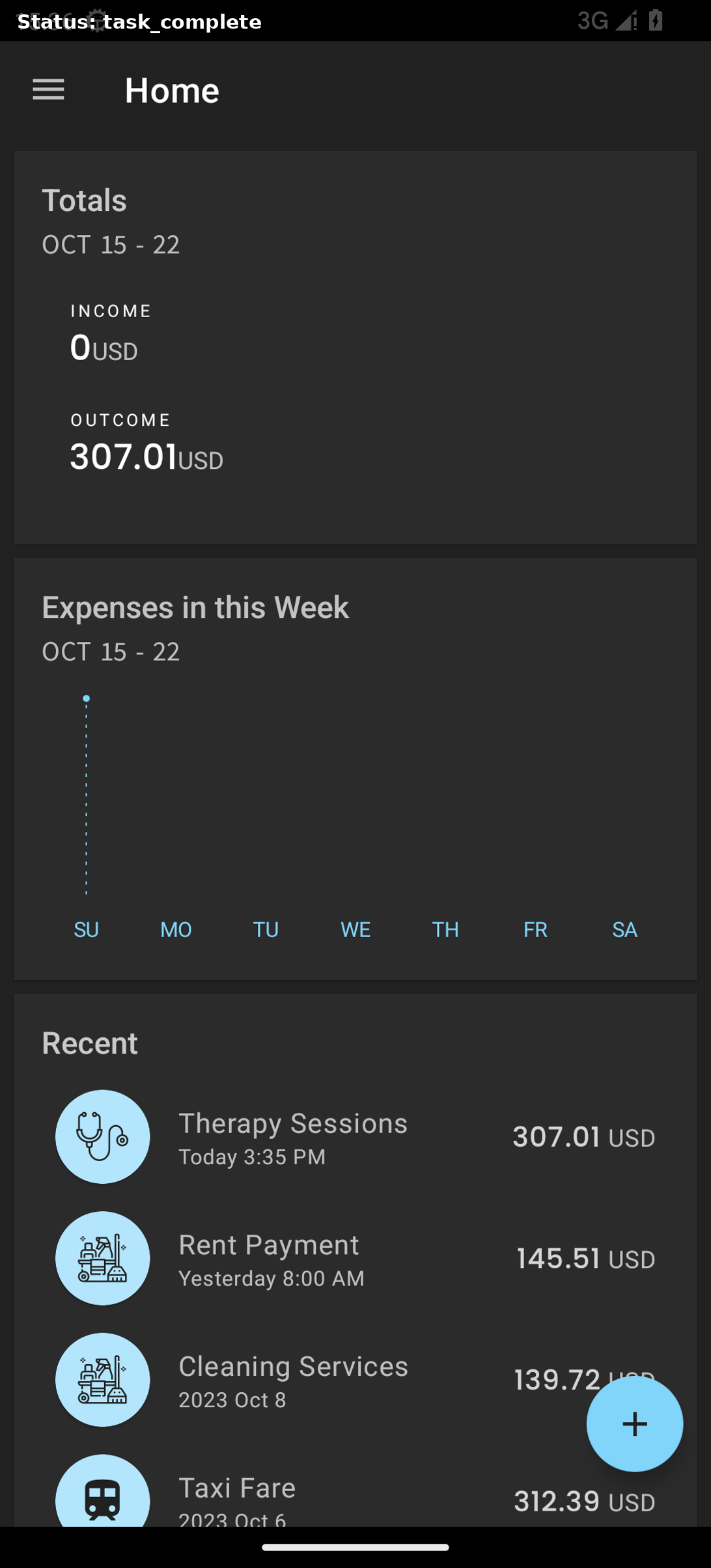}
    }{
        \caseinfo{0.235\textwidth}{
            \casejson{{"action": "terminate", "status": "success"}}
        }
    }
    \caption[]{\textbf{AnchorGUI second attempt with cross-trial experience (continued), steps 6--10.} Again, the local verdicts remain positive throughout, but unlike the first trial they now correspond to the \emph{right} semantics: the agent reveals \emph{Health}, enters the requested note, saves the corrected record, and then terminates.}
\end{figure*}

\clearpage
\section{Future Directions}
\label{sec:supp_future}

\subsection{Continual Learning}
\label{sec:supp_continual_learning}

Our current cross-trial setting distills experience within repeated attempts of the same task. A natural next step is to extend the experience bank into a persistent continual-learning memory that accumulates reusable knowledge across different users, apps, and tasks. Such a memory could store higher-level GUI regularities, such as common pop-up handling patterns, navigation shortcuts, or app-specific interaction conventions, allowing the policy to start from stronger priors even before encountering task-specific failures. An important challenge, however, is to retrieve only the most relevant prior experience while avoiding negative transfer from outdated or user-specific rules. This suggests future work on memory indexing, conflict resolution, and privacy-aware personalization for long-horizon GUI agents.

\subsection{Human-in-the-Loop Refinement}
\label{sec:supp_human_loop}

Another promising direction is to incorporate lightweight human feedback into the CSA loop. In our current framework, evaluator verdicts and distilled rules are produced automatically by the model. In practice, users could intervene when the system assigns a false verdict, misses an important task constraint, or extracts an overly generic rule from a failed trial. Allowing users to correct these intermediate outputs could turn the experience bank into an interactive knowledge base that is progressively refined over time. Beyond improving robustness, such a human-in-the-loop design may also enable faster personalization, since users can directly teach the agent preferred workflows, app-specific habits, or safety constraints that are difficult to infer from visual trajectories alone.

\end{document}